\documentclass[11pt]{article}

\PassOptionsToPackage{dvipsnames}{xcolor}
\usepackage[final]{acl}

\usepackage{microtype}
\usepackage{makecell}
\usepackage{multirow}
\usepackage{amssymb}
\usepackage{amsmath}
 
\usepackage{booktabs}
\usepackage{xcolor}
\usepackage{pgf}
\usepackage{colortbl}

\usepackage{rotating}
\usepackage{setspace}

\usepackage[english,bidi=default]{babel} % English as the main language.
\babelfont{rm}[Path=fonts/,%
  BoldFont=texgyretermes-bold.otf,%
  ItalicFont=texgyretermes-italic.otf,%
  BoldItalicFont=texgyretermes-bolditalic.otf]{texgyretermes-regular.otf} % similar to Times
\babelprovide[import]{hindi}
\babelfont[*devanagari]{rm}[Path=fonts/]{Lohit-Devanagari.ttf}
\babelprovide[import]{arabic}
\babelfont[*arabic]{rm}[Path=fonts/]{NotoSansArabic-Regular.ttf}
\babelprovide[import]{japanese}
\babelfont[*japanese]{rm}[Path=fonts/]{NotoSansCJK-Regular.ttc}

\definecolor{cs_color}{RGB}{124, 88, 55}
\definecolor{az_color}{RGB}{98, 106, 98}
\definecolor{Goldenrod}{RGB}{218, 165, 32}

\usepackage{todonotes}

\newcommand{\gradientcell}[6]{%
    \def\value{#1}%
    \def\minvalue{#2}%
    \def\maxvalue{#3}%
    \def\mincolor{#4}%
    \def\maxcolor{#5}%
    \def\transparency{#6}%
    \ifdimcomp{\value pt}{>}{\maxvalue pt}{\cellcolor{#5!100.0!#4!#6}\value}{%
    \ifdimcomp{\value pt}{<}{\minvalue pt}{\cellcolor{#5!0.0!#4!#6}\value}{%
         \pgfmathparse{int(round(100*(#1/(\maxvalue-\minvalue))-(\minvalue *(100/(\maxvalue-\minvalue)))))}%
        \xdef\tempa{\pgfmathresult}%
        \cellcolor{#5!\tempa!#4!#6}\value%
    }}%
}

\definecolor{LightGray}{gray}{0.95}
\newcommand{\Acc}[1]{\gradientcell{#1}{.3}{.7}{LightGray}{Goldenrod}{60}}
\newcommand{\Agree}[1]{\gradientcell{#1}{.47}{.82}{LightGray}{Green}{60}}
\newcommand{\Corr}[1]{\gradientcell{#1}{.10}{.82}{LightGray}{Red}{60}}

\newcommand{\FP}[1]{%
    \ifdimcomp{#1 pt}{<}{0 pt}{%
        \def\value{#1}%
        \def\minvalue{-.3}%
        \def\maxvalue{0}%
        \def\mincolor{Red}%
        \def\maxcolor{LightGray}%
        \def\transparency{60}%
        \ifdimcomp{\value pt}{>}{\maxvalue pt}{\cellcolor{LightGray!100.0!Red!60}\value}{%
        \ifdimcomp{\value pt}{<}{\minvalue pt}{\cellcolor{LightGray!0.0!Red!60}\value}{%
             \pgfmathparse{int(round(100*(#1/(\maxvalue-\minvalue))-(\minvalue *(100/(\maxvalue-\minvalue)))))}%
            \xdef\tempa{\pgfmathresult}%
            \cellcolor{LightGray!\tempa!Red!60}\value%
        }}%
    }{%
        \def\value{#1}%
        \def\minvalue{0}%
        \def\maxvalue{.3}%
        \def\mincolor{LightGray}%
        \def\maxcolor{Blue}%
        \def\transparency{60}%
        \ifdimcomp{\value pt}{>}{\maxvalue pt}{\cellcolor{Blue!100.0!LightGray!60}\hphantom{-}\value}{%
        \ifdimcomp{\value pt}{<}{\minvalue pt}{\cellcolor{Blue!0.0!LightGray!60}\hphantom{-}\value}{%
             \pgfmathparse{int(round(100*(#1/(\maxvalue-\minvalue))-(\minvalue *(100/(\maxvalue-\minvalue)))))}%
            \xdef\tempa{\pgfmathresult}%
            \cellcolor{Blue!\tempa!LightGray!60}\hphantom{-}\value%
        }}%
    }%
}

\newcommand{\focuscell}[1]{\begin{minipage}{18pt}\centering\setstretch{0.6}#1\end{minipage}}
\newcommand{\AccCell}[2]{%
  \focuscell{\gradientcell{#1}{.3}{.7}{LightGray}{Goldenrod}{60} \\ \scriptsize $\pm$ #2}%
}

\title{MultiInpaintCOCO: Fine‑Grained \\ Visual Understanding in Multiple Languages}

\author{Andrei-Alexandru Manea \\
  Faculty of Mathematics and Physics, Charles University / V Holešovičkách 747/2, Prague, Czech Republic \\
  \texttt{manea@ufal.mff.cuni.cz} \\\And
  Jindřich Libovický \\
  Faculty of Mathematics and Physics, Charles University / V Holešovičkách 747/2, Prague, Czech Republic \\
  \texttt{libovicky@ufal.mff.cuni.cz} \\\And
  Philipp J. Rösch \\
  University of the Bundeswehr Munich \\
  \texttt{philipp.roesch@unibw.de}
}

\author{Andrei-Alexandru Manea$^1$ \and Jindřich Libovický$^1$ \and Philipp J. Rösch$^2$ \\
$^1$Faculty of Mathematics and Physics, Charles University, Czech Republic \\
$^2$Bundeswehr University Munich, Germany \\
\texttt{\{manea, libovicky\}@ufal.mff.cuni.cz} \quad \texttt{philipp.roesch@unibw.de}
}

\begin{document}

\maketitle
\begin{abstract}
Previous research shows that vision-language (VL) models struggle to understand fine-grained image details and tend to focus on the overall layout.
Moreover, any weakness in English tends to be further amplified in the multilingual setting.
To study that, we extend InpaintCOCO, an English dataset focused on fine-grained visual details, to 13 typologically diverse languages across five language families, creating \href{https://huggingface.co/datasets/ufal/multiInpaintCOCO}{MultiInpaintCOCO} -- a new multilingual benchmark of 953 minimal-pair examples.
We further test several state-of-the-art models in multilingual multimodal retrieval, considering both light VL CLIP-like encoders and large language models.
Our results show that performance degrades substantially for lower-resource languages, with the strongest predictor being the data resources available for the languages rather than typological similarity to English.
In cross-lingual retrieval, most models exhibit a systematic bias toward English: when given a correct caption in a non-English language and an incorrect one in English, models preferentially select the English caption.
This bias varies considerably across architectures. Among the tested models, Gemma~3 shows the strongest English preference while Qwen~3 is the least biased, despite both reaching similar monolingual accuracy across languages. The code is available at \url{https://github.com/ufal/multi-inpaint-coco}.
\end{abstract}

\section{Introduction}

Vision-language (VL) models take both images and text as input. Depending on the task, two main types of models are currently used. \emph{Dual encoders}, such as CLIP \citep{pmlr-v139-radford21a} and SigLIP 2 \citep{tschannen2025siglip2multilingualvisionlanguage}, encode images and texts independently into a shared embedding space; they are the standard choice for retrieval and similarity tasks. \emph{Decoder-only models}, such as Gemma 3 \citep{gemmateam2025gemma3technicalreport} and Qwen 2.5-VL \citep{bai2025qwen25vltechnicalreport}, are decoder-based and generate text conditioned on both visual and textual input. Both model types are pre-trained on large image-text corpora using objectives (contrastive or generative) that do not require distinguishing between images or captions that differ only in fine-grained details.

\begin{figure}[t]

\footnotesize
\setlength{\tabcolsep}{1pt}
\begin{tabular}{p{10pt} >{\centering\arraybackslash}p{100pt} >{\centering\arraybackslash}p{100pt}}
\toprule
& Original image & Edited image \\
&
\begin{minipage}[t]{100pt}
    \includegraphics[width=\textwidth]{images/coco_110.png}
\end{minipage}
&
\begin{minipage}[t]{100pt}
    \includegraphics[width=\textwidth]{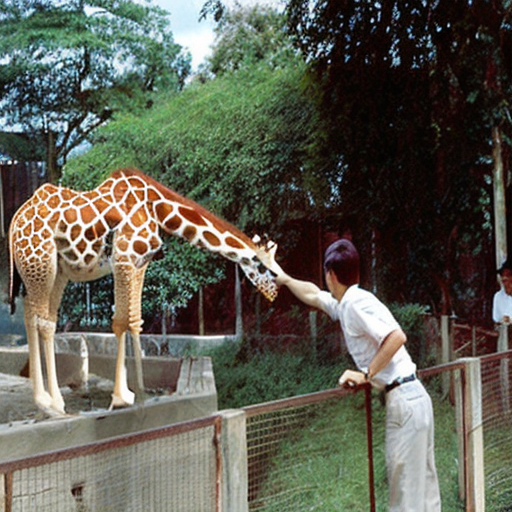}
\end{minipage} \\ \midrule
en & A man is feeding an elephant over a fence & A man is feeding a giraffe over a fence \\
\midrule
\textcolor{cs_color}{cs} & \textcolor{cs_color}{Muž krmí slona přes plot} & \textcolor{cs_color}{Muž krmí žirafu přes plot} \\
\midrule
jp & \foreignlanguage{japanese}{男性が柵越しに象に餌をあげている} & \foreignlanguage{japanese}{男性が柵越しにキリンに餌をあげている} \\
\midrule
\textcolor{az_color}{az} & \textcolor{az_color}{Bir kişi hasarın üstündən bir fili yemləyir} & \textcolor{az_color}{Bir kişi hasarın üstündə zürafəni qidalandırı} \\
\bottomrule

\end{tabular}

\caption{Example from the MultiInpaintCOCO dataset. The left image and English caption are from the COCO dataset, the right equivalent part is from InpaintCOCO.}
\label{fig:mic_sample_1}

\end{figure}

Several studies have shown that VL models fail to capture fine-grained visual information and instead rely on coarse features, such as global scene layout or the most salient objects \citep{fu2025hiddenplainsightvlms}. To probe this limitation precisely, \citet{rosch-etal-2024-enhancing} proposes minimal pairs: each example consists of two images that differ by a single inpainted object and two captions that differ by exactly one keyword, so that the correct answer cannot be reached without identifying the specific visual detail. A similar strategy applied at the caption level underlies benchmarks such as SugarCrepe \citep{hsieh-etal-2023-sugarcrepe}. The image-editing approach has a key advantage: it introduces genuine visual ambiguity, because neither the image nor the caption alone is sufficient to determine the correct match.

All of this work is English-only. It is well established that VL model performance drops substantially when moving from English to lower-resource languages \citep{liu-etal-2021-visually, pmlr-v162-bugliarello22a}, and that multilingual models are not equally calibrated across languages. Errors in less-resourced languages may therefore not only be more frequent but also qualitatively different. The combination of fine-grained visual understanding and multilingual evaluation is therefore almost entirely unexplored. To close this gap, we extend InpaintCOCO \citep{rosch-etal-2024-enhancing} to 13 typologically diverse languages (Azerbaijani, Arabic, Czech, German, Greek, Hindi, Italian, Japanese, Romanian, Russian, Slovak, Ukrainian, Vietnamese), producing the MultiInpaintCOCO dataset. Captions were first translated automatically and then post-edited by native speakers.

We evaluate several state-of-the-art dual-encoder and generative decoder models on MultiInpaintCOCO and report two main findings. First, performance on lower-resource languages is substantially worse than on English and other high-resource languages, confirming that fine-grained retrieval is harder in the multilingual setting. Second, and more surprisingly, when a model is given one correct caption in a non-English language and one incorrect caption in English, most models systematically prefer the English caption. This bias appears in both dual-encoder and generative decoder-only models, with Azerbaijani the least favored language. The strength of the bias varies considerably across models: Gemma~3 exhibits the strongest preference for English, while Qwen~3 is the least biased. This is not just a reflection of lower overall accuracy in weaker languages. It reveals a systematic preference for English that persists even when the model is otherwise capable of making the correct match.

\section{Dataset Description}

MultiInpaintCOCO extends the English InpaintCOCO dataset \citep{rosch-etal-2024-enhancing} to 13 more languages, producing a multilingual minimal-pair benchmark for fine-grained vision-language retrieval. The dataset covers 14 languages in total and spans five language families, with translations produced automatically and then post-edited by native speakers.

\subsection{InpaintCOCO}

InpaintCOCO \citep{rosch-etal-2024-enhancing} is a minimal-pair dataset built on top of COCO \citep{lin2014microsoft}. For each example, a keyword is replaced in the original caption, and the corresponding image region is edited using Stable Diffusion inpainting \citep{rombach2022high}, producing a pair of images that differ in a single localized detail and a pair of captions that reflect that difference. This construction ensures that a model can only retrieve the correct image--caption match by attending to fine-grained visual content rather than overall scene similarity.

Before translation, we manually reviewed all 1{,}260 examples in InpaintCOCO and removed or corrected low-quality instances. We discarded examples in which the inpainted image contained visible artifacts or the edit was too subtle to warrant a different caption, and we corrected captions that were inaccurate or inconsistent with the edited image. After this review, 953 examples remained.

\subsection{Translation Process}

For each language, captions were first translated automatically and then post-edited by a native speaker. Three translation systems were available to the annotators: Hunyuan \citep{zheng-etal-2025-shy} (the WMT25 winner; \citealp{kocmi-etal-2025-findings}), Google Translate, and GPT-4. Annotators were free to choose whichever system they found most useful as a starting point and were instructed to correct any errors in the output. All annotators were native speakers of the target language with an excellent command of English. The translation into Russian was post-edited, with German as the source language. More details about the annotation process are in the Appendix~\ref{ap:annotation}.

\subsection{The MultiInpaintCOCO Dataset}

The final dataset covers 14 languages across five language families: Indo-European, Japonic, Turkic, Austroasiatic, and Afro-Asiatic, with Indo-European being the most represented (10 languages). The full list of languages with their metadata is given in Table~\ref{tab:languages}. The languages vary substantially in their resource levels: following the tiers of \citet{joshi-etal-2020-state}, the dataset includes four tier-5 languages (English, German, Japanese, Arabic), five tier-4 languages, four tier-3 languages, and one tier-1 language (Azerbaijani). Morphologically, the majority are fusional (10), with two agglutinative languages (Japanese, Azerbaijani) and two isolating languages (English, Vietnamese).

The disparity in the availability of pretraining data across our languages is stark. In FineWeb2 \citep{penedo2025fineweb}, a 20 TB multilingual web corpus covering over 1,000 languages, Russian alone accounts for 6,400 GB, while German and Japanese each contribute around 1,650 GB. At the other extreme, Azerbaijani has only 29 GB of data, roughly 220 times less than Russian. Slovak (92 GB) and Hindi (130 GB) also fall well below the high-resource languages despite having tens of millions of native speakers. This orders-of-magnitude gap in raw text data directly motivates our expectation (and findings) of performance differences across languages in our experiments.

Table~\ref{tab:datasets} places MultiInpaintCOCO in the context of existing multilingual vision-language benchmarks \citep{liu-etal-2021-visually, nooralahzadeh2022, bugliarello2022, romero2024cvqaculturallydiversemultilingualvisual, pfeiffer-etal-2022-xgqa, schneider-sitaram-2024-m5}. Most existing benchmarks are either translation-based with a narrow language selection or culturally grounded but limited in scale. MultiInpaintCOCO occupies a distinct position: it is translation-based, enabling direct cross-lingual comparison, and the only benchmark that combines caption-level and image-level minimal pairs, allowing retrieval errors to be attributed to either the text or the visual representation. In terms of linguistic diversity, measured by the Grambank feature inclusion index (GBI; \citealp{ploeger-etal-2024-typological}), our dataset scores 79, which is higher than all other translation-based benchmarks in Table~\ref{tab:datasets}, such as xFlickrCo \citep{bugliarello2022} (65) or XTD \citep{xtd} (71), and comparable to culturally grounded ones such as KnowRecall \citep{wang2025travelinglanguagesbenchmarkingcrosslingual} (77.94) and CVQA \citep{romero2024cvqaculturallydiversemultilingualvisual} (88). Still, MultiInpaintCOCO's 14 languages represent only 19\% of the world's population when considering languages spoken as first languages.

\section{Multilingual Experiments}\label{sec:multingual}

We evaluate both dual-encoder and decoder-only models on MultiInpaintCOCO across all 14 languages to assess how well current VL models handle fine-grained visual details in a multilingual setting. Beyond raw model performance, we measure consistency across languages and across models.

\subsection{Experimental Setup}

\paragraph{Task formulation.} Each example in MultiInpaintCOCO consists of an original image--caption pair and an edited image--caption pair. We construct two retrieval subtasks from this structure. In the \emph{two-image} task (\textit{2img}), the model is given one caption and two images and must select the image that matches the caption. In the \emph{two-caption} task (\textit{2txt}), the model is given one image and two captions and must select the caption that describes the image. In both cases, either the original or the edited item serves as the query, while the two remaining items of the other modality serve as the context.

\paragraph{Dual encoders.} Dual encoders map images and texts independently into a shared embedding space. For each example, we embed both images and both captions and compute all pairwise cosine similarities. The predicted match is the image (or caption) that is most similar to the query. This evaluation protocol follows \citet{rosch-etal-2024-enhancing}.

\paragraph{Decoder-only models.} Decoder-only models process inputs sequentially. We present the model with both options (\emph{two-image} or \emph{two-caption}) and a prompt asking which option is correct. The instruction prompt (see Appendix~\ref{app:hyperparams}) is written in English, regardless of the language of the captions. We are aware that this practice might amplify the English bias. To explore this, we also test with French (a language not in the current dataset), and the \textit{strict accuracy} is generally lower (Table~\ref{tab:strict_acc_fr_prompt}), but the differences between languages follow the same patterns. We therefore decide to stick with the English prompts. Because decoder-only models are known to be sensitive to input order \citep{wardle2024imagetextfirstoptimising,tian2025identifying}, we evaluate each example in both orders and require both to be correct, yielding eight comparison configurations per example in total (see Table~\ref{tab:configs_2img}). A detailed evaluation of the position bias of the models used in this paper is provided in Appendix~\ref{app:position}.

\paragraph{Model performance.} We report \textit{strict accuracy} as a primary metric. A dataset item is counted as correct only if the model answers correctly across all comparison configurations: four for order-invariant dual encoders and eight for decoder-only models.

Formally, each example consists of an original image-caption pair $(I_\text{o}, T_\text{o})$ and an edited pair $(I_\text{e}, T_\text{e})$. We construct four binary correctness
indicators: $c(I_\text{o}, \{T_\text{o}, T_\text{e}\})$ and
$c(I_\text{e}, \{T_\text{o}, T_\text{e}\})$ for the \textit{2txt} task, and
$c(T_\text{o}, \{I_\text{o}, I_\text{e}\})$ and
$c(T_\text{e}, \{I_\text{o}, I_\text{e}\})$ for the \textit{2img} task, where $c(q, \{a, b\}) = 1$, if the model selects the correct item from the pair given the query $q$. 
For the sake of notation simplicity, we consider that $c(q, \{a, b\})$ contains the results in both input orders for decoder-only models; therefore, $ c(q, \{a, b\}) = c(q, (a, b)) \cdot c(q, (b, a))$. 

The \textit{2img} and \textit{2txt} accuracies are the averages of the corresponding
indicators over all examples. \textit{Strict accuracy} requires all four to be correct:
\begin{eqnarray}
    c_{\textit{2img}} & = & c(T_\text{o}, \{I_\text{o}, I_\text{e}\})
    \cdot c(T_\text{e}, \{I_\text{o}, I_\text{e}\})  \label{eq:c2img} \\
    c_\textit{2txt} & = & c(I_\text{o}, \{T_\text{o}, T_\text{e}\}) 
    \cdot c(I_\text{e}, \{T_\text{o}, T_\text{e}\}) \label{eq:c2txt} \\
    c_{\text{strict}} & = &  c_\textit{2img} \cdot c_\textit{2txt} \label{eq:cstrict} 
\end{eqnarray}
and \textit{strict accuracy} is the average of $c_{\text{strict}}$ over all examples.

\paragraph{Factors influencing model performance.} Because we do not have access to the training data or methodology of the models we study, we can only partially explain cross-lingual performance differences. To assess that, we compute the Spearman correlation between \textit{strict accuracy} and a set of language-level variables listed in Table~\ref{tab:lang_proper}: resource tier \citep{joshi-etal-2020-state}, number of native speakers according to Wikidata\footnote{\url{https://www.wikidata.org/wiki/Property:P1098}} \citep{vrandecic2014wikidata}, typological similarity to English and Chinese from the URIEL+ database \citep{khan-etal-2025-uriel}, number of Wikipedia articles, and the amount of available text in the FineWeb2 corpus \citep{penedo2025fineweb}. Further, we consider the number of tokens into which our dataset is tokenized per language (see Table~\ref{tab:tokenizer} in the Appendix).

\paragraph{Language agreement.}

Similar accuracy scores across languages do not necessarily mean that models make the same errors.
To measure how similar the error patterns are across languages, we compute Cohen's $\kappa$, a standard metric for inter-annotator agreement, between the per-language model predictions. The agreement between two languages, $l_1$ and $l_2$, inside the model $m$ is represented by $C(m, l_1, l_2)$, and the average agreement across languages $A(l_1, l_2)$ is computed as follows:
\begin{eqnarray}
    C(m, l_1, l_2) & = & \kappa(\hat{y}_{m}(l_1), \hat{y}_{m}(l_2)), \\
    A(l_1, l_2) & = &\frac{1}{|M|}\sum_{m} C(m, l_1, l_2)
\end{eqnarray}
where $\hat{y}_{m}(l)$ is the output of the model $m$, in language $l$. A value of 0 indicates that two languages make errors on completely different examples, while a value of 1 means their error patterns are identical. 

\paragraph{Model agreement.}

We use the same method to compare error patterns across models.
For two models $m_1$ and $m_2$, in the language $l$, we compute the individual agreement $C(m_1, m_2, l)$ and the average agreement $A(m_1, m_2)$ as:
\begin{eqnarray}
    C(m_1, m_2, l) & = & \kappa(\hat{y}_{m_1}(l), \hat{y}_{m_2}(l)), \\
    A(m_1, m_2) & = & \frac{1}{|L|}\sum_{l} C(m_1, m_2, l).
\end{eqnarray}

\begin{table*}[htbp]
\centering\footnotesize
\setlength{\tabcolsep}{4.2pt}
\begin{tabular}{lcccccccccccccc@{\hskip 16pt}c}
\toprule
Model & ar & az & cs & de & el & en & hi & it & ja & ro & ru & sk & uk & vi & Avg. \\
\midrule
Gemma 3 12B & \Acc{.581} & \Acc{.425} & \Acc{.507} & \Acc{.581} & \Acc{.587} & \Acc{.569} & \Acc{.523} & \Acc{.580} & \Acc{.510} & \Acc{.570} & \Acc{.610} & \Acc{.472} & \Acc{.594} & \Acc{.599} & \Acc{.551} \\
Qwen 2.5 VL 7B & \Acc{.595} & \Acc{.335} & \Acc{.586} & \Acc{.690} & \Acc{.571} & \Acc{.714} & \Acc{.528} & \Acc{.664} & \Acc{.607} & \Acc{.571} & \Acc{.686} & \Acc{.544} & \Acc{.630} & \Acc{.621} & \Acc{.596} \\
Qwen 3 VL 8B & \Acc{.677} & \Acc{.504} & \Acc{.654} & \Acc{.700} & \Acc{.705} & \Acc{.723} & \Acc{.695} & \Acc{.705} & \Acc{.686} & \Acc{.678} & \Acc{.698} & \Acc{.623} & \Acc{.677} & \Acc{.685} & \Acc{.672} \\
Aya Vision 8B & \Acc{.244} & \Acc{.125} & \Acc{.296} & \Acc{.410} & \Acc{.407} & \Acc{.496} & \Acc{.301} & \Acc{.366} & \Acc{.335} & \Acc{.359} & \Acc{.379} & \Acc{.225} & \Acc{.346} & \Acc{.424} & \Acc{.337} \\
Jina VL & \Acc{.017} & \Acc{.002} & \Acc{.005} & \Acc{.034} & \Acc{.015} & \Acc{.019} & \Acc{.007} & \Acc{.021} & \Acc{.022} & \Acc{.012} & \Acc{.010} & \Acc{.007} & \Acc{.015} & \Acc{.033} & \Acc{.016} \\
\midrule
NLLB-SigLIP Base & \Acc{.665} & \Acc{.338} & \Acc{.381} & \Acc{.488} & \Acc{.661} & \Acc{.678} & \Acc{.651} & \Acc{.478} & \Acc{.635} & \Acc{.450} & \Acc{.682} & \Acc{.405} & \Acc{.654} & \Acc{.516} & \Acc{.549} \\
NLLB-SigLIP Large & \Acc{.681} & \Acc{.443} & \Acc{.570} & \Acc{.599} & \Acc{.689} & \Acc{.699} & \Acc{.698} & \Acc{.643} & \Acc{.656} & \Acc{.608} & \Acc{.685} & \Acc{.573} & \Acc{.671} & \Acc{.551} & \Acc{.626} \\
MEXMA-SigLIP2 & \Acc{.653} & \Acc{.657} & \Acc{.668} & \Acc{.687} & \Acc{.654} & \Acc{.678} & \Acc{.684} & \Acc{.677} & \Acc{.642} & \Acc{.677} & \Acc{.679} & \Acc{.671} & \Acc{.674} & \Acc{.654} & \Acc{.668} \\
SigLIP2 Base & \Acc{.630} & \Acc{.339} & \Acc{.645} & \Acc{.686} & \Acc{.587} & \Acc{.719} & \Acc{.501} & \Acc{.674} & \Acc{.450} & \Acc{.591} & \Acc{.687} & \Acc{.620} & \Acc{.623} & \Acc{.641} & \Acc{.599} \\
SigLIP2 Large & \Acc{.639} & \Acc{.397} & \Acc{.672} & \Acc{.709} & \Acc{.634} & \Acc{.722} & \Acc{.617} & \Acc{.696} & \Acc{.460} & \Acc{.639} & \Acc{.698} & \Acc{.643} & \Acc{.655} & \Acc{.668} & \Acc{.632} \\
SigLIP2 SO400M & \Acc{.633} & \Acc{.427} & \Acc{.664} & \Acc{.705} & \Acc{.659} & \Acc{.705} & \Acc{.611} & \Acc{.689} & \Acc{.511} & \Acc{.661} & \Acc{.698} & \Acc{.659} & \Acc{.660} & \Acc{.653} & \Acc{.638} \\
SigLIP2 Giant & \Acc{.627} & \Acc{.379} & \Acc{.674} & \Acc{.679} & \Acc{.641} & \Acc{.684} & \Acc{.573} & \Acc{.672} & \Acc{.490} & \Acc{.655} & \Acc{.677} & \Acc{.659} & \Acc{.650} & \Acc{.660} & \Acc{.623} \\
\midrule
Avg. & \Acc{.554} & \Acc{.364} & \Acc{.527} & \Acc{.581} & \Acc{.567} & \Acc{.617} & \Acc{.532} & \Acc{.572} & \Acc{.500} & \Acc{.539} & \Acc{.599} & \Acc{.508} & \Acc{.571} & \Acc{.559} & \Acc{.542} \\
\bottomrule
\end{tabular}
\caption{\textit{Strict accuracy} for bidirectional retrieval, i.e. $c_{\text{strict}}$ from Equation~(\ref{eq:cstrict}). The upper part of the tables contains LLMs, while the lower part shows the dual encoders. The breakdown into \textit{2img} and \textit{2txt} accuracy is reported in the Appendix in Tables~\ref{tab:results_1} and~\ref{tab:results_2}.}
\label{tab:results}
\end{table*}

\subsection{Tested Models}

We evaluate two model families: decoder-only generative VL models and dual-encoder CLIP-like models. Tables~\ref{tab:model-details} and~\ref{tab:training_data} provide additional model details.
% and Section~\ref{app:hyperparams} describes the instruction prompt and other hyperparameters.

%Table contains details about the languages to which every model was exposed during pre-training, based on the available information in their technical reports. For some models, we know only the number of languages in the pre-training, but the exact list is not reported, so we mark these cases with $?$.

\paragraph{Decoder-only models.} We evaluate five open-weight VL models with a parameter count ranging from 2B to 12B. Gemma 3 \citep{gemmateam2025gemma3technicalreport} (12B parameters), Qwen VL 2.5 \citep{bai2025qwen25vltechnicalreport} (7B), Qwen VL 3 \citep{bai2025qwen3vltechnicalreport} (8B), Aya Vision \citep{dash2025ayavisionadvancingfrontier} (8B), and Jina VL \citep{koukounas2026jinavlmsmallmultilingualvision} (2B), the smallest model in our evaluation.

\paragraph{Dual encoder models.} We evaluate seven CLIP-like models \citep{pmlr-v139-radford21a}, all based on SigLIP \citep{zhai2023sigmoidlosslanguageimage}. This includes NLLB-SigLIP \citep{visheratin2023nllbcliptrainperformant} in base and large sizes, MEXMA-SigLIP 2, and SigLIP 2 \citep{tschannen2025siglip2multilingualvisionlanguage} from base to giant sizes. The vision encoders of SigLIP 2 are also used in Gemma and Qwen. 
The NLLB versions consist of the first SigLIP model, pre-trained across all languages in NLLB-200 \citep{nllbteam2022languageleftbehindscaling}. MEXMA-SigLIP 2 combines the MEXMA text encoder \citep{janeiro2024mexmatokenlevelobjectivesimprove} with the SigLIP visual backbone.

\subsection{Results}

\paragraph{Model performance.} We report the \textit{strict accuracy} in Table~\ref{tab:results}. Table~\ref{tab:results_1} in the appendix contains the accuracy for the \textit{2img} task, and Table~\ref{tab:results_2} for $\textit{2txt}$. On average, deciding between two images is slightly harder than deciding between two texts for all the models. This is expected, given that in the \textit{2txt} task, the query image is fixed across languages, and errors are driven mainly by shared visual ambiguity. However, in the \textit{2img} setup, the query is language-specific, so the per-language text-encoding quality introduces additional variation. The Jina model has the highest modality performance gap, while Qwen 3 has the lowest. Azerbaijani (tier-1) has the lowest performance, which is correlated with it being the least-resourced language we have. On the other hand, with dual encoders that rely on the SigLIP 2 text encoder, we observe a performance drop for Japanese, a high-resource language (tier-5). Moreover, for the same models, the gap between the \textit{2img}/\textit{2txt} accuracies and \textit{strict accuracy} is unusually large for Hindi. Finally, MEXMA-SigLIP 2 performs best on Azerbaijani.

\paragraph{Statistical significance.} To verify the robustness of our findings, we estimated 95\% confidence intervals using bootstrap resampling with 10K samples (Table~\ref{tab:strict_acc_bootstrap}). With 953 examples per language, the intervals are roughly $\pm 3$ percentage points, and the main effects discussed above are statistically significant: the drop for Azerbaijani (with MEXMA-SigLIP 2 significantly outperforming all other models there), the drop for Japanese for SigLIP 2, and the English lead for Aya Vision. The overlapping intervals appear mostly among higher-resource languages and among the strongest models, where performance is similar: e.g., for Qwen 3, all languages except Azerbaijani, Czech, and Slovak overlap with English, and for MEXMA-SigLIP 2, all 14 languages overlap.

\paragraph{Factors influencing model performance.}

Table~\ref{tab:lang_proper} shows Spearman correlations between \textit{strict accuracy} and a set of language-level variables. The strongest predictor across models is the amount of text in the FineWeb2 corpus ($\bar{\rho} = .56$), an open-source web crawl derived from CommonCrawl that reflects the amount of data realistically available for a given language on the web. The number of Wikipedia articles follows closely ($\bar{\rho} = .53$) and can be understood as a proxy for the same underlying quantity. 
Even though the exact amount and composition of the training data are unknown to the public, these two proxy variables suggest that the volume of pretraining data might be the dominant factor in cross-lingual performance differences.

Typological similarity to English correlates moderately ($\bar{\rho} = .46$) and less strongly than data-size proxies, indicating that the volume of training data matters more than cross-lingual transfer from English. 
The notably low correlation between English similarity and performance for NLLB-SigLIP models ($.18$ for both Base and Large) is consistent with the finding in Section~\ref{sec:cross_lingual} that these models are less English-centric than the rest.
%Interestingly, for models with a stronger Chinese component (Qwen family), similarity to English still correlates more than similarity to Chinese ($\bar{\rho} = .34$). 
For models with a stronger Chinese component (Qwen family), similarity to Chinese correlates slightly more than similarity to English; averaged over all models, however, the correlation with similarity to Chinese is only $.34$. 

%Whereas tokenization length appears to be a relatively reliable proxy for training data size in language and domain match for some models, the average correlation is only 0.41. Tokenizers are often reused (Qwen 2.5, Qwen 3, and Jina have the same tokenizers; tokenizers are shared in the SigLIP families, too) and might not reflect the actual training data. 
Languages that require fewer tokens tend to perform better, especially in the SigLIP 2 family and Jina VL, but not in the MEXMA and NLLB families. Among the two largest decoder-only models, this effect is relatively strong for Gemma 3 but much weaker for Qwen 3 VL. However, the average correlation of $.30$ shows that token count remains only a rough proxy.
%A notable counter example is that the Qwen tokenizers overtokenize Greek and Hindi 2.8$\times$ compared to English, and still, these languages perform comparably to other languages, and Azerbaidjani, with relatively long tokens, is much worse.
A notable counterexample, shown in Table~\ref{tab:tokenizer}, is that the Qwen tokenizers overtokenize Greek and Hindi by 2.8$\times$ compared to English, yet these languages perform comparably to other languages. Meanwhile, Azerbaijani, despite having relatively long tokens, performs much worse.

Finally, Joshi's resource tiers are the weakest explanatory factor ($\bar{\rho} =.34$). This is likely for two reasons: the tier classification partly reflects the availability of annotated linguistic resources (treebanks, lexicons, and so on), which are not directly relevant for vision-language pretraining; and the taxonomy is now six years old~\citep{joshi-etal-2020-state} and might not reflect current data availability. The total number of native speakers is slightly more relevant ($\bar{\rho} =.37$). 

\paragraph{Language agreement.} The average agreement between languages is plotted in Table~\ref{tab:cohen_kappa_lang_corr}. The per-model breakdown is in the appendix (Table~\ref{tab:cohen_kappa_lang_corr_all}). Above the diagonal, the correlations within the \textit{2img} task are considerably lower than those in the \textit{2txt} task below. All the languages agree the least with Azerbaijani, making it an outlier language. All four SigLIP 2 models have a similar disregard for Japanese. On the other hand, the Qwen family of models shows the highest correlations, while Jina shows the lowest, which suggests a chaotic pattern in its output error. The individual model tables show no language outliers for Gemma 3, Qwen 3, and MEXMA-SigLIP 2. Jina pattern is totally random in the \textit{2img} task, and Czech is an outlier in the NLLB-SigLIP base \textit{2txt} setup.

\paragraph{Model agreement.} The average agreement between models is plotted in Table~\ref{tab:cohen_kappa_model_corr}. The per-language breakdown is in the appendix (Table~\ref{tab:cohen_kappa_model_corr_1} and Table~\ref{tab:cohen_kappa_model_corr_2}). The average agreement is not as consistent as it is between languages. Gemma 3 and the Qwen family of models agree with each other. So do the SigLIP 2 models, but not with the NLLB-SigLIP. The highest correlation numbers are found in English, and the lowest in Azerbaijani. Moreover, in the Azerbaijani table, models do not agree with each other at all. In Japanese, the disagreement between SigLIP 2 models differs from that of other models.

\begin{table*}[t]
\centering\footnotesize
\begin{tabular}{lccccc@{\hskip 16pt}ccccccc@{\hskip 16pt}c}
\toprule
Property & \rotatebox{90}{Gemma 3 12B} & \rotatebox{90}{Qwen 2.5 VL 7B} & \rotatebox{90}{Qwen 3 VL 8B} & \rotatebox{90}{Aya Vision 8B} & \rotatebox{90}{Jina VL} & \rotatebox{90}{MEXMA-SigLIP2} & \rotatebox{90}{NLLB-SigLIP Base} & \rotatebox{90}{NLLB-SigLIP Large} & \rotatebox{90}{SigLIP2 Base} & \rotatebox{90}{SigLIP2 Large} & \rotatebox{90}{SigLIP2 SO400M} & \rotatebox{90}{SigLIP2 Giant} & \rotatebox{90}{Average} \\
\midrule
FineWeb2 (GB) & \Corr{.76} & \Corr{.58} & \Corr{.75} & \Corr{.45} & \Corr{.36} & \Corr{.21} & \Corr{.62} & \Corr{.39} & \Corr{.72} & \Corr{.68} & \Corr{.58} & \Corr{.60} & \Corr{.56} \\
Wikipedia Articles & \Corr{.73} & \Corr{.43} & \Corr{.70} & \Corr{.21} & \Corr{.29} & \Corr{.61} & \Corr{.47} & \Corr{.42} & \Corr{.66} & \Corr{.66} & \Corr{.56} & \Corr{.56} & \Corr{.53} \\
Similarity to en & \Corr{.60} & \Corr{.36} & \Corr{.42} & \Corr{.54} & \Corr{.25} & \Corr{.68} & \Corr{.18} & \Corr{.18} & \Corr{.49} & \Corr{.58} & \Corr{.62} & \Corr{.55} & \Corr{.46} \\
Similarity to zh & \Corr{.51} & \Corr{.46} & \Corr{.44} & \Corr{.45} & \Corr{.10} & \Corr{.53} & \Corr{.37} & \Corr{.40} & \Corr{.13} & \Corr{.19} & \Corr{.32} & \Corr{.20} & \Corr{.34} \\
Number of tokens & \Corr{.38} & \Corr{.27} & \Corr{.22} & \Corr{.14} & \Corr{.66} & \Corr{-.08} & \Corr{-.24} & \Corr{-.12} & \Corr{.64} & \Corr{.62} & \Corr{.53} & \Corr{.59} & \Corr{.30} \\
Joshi's Tier & \Corr{.52} & \Corr{.22} & \Corr{.50} & \Corr{-.04} & \Corr{.53} & \Corr{.08} & \Corr{.44} & \Corr{.33} & \Corr{.50} & \Corr{.42} & \Corr{.27} & \Corr{.35} & \Corr{.34} \\
Speakers (M) & \Corr{.70} & \Corr{.46} & \Corr{.61} & \Corr{-.01} & \Corr{.33} & \Corr{.30} & \Corr{.70} & \Corr{.63} & \Corr{.32} & \Corr{.22} & \Corr{.10} & \Corr{.12} & \Corr{.37} \\
\bottomrule
\end{tabular}
\caption{Spearman correlation of the \textit{strict accuracy} with the language properties.}
\label{tab:lang_proper}
\end{table*}

\begin{table}[t]
\centering\footnotesize
\setlength{\tabcolsep}{1.5pt}
\begin{tabular}{lcccccccccccccc}
\toprule
& ar & az & cs & de & el & en & hi & it & ja & ro & ru & sk & uk & vi \\
\midrule
ar & --- & \Agree{.58} & \Agree{.73} & \Agree{.77} & \Agree{.76} & \Agree{.79} & \Agree{.75} & \Agree{.78} & \Agree{.72} & \Agree{.75} & \Agree{.80} & \Agree{.73} & \Agree{.78} & \Agree{.76} \\
az & \Agree{.64} & --- & \Agree{.57} & \Agree{.58} & \Agree{.57} & \Agree{.58} & \Agree{.57} & \Agree{.59} & \Agree{.54} & \Agree{.58} & \Agree{.60} & \Agree{.59} & \Agree{.59} & \Agree{.56} \\
cs & \Agree{.76} & \Agree{.62} & --- & \Agree{.76} & \Agree{.72} & \Agree{.76} & \Agree{.70} & \Agree{.76} & \Agree{.68} & \Agree{.74} & \Agree{.78} & \Agree{.79} & \Agree{.76} & \Agree{.72} \\
de & \Agree{.82} & \Agree{.64} & \Agree{.79} & --- & \Agree{.77} & \Agree{.83} & \Agree{.75} & \Agree{.82} & \Agree{.73} & \Agree{.76} & \Agree{.83} & \Agree{.75} & \Agree{.80} & \Agree{.77} \\
el & \Agree{.82} & \Agree{.63} & \Agree{.76} & \Agree{.82} & --- & \Agree{.78} & \Agree{.76} & \Agree{.77} & \Agree{.71} & \Agree{.74} & \Agree{.79} & \Agree{.71} & \Agree{.77} & \Agree{.76} \\
en & \Agree{.84} & \Agree{.64} & \Agree{.79} & \Agree{.87} & \Agree{.85} & --- & \Agree{.76} & \Agree{.82} & \Agree{.74} & \Agree{.77} & \Agree{.84} & \Agree{.75} & \Agree{.80} & \Agree{.78} \\
hi & \Agree{.81} & \Agree{.64} & \Agree{.75} & \Agree{.80} & \Agree{.80} & \Agree{.83} & --- & \Agree{.74} & \Agree{.72} & \Agree{.71} & \Agree{.76} & \Agree{.70} & \Agree{.75} & \Agree{.73} \\
it & \Agree{.83} & \Agree{.64} & \Agree{.78} & \Agree{.85} & \Agree{.82} & \Agree{.86} & \Agree{.80} & --- & \Agree{.72} & \Agree{.77} & \Agree{.82} & \Agree{.75} & \Agree{.79} & \Agree{.77} \\
ja & \Agree{.78} & \Agree{.63} & \Agree{.71} & \Agree{.77} & \Agree{.77} & \Agree{.80} & \Agree{.78} & \Agree{.77} & --- & \Agree{.68} & \Agree{.74} & \Agree{.67} & \Agree{.72} & \Agree{.71} \\
ro & \Agree{.80} & \Agree{.64} & \Agree{.77} & \Agree{.81} & \Agree{.80} & \Agree{.82} & \Agree{.78} & \Agree{.82} & \Agree{.74} & --- & \Agree{.78} & \Agree{.74} & \Agree{.76} & \Agree{.73} \\
ru & \Agree{.84} & \Agree{.65} & \Agree{.80} & \Agree{.86} & \Agree{.84} & \Agree{.88} & \Agree{.82} & \Agree{.85} & \Agree{.80} & \Agree{.82} & --- & \Agree{.77} & \Agree{.84} & \Agree{.78} \\
sk & \Agree{.78} & \Agree{.64} & \Agree{.81} & \Agree{.80} & \Agree{.78} & \Agree{.80} & \Agree{.76} & \Agree{.79} & \Agree{.73} & \Agree{.79} & \Agree{.81} & --- & \Agree{.76} & \Agree{.71} \\
uk & \Agree{.83} & \Agree{.65} & \Agree{.79} & \Agree{.84} & \Agree{.83} & \Agree{.85} & \Agree{.81} & \Agree{.83} & \Agree{.78} & \Agree{.81} & \Agree{.88} & \Agree{.80} & --- & \Agree{.76} \\
vi & \Agree{.82} & \Agree{.62} & \Agree{.75} & \Agree{.82} & \Agree{.81} & \Agree{.84} & \Agree{.79} & \Agree{.82} & \Agree{.76} & \Agree{.79} & \Agree{.83} & \Agree{.77} & \Agree{.82} & --- \\
\midrule
Avg. & \Agree{.80} & \Agree{.63} & \Agree{.75} & \Agree{.80} & \Agree{.78} & \Agree{.80} & \Agree{.76} & \Agree{.78} & \Agree{.72} & \Agree{.75} & \Agree{.79} & \Agree{.73} & \Agree{.76} & \Agree{.73} \\
\bottomrule
\end{tabular}
\caption{Languages' agreement as Cohen's Kappa coefficient averaged over all models. Above the diagonal: \textit{2img} task. Below the diagonal \textit{2txt} task.}
\label{tab:cohen_kappa_lang_corr}
\end{table}

\begin{table}[t]
\centering\footnotesize
\setlength{\tabcolsep}{1.5pt}

\scalebox{0.86}{
\begin{tabular}{lcccccccccccc}
\toprule
& \rotatebox{90}{Gemma 3 12B} & \rotatebox{90}{Qwen 2.5 VL 7B} & \rotatebox{90}{Qwen 3 VL 8B} & \rotatebox{90}{Aya Vision 8B} & \rotatebox{90}{Jina VL} & \rotatebox{90}{NLLB-SigLIP Base} & \rotatebox{90}{NLLB-SigLIP Large} & \rotatebox{90}{MEXMA-SigLIP2} & \rotatebox{90}{SigLIP2 Base} & \rotatebox{90}{SigLIP2 Large} & \rotatebox{90}{SigLIP2 SO400M} & \rotatebox{90}{SigLIP2 Giant} \\
\midrule
Gemma 3 12B & --- & \Agree{.65} & \Agree{.70} & \Agree{.51} & \Agree{.09} & \Agree{.57} & \Agree{.61} & \Agree{.63} & \Agree{.58} & \Agree{.61} & \Agree{.62} & \Agree{.62} \\
Qwen 2.5 VL 7B & \Agree{.79} & --- & \Agree{.76} & \Agree{.46} & \Agree{.09} & \Agree{.60} & \Agree{.65} & \Agree{.67} & \Agree{.60} & \Agree{.62} & \Agree{.64} & \Agree{.63} \\
Qwen 3 VL 8B & \Agree{.82} & \Agree{.83} & --- & \Agree{.51} & \Agree{.08} & \Agree{.64} & \Agree{.69} & \Agree{.72} & \Agree{.64} & \Agree{.66} & \Agree{.68} & \Agree{.68} \\
Aya Vision 8B & \Agree{.71} & \Agree{.72} & \Agree{.74} & --- & \Agree{.08} & \Agree{.45} & \Agree{.48} & \Agree{.49} & \Agree{.45} & \Agree{.47} & \Agree{.48} & \Agree{.48} \\
Jina VL & \Agree{.60} & \Agree{.67} & \Agree{.62} & \Agree{.56} & --- & \Agree{.06} & \Agree{.06} & \Agree{.06} & \Agree{.06} & \Agree{.06} & \Agree{.06} & \Agree{.06} \\
\midrule
NLLB-SigLIP Base & \Agree{.63} & \Agree{.65} & \Agree{.66} & \Agree{.60} & \Agree{.52} & --- & \Agree{.70} & \Agree{.67} & \Agree{.55} & \Agree{.55} & \Agree{.57} & \Agree{.57} \\
NLLB-SigLIP Large & \Agree{.69} & \Agree{.71} & \Agree{.73} & \Agree{.65} & \Agree{.56} & \Agree{.71} & --- & \Agree{.72} & \Agree{.59} & \Agree{.60} & \Agree{.62} & \Agree{.62} \\
MEXMA-SigLIP2 & \Agree{.75} & \Agree{.76} & \Agree{.78} & \Agree{.68} & \Agree{.59} & \Agree{.68} & \Agree{.74} & --- & \Agree{.60} & \Agree{.61} & \Agree{.63} & \Agree{.63} \\
SigLIP2 Base & \Agree{.62} & \Agree{.61} & \Agree{.64} & \Agree{.57} & \Agree{.49} & \Agree{.51} & \Agree{.57} & \Agree{.60} & --- & \Agree{.74} & \Agree{.74} & \Agree{.74} \\
SigLIP2 Large & \Agree{.64} & \Agree{.63} & \Agree{.66} & \Agree{.58} & \Agree{.49} & \Agree{.53} & \Agree{.58} & \Agree{.61} & \Agree{.74} & --- & \Agree{.78} & \Agree{.78} \\
SigLIP2 SO400M & \Agree{.63} & \Agree{.62} & \Agree{.65} & \Agree{.58} & \Agree{.48} & \Agree{.52} & \Agree{.57} & \Agree{.61} & \Agree{.73} & \Agree{.77} & --- & \Agree{.80} \\
SigLIP2 Giant & \Agree{.61} & \Agree{.60} & \Agree{.63} & \Agree{.56} & \Agree{.47} & \Agree{.51} & \Agree{.56} & \Agree{.60} & \Agree{.70} & \Agree{.76} & \Agree{.78} & --- \\
\midrule
Avg. & \Agree{.68} & \Agree{.68} & \Agree{.69} & \Agree{.57} & \Agree{.36} & \Agree{.53} & \Agree{.57} & \Agree{.58} & \Agree{.57} & \Agree{.59} & \Agree{.60} & \Agree{.60} \\
\bottomrule
\end{tabular}
}

\caption{Models' agreement as Cohen's Kappa coefficient averaged over all languages. Above the diagonal: \textit{2img} task. Below the diagonal \textit{2txt} task.}
\label{tab:cohen_kappa_model_corr}
\end{table}

\section{Cross-lingual Experiments}
\label{sec:cross_lingual}

Unlike the previous set of experiments, where we tested the models monolingually across all languages, here we test them in a more challenging \textit{two-text} setup, in which each text is in a different language. This allows us to evaluate whether the models exhibit biases.
We design a set of cross-lingual experiments to measure whether models are biased toward better-resourced languages.

\subsection{Experimental Setup}

In these experiments, the model is given a single image together with two captions in different languages (one correct and one incorrect) and must identify which caption matches the image. We use the same evaluation protocol for dual encoders and decoder-only models as in Section~\ref{sec:multingual}. In addition to accuracy (equivalent to the \textit{2txt} task in Section~\ref{sec:multingual}), we report differences in per-language mistakes.

For two languages $x$ and $y$, the mistake $\mathrm{M}(x, y)$ measures how often the model makes an error by preferring $y$ over $x$ when $x$ is correct. If the model makes mistakes randomly, $\mathrm{M}(x, y) \approx \mathrm{M}(y, x)$, and the difference $\mathrm{M}_{\text{dif}}(x, y)$ is close to zero. 
A positive value for $\mathrm{M}_{\text{dif}}(x, y)$ therefore indicates a systematic bias in favor of language $x$, while a negative value indicates a bias towards the language $y$. 
More concretely:
\begin{eqnarray}
    \mathrm{M}(x, y) = \mathbb{0}_{c(I^\text{x}_\text{o}, \{T^\text{x}_\text{o}, T^\text{y}_\text{e}\})} & + & \mathbb{0}_{c(I^\text{x}_\text{e}, \{T^\text{y}_\text{o}, T^\text{x}_\text{e}\})} \\
    \mathrm{M}_{\text{dif}}(x, y) = \mathrm{M}(y, x) & - & \mathrm{M}(x, y) \label{eq:mdiff}
\end{eqnarray}
where $\mathbb{0}$ is the error indicator function.

\begin{table*}[t]
\centering\footnotesize
\setlength{\tabcolsep}{5.0pt}
\begin{tabular}{lcccccccccccccc@{\hskip 16pt}c}
\toprule
Lng. & ar & az & cs & de & el & en & hi & it & ja & ro & ru & sk & uk & vi & Avg. \\
\midrule
ar & --- & \FP{.17} & \FP{.04} & \FP{.01} & \FP{-.01} & \FP{-.05} & \FP{-.02} & \FP{.00} & \FP{.00} & \FP{.00} & \FP{-.02} & \FP{.05} & \FP{.01} & \FP{-.01} & \FP{.01} \\
az & \FP{-.17} & --- & \FP{-.16} & \FP{-.17} & \FP{-.21} & \FP{-.24} & \FP{-.21} & \FP{-.18} & \FP{-.18} & \FP{-.20} & \FP{-.19} & \FP{-.14} & \FP{-.17} & \FP{-.20} & \FP{-.19} \\
cs & \FP{-.04} & \FP{.16} & --- & \FP{-.05} & \FP{-.07} & \FP{-.13} & \FP{-.06} & \FP{-.05} & \FP{-.04} & \FP{-.05} & \FP{-.06} & \FP{.03} & \FP{-.04} & \FP{-.07} & \FP{-.04} \\
de & \FP{-.01} & \FP{.17} & \FP{.05} & --- & \FP{-.02} & \FP{-.09} & \FP{-.02} & \FP{.00} & \FP{.00} & \FP{.00} & \FP{-.02} & \FP{.07} & \FP{.01} & \FP{-.02} & \FP{.01} \\
el & \FP{.01} & \FP{.21} & \FP{.07} & \FP{.02} & --- & \FP{-.06} & \FP{-.01} & \FP{.02} & \FP{.03} & \FP{.02} & \FP{.00} & \FP{.09} & \FP{.03} & \FP{.00} & \FP{.03} \\
en & \FP{.05} & \FP{.24} & \FP{.13} & \FP{.09} & \FP{.06} & --- & \FP{.05} & \FP{.09} & \FP{.08} & \FP{.09} & \FP{.05} & \FP{.15} & \FP{.08} & \FP{.05} & \FP{.09} \\
hi & \FP{.02} & \FP{.21} & \FP{.06} & \FP{.02} & \FP{.01} & \FP{-.05} & --- & \FP{.02} & \FP{.02} & \FP{.01} & \FP{.00} & \FP{.08} & \FP{.03} & \FP{-.01} & \FP{.03} \\
it & \FP{.00} & \FP{.18} & \FP{.05} & \FP{.00} & \FP{-.02} & \FP{-.09} & \FP{-.02} & --- & \FP{.01} & \FP{.00} & \FP{-.02} & \FP{.07} & \FP{.01} & \FP{-.02} & \FP{.01} \\
ja & \FP{.00} & \FP{.18} & \FP{.04} & \FP{.00} & \FP{-.03} & \FP{-.08} & \FP{-.02} & \FP{-.01} & --- & \FP{-.01} & \FP{-.02} & \FP{.06} & \FP{.01} & \FP{-.02} & \FP{.01} \\
ro & \FP{.00} & \FP{.20} & \FP{.05} & \FP{.00} & \FP{-.02} & \FP{-.09} & \FP{-.01} & \FP{.00} & \FP{.01} & --- & \FP{-.02} & \FP{.08} & \FP{.01} & \FP{-.02} & \FP{.01} \\
ru & \FP{.02} & \FP{.19} & \FP{.06} & \FP{.02} & \FP{.00} & \FP{-.05} & \FP{.00} & \FP{.02} & \FP{.02} & \FP{.02} & --- & \FP{.09} & \FP{.03} & \FP{.00} & \FP{.03} \\
sk & \FP{-.05} & \FP{.14} & \FP{-.03} & \FP{-.07} & \FP{-.09} & \FP{-.15} & \FP{-.08} & \FP{-.07} & \FP{-.06} & \FP{-.08} & \FP{-.09} & --- & \FP{-.06} & \FP{-.09} & \FP{-.06} \\
uk & \FP{-.01} & \FP{.17} & \FP{.04} & \FP{-.01} & \FP{-.03} & \FP{-.08} & \FP{-.03} & \FP{-.01} & \FP{-.01} & \FP{-.01} & \FP{-.03} & \FP{.06} & --- & \FP{-.03} & \FP{.00} \\
vi & \FP{.01} & \FP{.20} & \FP{.07} & \FP{.02} & \FP{.00} & \FP{-.05} & \FP{.01} & \FP{.02} & \FP{.02} & \FP{.02} & \FP{.00} & \FP{.09} & \FP{.03} & --- & \FP{.03} \\
\bottomrule
\end{tabular}
\caption{Difference between language-specific mistakes, $\mathrm{M}_{\text{dif}}(x,y)$ as defined in Equation~(\ref{eq:mdiff}) averaged over all models. Languages on the vertical axis correspond to $x$, while languages on the horizontal axis correspond to $y$.}
\label{tab:bilingual_fp}
\end{table*}

\subsection{Results}

We report the detailed \textit{2txt} accuracy in this bilingual setup in Table~\ref{tab:bilingual_acc} in the appendix. Table~\ref{tab:bilingual_fp} presents the language-specific mistakes averaged across all models (above the diagonal; below are the negatives, making the matrix antisymmetric). The values in the table are $\mathrm{M}_{\text{dif}}(x, y)$ divided by the total number of samples. For completeness, we also compute the same table with the French prompts in Table~\ref{tab:lang_mistakes_fr_prompt}. We notice that the bias patterns are very similar to those in the English-prompt condition.

\paragraph{Overall trends.} Averaged across models, there is a clear bias toward English: models systematically prefer an incorrect English caption over a correct one in another language. Azerbaijani is the most disfavored language overall, with Czech and Slovak also showing a mild negative bias.

\paragraph{Decoder-only models.} The least biased model is Qwen~3 with close-to-zero language-specific mistakes in all languages, as displayed in Table~\ref{tab:bilingual_fp_llms}. At the opposite extreme, Gemma~3 shows a stronger preference for English than other models, despite its competitive \textit{strict accuracy}. Gemma disfavors languages using non-Latin scripts (Arabic, Japanese, Ukrainian, Greek, Hindi) as well as Azerbaijani. Aya and Qwen~2.5 show intermediate levels of English bias without strong script-related patterns. Jina~VL exhibits an unusual pattern: in addition to preferring English, it strongly favors Hindi while systematically disfavoring Slavic languages, regardless of script.

\paragraph{Dual encoder models.} Dual encoders show less English bias overall (Table~\ref{tab:bilingual_fp_encoders}) but differ substantially from one another. The first point is likely because their text towers are trained with a symmetric cross-lingual contrastive objective, whereas decoder-only LMs are pretrained on English-dominant text with vision attached afterward, which plausibly induces a stronger English preference at generation time. NLLB-SigLIP Base is a strong outlier: it is biased against European languages, suggesting that its training on 200 languages with NLLB representations may have introduced an unusual skew.
The Large variant is more balanced, though it retains a dispreference for Czech and Slovak.
MEXMA-SigLIP 2 and the SigLIP~2 family (especially in the base and giant versions) show little bias toward English but exhibit a consistent tendency to penalize Hindi and Japanese, similar to the accuracy drop for Japanese noted in Section~\ref{sec:multingual}.

\paragraph{Examples of English preferences.} We further analyze a subset of the mistakes: for every sample in the bilingual setup with English and any other target language, we count the number of times the model considers the wrong caption in English over the correct one in the target language. Figure~\ref{fig:mic_sample_2} contains the sample with one of the highest English preferences, $ep_{536}=161$, showing that the difference between \textit{small} and \textit{tiny} is not as clear in the target languages as it is in English. On the opposite pole, there are 16 samples for which the models display no such preference at all; one example is shown in Figure~\ref{fig:mic_sample_3}, which demonstrates that the models are capable of distinguishing between animals. A similar phenomenon can be observed in the headline picture (Figure~\ref{fig:mic_sample_1}), where $ep_{110}=13$. Additionally, for the image from the annotators' instruction (Figure~\ref{fig:screenshot}), $ep_{672}=64$ shows that the models find it difficult to distinguish between color palettes.

\paragraph{Random pairs ablation.} As a final ablation study, we aim to check the bilingual retrieval task with random pairs of captions. Table~\ref{tab:bilingual_acc_in_random_pairs} and Table~\ref{tab:lang_mistakes_random_pairs} display the accuracy and mistakes in this setup, showing that the task is almost trivial for our models, with high accuracies and few mistakes.

\section{Related Work}

\paragraph{Minimal pairs benchmarks.}
Minimal pairs, items differing in exactly one controlled dimension, are a standard diagnostic methodology in NLP.
The most notable examples include BLiMP \citep{warstadt-etal-2020-blimp-benchmark} and MultiBLiMP \citep{jumelet2025multiblimp10massivelymultilingual}, which are sentence pairs isolating specific contrasts in syntax, morphology, and semantics. HANS \citep{mccoy-etal-2019-right} probes whether natural language inference models exploit syntactic heuristics rather than genuine reasoning;
WinoBias \citep{zhao-etal-2018-gender} with Winograd-style coreference pairs measures gender bias in NLP systems.

The methodology has been adapted to vision-language models.
Winoground \citep{thrush2022winoground} tests whether models can match two images to two captions containing identical words in different orders.
ARO \citep{yuksekgonul2022when} scales this to 50,000 pairs, targeting relational and attributive understanding.
SugarCrepe \citep{hsieh-etal-2023-sugarcrepe} addresses artifacts in earlier benchmarks by using an LLM to generate fluent hard-negative captions.
All of these works focus exclusively on English.
InpaintCOCO \citep{rosch-etal-2024-enhancing} extends the paradigm by modifying both the caption and the image via inpainting, introducing genuine visual ambiguity.
We extend InpaintCOCO to 13 more languages.

\paragraph{Multilingual vision-language benchmarks.}

A survey by \citet{manea2025multilingual} identified 23 benchmarks for multilingual vision-language tasks. We refer the reader to that section for a broader overview.

Most existing multilingual benchmarks are either translation-based (XVNLI, xGQA, xFlickrCo; \citealt{bugliarello2022,nooralahzadeh2022}) or with region-specific imagery (MARVL, \citealt{liu-etal-2021-visually}; CVQA, \citealt{romero2024cvqaculturallydiversemultilingualvisual}).
MultiInpaintCOCO is translation-based, enables direct cross-lingual comparison, and is the only benchmark that combines caption-level and image-level minimal pairs, allowing retrieval errors to be attributed to either text or vision understanding.

\paragraph{Language bias.}

Language bias in NLP refers to the tendency of multilingual models to favor high-resource languages over lower-resource ones.
Early work showed that multilingual BERT encodes a structural bias toward English, with the grammatical properties of higher-resource languages bleeding into the representations of lower-resource ones \citep{papadimitriou-etal-2023-multilingual}.
Nowadays, this bias is particularly pronounced in retrieval settings: in multilingual retrieval-augmented generation, models systematically prefer high-resource language sources, especially English, even when lower-resource language documents are more relevant \citep{li-etal-2024-not,chen-etal-2024-linguisticnepotism}.

In vision-language models, related phenomena have been observed, such as models defaulting to English responses regardless of the query language when an image is present \citep{hinck-etal-2024-llava}, and degraded understanding of non-English text within images \citep{geigle-etal-2025-centurio}.

However, to the best of our knowledge, none of the previous works conducted a direct study of language bias in a \emph{multilingual vision-language retrieval} setup.

\section{Conclusions}

We introduced MultiInpaintCOCO, a multilingual minimal-pair benchmark for fine-grained vision-language retrieval, extending InpaintCOCO to 13 more typologically diverse languages via automatic translation followed by native-speaker post-editing. The benchmark combines caption-level and image-level minimal pairs, enabling retrieval errors to be attributed to either the textual or visual representation, a combination absent from prior multilingual VL benchmarks.

We evaluated five decoder-only and seven dual-encoder models and reported two main findings. First, performance drops substantially for lower-resource languages, with Azerbaijani—the least-resourced language in our study—consistently being the hardest across all models. The strongest predictor of this gap is the volume of pretraining data available for a language, as reflected by web corpus size, rather than typological proximity to English. Second, in cross-lingual retrieval, models exhibit a systematic preference for English captions over correct captions in other languages. This bias is not merely a consequence of lower monolingual accuracy; the tendency persists even in cases where a model would otherwise handle the non-English language well. The bias is strongest in Gemma 3 and weakest in Qwen 3, and it is present in both encoder and decoder architectures, though dual encoders tend to show less English preference overall.

These findings suggest that the English-centric nature of VL pretraining manifests in at least two separable ways: reduced representational capacity for lower-resource languages and an active preference for English during retrieval when English is one of the candidate options. %Addressing both will likely require targeted multilingual training, rather than simply including more languages in pretraining data.

\section*{Limitations}

MultiInpaintCOCO covers 14 typologically diverse languages across five language families. This remains a small fraction of the world's approximately 7,000 languages, and our findings may not generalize beyond the languages included.

Automatic translation followed by native-speaker post-editing may introduce a bias toward grammatically correct but unnaturally formal phrasing, as post-editors tend to preserve the structure of machine-translated output rather than produce fully idiomatic text. Moreover, we did not evaluate the translation quality (e.g., for naturalness, consistency, or difficulty) of our annotated captions across languages. Therefore, we assumed that the low performance is the direct consequence of the models' abilities.

Finally, the images in our dataset are drawn from COCO, which predominantly depicts scenes and objects from North America. This introduces a potential confound: strong performance on our benchmark does not necessarily imply that a model handles a given language well in its natural cultural context, since the visual content may be more familiar to models trained on Western-centric data, regardless of the language of the caption.

\section*{Acknowledgments}

We thank our translators: Konstantinos Diamantopulos, Turan Isgandarli, Nalin Kumar, Nam Luu, Yuka Miyake, Sabína Ságová, Tim Shchetilin, Ibrahim Qatato, and Gianluca Vico.

This research was supported by the Charles University project PRIMUS/23/SCI/023 and project CZ.02.01.01/00/23\_020/0008518 of the Ministry of Education, Youth and Sports of the Czech Republic.

We used several AI assistants for coding (Microsoft Copilot) and paper editing (Grammarly and Writefull).

\bibliography{custom,anthology.1,anthology.2}

\begin{thebibliography}{53}
\providecommand{\natexlab}[1]{#1}

\bibitem[{Aggarwal and Kale(2020)}]{xtd}
Pranav Aggarwal and Ajinkya Kale. 2020.
\newblock \href {https://arxiv.org/abs/2012.05107} {Towards zero-shot
  cross-lingual image retrieval}.
\newblock \emph{CoRR}, abs/2012.05107.

\bibitem[{Bai et~al.(2025{\natexlab{a}})Bai, Cai, Chen, Chen, Chen, Cheng,
  Deng, Ding, Gao, Ge et~al.}]{bai2025qwen3vltechnicalreport}
Shuai Bai, Yuxuan Cai, Ruizhe Chen, Keqin Chen, Xionghui Chen, Zesen Cheng,
  Lianghao Deng, Wei Ding, Chang Gao, Chunjiang Ge, and 1 others.
  2025{\natexlab{a}}.
\newblock \href {https://arxiv.org/abs/2511.21631} {Qwen3-vl technical report}.
\newblock \emph{Preprint}, arXiv:2511.21631.

\bibitem[{Bai et~al.(2025{\natexlab{b}})Bai, Chen, Liu, Wang, Ge, Song, Dang,
  Wang, Wang, Tang, Zhong, Zhu, Yang, Li, Wan, Wang, Ding, Fu, Xu, Ye, Zhang,
  Xie, Cheng, Zhang, Yang, Xu, and Lin}]{bai2025qwen25vltechnicalreport}
Shuai Bai, Keqin Chen, Xuejing Liu, Jialin Wang, Wenbin Ge, Sibo Song, Kai
  Dang, Peng Wang, Shijie Wang, Jun Tang, Humen Zhong, Yuanzhi Zhu, Mingkun
  Yang, Zhaohai Li, Jianqiang Wan, Pengfei Wang, Wei Ding, Zheren Fu, Yiheng
  Xu, and 8 others. 2025{\natexlab{b}}.
\newblock \href {https://arxiv.org/abs/2502.13923} {Qwen2.5-vl technical
  report}.
\newblock \emph{Preprint}, arXiv:2502.13923.

\bibitem[{Barrault et~al.(2018)Barrault, Bougares, Specia, Lala, Elliott, and
  Frank}]{barrault-etal-2018-findings}
Lo{\"i}c Barrault, Fethi Bougares, Lucia Specia, Chiraag Lala, Desmond Elliott,
  and Stella Frank. 2018.
\newblock \href {https://doi.org/10.18653/v1/W18-6402} {Findings of the third
  shared task on multimodal machine translation}.
\newblock In \emph{Proceedings of the Third Conference on Machine Translation:
  Shared Task Papers}, pages 304--323, Belgium, Brussels. Association for
  Computational Linguistics.

\bibitem[{Bugliarello et~al.(2022{\natexlab{a}})Bugliarello, Liu, Pfeiffer,
  Reddy, Elliott, Ponti, and Vuli{\'c}}]{pmlr-v162-bugliarello22a}
Emanuele Bugliarello, Fangyu Liu, Jonas Pfeiffer, Siva Reddy, Desmond Elliott,
  Edoardo~Maria Ponti, and Ivan Vuli{\'c}. 2022{\natexlab{a}}.
\newblock \href {https://proceedings.mlr.press/v162/bugliarello22a.html}
  {{IGLUE}: A benchmark for transfer learning across modalities, tasks, and
  languages}.
\newblock In \emph{Proceedings of the 39th International Conference on Machine
  Learning}, volume 162 of \emph{Proceedings of Machine Learning Research},
  pages 2370--2392. PMLR.

\bibitem[{Bugliarello et~al.(2022{\natexlab{b}})Bugliarello, Liu, Pfeiffer,
  Reddy, Elliott, Ponti, and Vulic}]{bugliarello2022}
Emanuele Bugliarello, Fangyu Liu, Jonas Pfeiffer, Siva Reddy, Desmond Elliott,
  Edoardo~Maria Ponti, and Ivan Vulic. 2022{\natexlab{b}}.
\newblock \href {https://proceedings.mlr.press/v162/bugliarello22a.html}
  {{IGLUE:} {A} {B}enchmark for {T}ransfer {L}earning across {M}odalities,
  {T}asks, and {L}anguages}.
\newblock In \emph{International Conference on Machine Learning, {ICML} 2022,
  17-23 July 2022}, volume 162 of \emph{Proceedings of Machine Learning
  Research}, pages 2370--2392, Baltimore, Maryland. {PMLR}.

\bibitem[{Changpinyo et~al.(2023)Changpinyo, Xue, Yarom, Thapliyal, Szpektor,
  Amelot, Chen, and Soricut}]{changpinyo-etal-2023-maxm}
Soravit Changpinyo, Linting Xue, Michal Yarom, Ashish Thapliyal, Idan Szpektor,
  Julien Amelot, Xi~Chen, and Radu Soricut. 2023.
\newblock \href {https://doi.org/10.18653/v1/2023.findings-emnlp.176}
  {{M}a{XM}: Towards multilingual visual question answering}.
\newblock In \emph{Findings of the Association for Computational Linguistics:
  EMNLP 2023}, pages 2667--2682, Singapore. Association for Computational
  Linguistics.

\bibitem[{Dash et~al.(2025)Dash, Nan, Dang, Ahmadian, Singh, Smith, Venkitesh,
  Shmyhlo, Aryabumi, Beller-Morales, Pekmez, Ozuzu, Richemond, Locatelli,
  Frosst, Blunsom, Gomez, Zhang, Fadaee, Govindassamy, Roy, Gallé, Ermis,
  Üstün, and Hooker}]{dash2025ayavisionadvancingfrontier}
Saurabh Dash, Yiyang Nan, John Dang, Arash Ahmadian, Shivalika Singh, Madeline
  Smith, Bharat Venkitesh, Vlad Shmyhlo, Viraat Aryabumi, Walter
  Beller-Morales, Jeremy Pekmez, Jason Ozuzu, Pierre Richemond, Acyr Locatelli,
  Nick Frosst, Phil Blunsom, Aidan Gomez, Ivan Zhang, Marzieh Fadaee, and 6
  others. 2025.
\newblock \href {https://arxiv.org/abs/2505.08751} {Aya vision: Advancing the
  frontier of multilingual multimodality}.
\newblock \emph{Preprint}, arXiv:2505.08751.

\bibitem[{Dumpala et~al.(2024)Dumpala, Jaiswal, Sastry, Milios, Oore, and
  Sajjad}]{hsieh-etal-2023-sugarcrepe}
Sri~Harsha Dumpala, Aman Jaiswal, Chandramouli~Shama Sastry, Evangelos~E.
  Milios, Sageev Oore, and Hassan Sajjad. 2024.
\newblock \href
  {http://papers.nips.cc/paper\_files/paper/2024/hash/200661bf8f4993b7828a45a2a90f2ecf-Abstract-Datasets\_and\_Benchmarks\_Track.html}
  {{SUGARCREPE++} dataset: Vision-language model sensitivity to semantic and
  lexical alterations}.
\newblock In \emph{Advances in Neural Information Processing Systems 38: Annual
  Conference on Neural Information Processing Systems 2024, NeurIPS 2024,
  Vancouver, BC, Canada, December 10 - 15, 2024}.

\bibitem[{Fu et~al.(2025)Fu, Bonnen, Guillory, and
  Darrell}]{fu2025hiddenplainsightvlms}
Stephanie Fu, Tyler Bonnen, Devin Guillory, and Trevor Darrell. 2025.
\newblock \href {https://arxiv.org/abs/2506.08008} {Hidden in plain sight: Vlms
  overlook their visual representations}.
\newblock \emph{Preprint}, arXiv:2506.08008.

\bibitem[{Futeral et~al.(2023)Futeral, Schmid, Laptev, Sagot, and
  Bawden}]{futeral-etal-2023-tackling}
Matthieu Futeral, Cordelia Schmid, Ivan Laptev, Beno{\^i}t Sagot, and Rachel
  Bawden. 2023.
\newblock \href {https://doi.org/10.18653/v1/2023.acl-long.295} {Tackling
  ambiguity with images: Improved multimodal machine translation and
  contrastive evaluation}.
\newblock In \emph{Proceedings of the 61st Annual Meeting of the Association
  for Computational Linguistics (Volume 1: Long Papers)}, pages 5394--5413,
  Toronto, Canada. Association for Computational Linguistics.

\bibitem[{Geigle et~al.(2025)Geigle, Schneider, Holtermann, Biemann, Timofte,
  Lauscher, and Glava{\v{s}}}]{geigle-etal-2025-centurio}
Gregor Geigle, Florian Schneider, Carolin Holtermann, Chris Biemann, Radu
  Timofte, Anne Lauscher, and Goran Glava{\v{s}}. 2025.
\newblock \href {https://doi.org/10.18653/v1/2025.acl-long.143} {Centurio: On
  drivers of multilingual ability of large vision-language model}.
\newblock In \emph{Proceedings of the 63rd Annual Meeting of the Association
  for Computational Linguistics (Volume 1: Long Papers)}, pages 2831--2881,
  Vienna, Austria. Association for Computational Linguistics.

\bibitem[{Hinck et~al.(2024)Hinck, Holtermann, Olson, Schneider, Yu,
  Bhiwandiwalla, Lauscher, Tseng, and Lal}]{hinck-etal-2024-llava}
Musashi Hinck, Carolin Holtermann, Matthew~Lyle Olson, Florian Schneider,
  Sungduk Yu, Anahita Bhiwandiwalla, Anne Lauscher, Shao-Yen Tseng, and Vasudev
  Lal. 2024.
\newblock \href {https://doi.org/10.18653/v1/2024.findings-emnlp.783} {Why do
  {LL}a{VA} vision-language models reply to images in {E}nglish?}
\newblock In \emph{Findings of the Association for Computational Linguistics:
  EMNLP 2024}, pages 13402--13421, Miami, Florida, USA. Association for
  Computational Linguistics.

\bibitem[{Janeiro et~al.(2024)Janeiro, Piwowarski, Gallinari, and
  Barrault}]{janeiro2024mexmatokenlevelobjectivesimprove}
João~Maria Janeiro, Benjamin Piwowarski, Patrick Gallinari, and Loïc
  Barrault. 2024.
\newblock \href {https://arxiv.org/abs/2409.12737} {Mexma: Token-level
  objectives improve sentence representations}.
\newblock \emph{Preprint}, arXiv:2409.12737.

\bibitem[{Joshi et~al.(2020)Joshi, Santy, Budhiraja, Bali, and
  Choudhury}]{joshi-etal-2020-state}
Pratik Joshi, Sebastin Santy, Amar Budhiraja, Kalika Bali, and Monojit
  Choudhury. 2020.
\newblock \href {https://doi.org/10.18653/v1/2020.acl-main.560} {The state and
  fate of linguistic diversity and inclusion in the {NLP} world}.
\newblock In \emph{Proceedings of the 58th Annual Meeting of the Association
  for Computational Linguistics}, pages 6282--6293, Online. Association for
  Computational Linguistics.

\bibitem[{Jumelet et~al.(2025)Jumelet, Weissweiler, and
  Bisazza}]{jumelet2025multiblimp10massivelymultilingual}
Jaap Jumelet, Leonie Weissweiler, and Arianna Bisazza. 2025.
\newblock \href {https://doi.org/10.48550/ARXIV.2504.02768} {Multiblimp 1.0:
  {A} massively multilingual benchmark of linguistic minimal pairs}.
\newblock \emph{CoRR}, abs/2504.02768.

\bibitem[{Khan et~al.(2025)Khan, Shipton, Anugraha, Duan, Hoang, Khiu,
  Do{\u{g}}ru{\"o}z, and Lee}]{khan-etal-2025-uriel}
Aditya Khan, Mason Shipton, David Anugraha, Kaiyao Duan, Phuong~H. Hoang, Eric
  Khiu, A.~Seza Do{\u{g}}ru{\"o}z, and En-Shiun~Annie Lee. 2025.
\newblock \href {https://aclanthology.org/2025.coling-main.463/} {{URIEL}+:
  Enhancing linguistic inclusion and usability in a typological and
  multilingual knowledge base}.
\newblock In \emph{Proceedings of the 31st International Conference on
  Computational Linguistics}, pages 6937--6952, Abu Dhabi, UAE. Association for
  Computational Linguistics.

\bibitem[{Ki et~al.(2025)Ki, Carpuat, McNamee, Khashabi, Yang, Lawrie, and
  Duh}]{chen-etal-2024-linguisticnepotism}
Dayeon Ki, Marine Carpuat, Paul McNamee, Daniel Khashabi, Eugene Yang, Dawn~J.
  Lawrie, and Kevin Duh. 2025.
\newblock \href {https://doi.org/10.48550/ARXIV.2509.13930} {Linguistic
  nepotism: Trading-off quality for language preference in multilingual {RAG}}.
\newblock \emph{CoRR}, abs/2509.13930.

\bibitem[{Kocmi et~al.(2025)Kocmi, Artemova, Avramidis, Bawden, Bojar, Dranch,
  Dvorkovich, Dukanov, Fishel, Freitag, Gowda, Grundkiewicz, Haddow, Karpinska,
  Koehn, Lakougna, Lundin, Monz, Murray, Nagata, Perrella, Proietti, Popel,
  Popovi{\'c}, Riley, Shmatova, Steingr{\'i}msson, Yankovskaya, and
  Zouhar}]{kocmi-etal-2025-findings}
Tom Kocmi, Ekaterina Artemova, Eleftherios Avramidis, Rachel Bawden,
  Ond{\v{r}}ej Bojar, Konstantin Dranch, Anton Dvorkovich, Sergey Dukanov, Mark
  Fishel, Markus Freitag, Thamme Gowda, Roman Grundkiewicz, Barry Haddow,
  Marzena Karpinska, Philipp Koehn, Howard Lakougna, Jessica Lundin, Christof
  Monz, Kenton Murray, and 10 others. 2025.
\newblock \href {https://doi.org/10.18653/v1/2025.wmt-1.22} {Findings of the
  {WMT}25 general machine translation shared task: Time to stop evaluating on
  easy test sets}.
\newblock In \emph{Proceedings of the Tenth Conference on Machine Translation},
  pages 355--413, Suzhou, China. Association for Computational Linguistics.

\bibitem[{Koukounas et~al.(2026)Koukounas, Mastrapas, Hönicke, Eslami,
  Roncari, Martens, and Xiao}]{koukounas2026jinavlmsmallmultilingualvision}
Andreas Koukounas, Georgios Mastrapas, Florian Hönicke, Sedigheh Eslami,
  Guillaume Roncari, Scott Martens, and Han Xiao. 2026.
\newblock \href {https://arxiv.org/abs/2512.04032} {jina-vlm: Small
  multilingual vision language model}.
\newblock \emph{Preprint}, arXiv:2512.04032.

\bibitem[{Lin et~al.(2014)Lin, Maire, Belongie, Hays, Perona, Ramanan,
  Doll{\'{a}}r, and Zitnick}]{lin2014microsoft}
Tsung{-}Yi Lin, Michael Maire, Serge~J. Belongie, James Hays, Pietro Perona,
  Deva Ramanan, Piotr Doll{\'{a}}r, and C.~Lawrence Zitnick. 2014.
\newblock \href {https://doi.org/10.1007/978-3-319-10602-1\_48} {Microsoft
  {COCO:} common objects in context}.
\newblock In \emph{Computer Vision - {ECCV} 2014 - 13th European Conference,
  Zurich, Switzerland, September 6-12, 2014, Proceedings, Part {V}}, volume
  8693 of \emph{Lecture Notes in Computer Science}, pages 740--755. Springer.

\bibitem[{Liu et~al.(2021)Liu, Bugliarello, Ponti, Reddy, Collier, and
  Elliott}]{liu-etal-2021-visually}
Fangyu Liu, Emanuele Bugliarello, Edoardo~Maria Ponti, Siva Reddy, Nigel
  Collier, and Desmond Elliott. 2021.
\newblock \href {https://doi.org/10.18653/v1/2021.emnlp-main.818} {Visually
  grounded reasoning across languages and cultures}.
\newblock In \emph{Proceedings of the 2021 Conference on Empirical Methods in
  Natural Language Processing}, pages 10467--10485, Online and Punta Cana,
  Dominican Republic. Association for Computational Linguistics.

\bibitem[{Manea and Libovick{\'{y}}(2025)}]{manea2025multilingual}
Andrei{-}Alexandru Manea and Jindrich Libovick{\'{y}}. 2025.
\newblock \href {https://doi.org/10.48550/ARXIV.2509.22123} {Multilingual
  vision-language models, {A} survey}.
\newblock \emph{CoRR}, abs/2509.22123.

\bibitem[{McCoy et~al.(2019)McCoy, Pavlick, and Linzen}]{mccoy-etal-2019-right}
R.~Thomas McCoy, Ellie Pavlick, and Tal Linzen. 2019.
\newblock \href {https://doi.org/10.18653/v1/P19-1334} {Right for the wrong
  reasons: Diagnosing syntactic heuristics in natural language inference}.
\newblock In \emph{Proceedings of the 57th Annual Meeting of the Association
  for Computational Linguistics}, pages 3428--3448, Florence, Italy.
  Association for Computational Linguistics.

\bibitem[{Nooralahzadeh and Sennrich(2023)}]{nooralahzadeh2022}
Farhad Nooralahzadeh and Rico Sennrich. 2023.
\newblock \href {https://doi.org/10.1609/AAAI.V37I11.26574} {{I}mproving the
  {C}ross-lingual {G}eneralisation in {V}isual {Q}uestion {A}nswering}.
\newblock In \emph{Thirty-Seventh {AAAI} Conference on Artificial Intelligence,
  {AAAI} 2023, Thirty-Fifth Conference on Innovative Applications of Artificial
  Intelligence, {IAAI} 2023, Thirteenth Symposium on Educational Advances in
  Artificial Intelligence, {EAAI} 2023}, pages 13419--13427, Washington, DC,
  USA. {AAAI} Press.

\bibitem[{Papadimitriou et~al.(2023)Papadimitriou, Lopez, and
  Jurafsky}]{papadimitriou-etal-2023-multilingual}
Isabel Papadimitriou, Kezia Lopez, and Dan Jurafsky. 2023.
\newblock \href {https://doi.org/10.18653/v1/2023.findings-eacl.89}
  {Multilingual {BERT} has an accent: Evaluating {E}nglish influences on
  fluency in multilingual models}.
\newblock In \emph{Findings of the Association for Computational Linguistics:
  EACL 2023}, pages 1194--1200, Dubrovnik, Croatia. Association for
  Computational Linguistics.

\bibitem[{Penedo et~al.(2025)Penedo, Kydl{\'\i}{\v{c}}ek, Sabol{\v{c}}ec,
  Messmer, Foroutan, Kargaran, Raffel, Jaggi, Werra, and
  Wolf}]{penedo2025fineweb}
Guilherme Penedo, Hynek Kydl{\'\i}{\v{c}}ek, Vinko Sabol{\v{c}}ec, Bettina
  Messmer, Negar Foroutan, Amir~Hossein Kargaran, Colin Raffel, Martin Jaggi,
  Leandro~Von Werra, and Thomas Wolf. 2025.
\newblock \href {https://openreview.net/forum?id=jnRBe6zatP} {Fineweb2: One
  pipeline to scale them all {\textemdash} adapting pre-training data
  processing to every language}.
\newblock In \emph{Second Conference on Language Modeling}.

\bibitem[{Pfeiffer et~al.(2022)Pfeiffer, Geigle, Kamath, Steitz, Roth,
  Vuli{\'c}, and Gurevych}]{pfeiffer-etal-2022-xgqa}
Jonas Pfeiffer, Gregor Geigle, Aishwarya Kamath, Jan-Martin~O. Steitz, Stefan
  Roth, Ivan Vuli{\'c}, and Iryna Gurevych. 2022.
\newblock \href {https://doi.org/10.18653/v1/2022.findings-acl.196} {x{GQA}:
  Cross-lingual visual question answering}.
\newblock In \emph{Findings of the Association for Computational Linguistics:
  ACL 2022}, pages 2497--2511, Dublin, Ireland. Association for Computational
  Linguistics.

\bibitem[{Ploeger et~al.(2024)Ploeger, Poelman, de~Lhoneux, and
  Bjerva}]{ploeger-etal-2024-typological}
Esther Ploeger, Wessel Poelman, Miryam de~Lhoneux, and Johannes Bjerva. 2024.
\newblock \href {https://doi.org/10.18653/v1/2024.emnlp-main.326} {What is
  ``typological diversity'' in {NLP}?}
\newblock In \emph{Proceedings of the 2024 Conference on Empirical Methods in
  Natural Language Processing}, pages 5681--5700, Miami, Florida, USA.
  Association for Computational Linguistics.

\bibitem[{Radford et~al.(2021)Radford, Kim, Hallacy, Ramesh, Goh, Agarwal,
  Sastry, Askell, Mishkin, Clark, Krueger, and
  Sutskever}]{pmlr-v139-radford21a}
Alec Radford, Jong~Wook Kim, Chris Hallacy, Aditya Ramesh, Gabriel Goh,
  Sandhini Agarwal, Girish Sastry, Amanda Askell, Pamela Mishkin, Jack Clark,
  Gretchen Krueger, and Ilya Sutskever. 2021.
\newblock \href {https://proceedings.mlr.press/v139/radford21a.html} {Learning
  transferable visual models from natural language supervision}.
\newblock In \emph{Proceedings of the 38th International Conference on Machine
  Learning}, volume 139 of \emph{Proceedings of Machine Learning Research},
  pages 8748--8763. PMLR.

\bibitem[{Rombach et~al.(2022)Rombach, Blattmann, Lorenz, Esser, and
  Ommer}]{rombach2022high}
Robin Rombach, Andreas Blattmann, Dominik Lorenz, Patrick Esser, and
  Bj{\"{o}}rn Ommer. 2022.
\newblock \href {https://doi.org/10.1109/CVPR52688.2022.01042} {High-resolution
  image synthesis with latent diffusion models}.
\newblock In \emph{{IEEE/CVF} Conference on Computer Vision and Pattern
  Recognition, {CVPR} 2022, New Orleans, LA, USA, June 18-24, 2022}, pages
  10674--10685. {IEEE}.

\bibitem[{Romero et~al.(2024)Romero, Lyu, Wibowo, G{\'{o}}ngora, Mandal,
  Purkayastha, Ortiz{-}Barajas, Villa{-}Cueva, Baek, Jeong, Hamed, Yong, Lim,
  Silva, Dunstan, Jouitteau, Meur, Nwatu, Batnasan, Otgonbold, Gochoo, Ivetta,
  Benotti, Alemany, Maina, Geng, Torrent, Belcavello, Viridiano, Cruz, Velasco,
  Ignat, Burzo, Whitehouse, Abzaliev, Clifford, Caulfield, Lynn, Palacios,
  Araujo, Kementchedjhieva, Mihaylov, Azime, Ademtew, Balcha, Etori, Adelani,
  Mihalcea, Tonja, Cabrera, Vallejo, Lovenia, Zhang, Estecha{-}Garitagoitia,
  Rodr{\'{\i}}guez{-}Cantelar, Ehsan, Chevi, Adilazuarda, Diandaru,
  Cahyawijaya, Koto, Kuribayashi, Song, Khandavally, Jayakumar, Dabre, Imam,
  Nagasinghe, Dragonetti, D'Haro, Niyomugisha, Gala, Chitale, Farooqui,
  Solorio, and Aji}]{romero2024cvqaculturallydiversemultilingualvisual}
David Romero, Chenyang Lyu, Haryo~Akbarianto Wibowo, Santiago G{\'{o}}ngora,
  Aishik Mandal, Sukannya Purkayastha, Jes{\'{u}}s{-}Germ{\'{a}}n
  Ortiz{-}Barajas, Emilio Villa{-}Cueva, Jinheon Baek, Soyeong Jeong, Injy
  Hamed, Zheng~Xin Yong, Zheng~Wei Lim, Paula~M{\'{o}}nica Silva, Jocelyn
  Dunstan, M{\'{e}}lanie Jouitteau, David~Le Meur, Joan Nwatu, Ganzorig
  Batnasan, and 57 others. 2024.
\newblock \href
  {http://papers.nips.cc/paper\_files/paper/2024/hash/1568882ba1a50316e87852542523739c-Abstract-Datasets\_and\_Benchmarks\_Track.html}
  {{CVQA:} {C}ulturally-diverse {M}ultilingual {V}isual {Q}uestion {A}nswering
  {B}enchmark}.
\newblock In \emph{Advances in Neural Information Processing Systems 38: Annual
  Conference on Neural Information Processing Systems 2024, NeurIPS 2024,
  Vancouver, BC}, Canada.

\bibitem[{R{\"o}sch et~al.(2024)R{\"o}sch, Oswald, Geierhos, and
  Libovick{\'y}}]{rosch-etal-2024-enhancing}
Philipp~J. R{\"o}sch, Norbert Oswald, Michaela Geierhos, and Jind{\v{r}}ich
  Libovick{\'y}. 2024.
\newblock \href {https://doi.org/10.18653/v1/2024.alvr-1.9} {Enhancing
  conceptual understanding in multimodal contrastive learning through hard
  negative samples}.
\newblock In \emph{Proceedings of the 3rd Workshop on Advances in Language and
  Vision Research (ALVR)}, pages 102--115, Bangkok, Thailand. Association for
  Computational Linguistics.

\bibitem[{Schneider and Sitaram(2024)}]{schneider-sitaram-2024-m5}
Florian Schneider and Sunayana Sitaram. 2024.
\newblock \href {https://doi.org/10.18653/v1/2024.findings-emnlp.250} {M5 {--}
  a diverse benchmark to assess the performance of large multimodal models
  across multilingual and multicultural vision-language tasks}.
\newblock In \emph{Findings of the Association for Computational Linguistics:
  EMNLP 2024}, pages 4309--4345, Miami, Florida, USA. Association for
  Computational Linguistics.

\bibitem[{Tang et~al.(2025)Tang, Liu, Ye, Lu, Wei, Lin, Li, Mahmood, Feng,
  Zhao, He, Lu, Wang, Liu, Liu, Bai, and
  Huang}]{tang2025mtvqabenchmarkingmultilingualtextcentric}
Jingqun Tang, Qi~Liu, Yongjie Ye, Jinghui Lu, Shu Wei, Chunhui Lin, Wanqing Li,
  Mohamad Fitri Faiz~Bin Mahmood, Hao Feng, Zhen Zhao, Yangfan He, Kuan Lu,
  Yanjie Wang, Yuliang Liu, Hao Liu, Xiang Bai, and Can Huang. 2025.
\newblock \href {https://arxiv.org/abs/2405.11985} {Mtvqa: Benchmarking
  multilingual text-centric visual question answering}.
\newblock \emph{Preprint}, arXiv:2405.11985.

\bibitem[{Team et~al.(2025)Team, Kamath, Ferret, Pathak, Vieillard, Merhej,
  Perrin, Matejovicova, Ramé et~al.}]{gemmateam2025gemma3technicalreport}
Gemma Team, Aishwarya Kamath, Johan Ferret, Shreya Pathak, Nino Vieillard,
  Ramona Merhej, Sarah Perrin, Tatiana Matejovicova, Alexandre Ramé, and 1
  others. 2025.
\newblock \href {https://arxiv.org/abs/2503.19786} {{G}emma 3 {T}echnical
  {R}eport}.
\newblock \emph{Preprint}, arXiv:2503.19786.

\bibitem[{Team et~al.(2022)Team, Costa-jussà, Cross, Çelebi, Elbayad,
  Heafield, Heffernan, Kalbassi, Lam, Licht, Maillard, Sun, Wang, Wenzek,
  Youngblood, Akula, Barrault, Gonzalez, Hansanti, Hoffman, Jarrett, Sadagopan,
  Rowe, Spruit, Tran, Andrews, Ayan, Bhosale, Edunov, Fan, Gao, Goswami,
  Guzmán, Koehn, Mourachko, Ropers, Saleem, Schwenk, and
  Wang}]{nllbteam2022languageleftbehindscaling}
NLLB Team, Marta~R. Costa-jussà, James Cross, Onur Çelebi, Maha Elbayad,
  Kenneth Heafield, Kevin Heffernan, Elahe Kalbassi, Janice Lam, Daniel Licht,
  Jean Maillard, Anna Sun, Skyler Wang, Guillaume Wenzek, Al~Youngblood, Bapi
  Akula, Loic Barrault, Gabriel~Mejia Gonzalez, Prangthip Hansanti, and 20
  others. 2022.
\newblock \href {https://arxiv.org/abs/2207.04672} {No language left behind:
  Scaling human-centered machine translation}.
\newblock \emph{Preprint}, arXiv:2207.04672.

\bibitem[{Thapliyal et~al.(2022)Thapliyal, Pont~Tuset, Chen, and
  Soricut}]{thapliyal-etal-2022-crossmodal}
Ashish~V. Thapliyal, Jordi Pont~Tuset, Xi~Chen, and Radu Soricut. 2022.
\newblock \href {https://doi.org/10.18653/v1/2022.emnlp-main.45}
  {Crossmodal-3600: A massively multilingual multimodal evaluation dataset}.
\newblock In \emph{Proceedings of the 2022 Conference on Empirical Methods in
  Natural Language Processing}, pages 715--729, Abu Dhabi, United Arab
  Emirates. Association for Computational Linguistics.

\bibitem[{Thrush et~al.(2022)Thrush, Jiang, Bartolo, Singh, Williams, Kiela,
  and Ross}]{thrush2022winoground}
Tristan Thrush, Ryan Jiang, Max Bartolo, Amanpreet Singh, Adina Williams, Douwe
  Kiela, and Candace Ross. 2022.
\newblock Winoground: Probing vision and language models for visio-linguistic
  compositionality.
\newblock In \emph{Proceedings of the IEEE/CVF Conference on Computer Vision
  and Pattern Recognition}, pages 5238--5248.

\bibitem[{Tian et~al.(2025)Tian, Zou, Yang, and Zhang}]{tian2025identifying}
Xinyu Tian, Shu Zou, Zhaoyuan Yang, and Jing Zhang. 2025.
\newblock \href {https://doi.org/10.1109/CVPR52734.2025.00991} {Identifying and
  mitigating position bias of multi-image vision-language models}.
\newblock In \emph{{IEEE/CVF} Conference on Computer Vision and Pattern
  Recognition, {CVPR} 2025, Nashville, TN, USA, June 11-15, 2025}, pages
  10599--10609. Computer Vision Foundation / {IEEE}.

\bibitem[{Tschannen et~al.(2025)Tschannen, Gritsenko, Wang, Naeem,
  Alabdulmohsin, Parthasarathy, Evans, Beyer, Xia, Mustafa, Hénaff, Harmsen,
  Steiner, and Zhai}]{tschannen2025siglip2multilingualvisionlanguage}
Michael Tschannen, Alexey Gritsenko, Xiao Wang, Muhammad~Ferjad Naeem, Ibrahim
  Alabdulmohsin, Nikhil Parthasarathy, Talfan Evans, Lucas Beyer, Ye~Xia, Basil
  Mustafa, Olivier Hénaff, Jeremiah Harmsen, Andreas Steiner, and Xiaohua
  Zhai. 2025.
\newblock \href {https://arxiv.org/abs/2502.14786} {Siglip 2: Multilingual
  vision-language encoders with improved semantic understanding, localization,
  and dense features}.
\newblock \emph{Preprint}, arXiv:2502.14786.

\bibitem[{Vayani et~al.(2025)Vayani, Dissanayake, Watawana, Ahsan, Sasikumar,
  Thawakar, Ademtew, Hmaiti, Kumar, Kuckreja, Maslych, Ghallabi, Mihaylov, Qin,
  Shaker, Zhang, Ihsani, Esplana, Gokani, Mirkin, Singh, Srivastava, Hamerlik,
  Izzati, Maani, Cavada, Chim, Gupta, Manjunath, Zhumakhanova, Rabevohitra,
  Amirudin, Ridzuan, Kareem, More, Li, Shakya, Saad, Ghasemaghaei, Djanibekov,
  Azizov, Jankovic, Bhatia, Cabrera, Obando-Ceron, Otieno, Farestam, Rabbani,
  Baliah, Sanjeev, Shtanchaev, Fatima, Nguyen, Kareem, Aremu, Xavier, Bhatkal,
  Toyin, Chadha, Cholakkal, Anwer, Felsberg, Laaksonen, Solorio, Choudhury,
  Laptev, Shah, Khan, and Khan}]{vayani2025languagesmatterevaluatinglmms}
Ashmal Vayani, Dinura Dissanayake, Hasindri Watawana, Noor Ahsan, Nevasini
  Sasikumar, Omkar Thawakar, Henok~Biadglign Ademtew, Yahya Hmaiti, Amandeep
  Kumar, Kartik Kuckreja, Mykola Maslych, Wafa~Al Ghallabi, Mihail Mihaylov,
  Chao Qin, Abdelrahman~M Shaker, Mike Zhang, Mahardika~Krisna Ihsani, Amiel
  Esplana, Monil Gokani, and 50 others. 2025.
\newblock \href {https://arxiv.org/abs/2411.16508} {All languages matter:
  Evaluating lmms on culturally diverse 100 languages}.
\newblock \emph{Preprint}, arXiv:2411.16508.

\bibitem[{Visheratin(2023)}]{visheratin2023nllbcliptrainperformant}
Alexander Visheratin. 2023.
\newblock \href {https://arxiv.org/abs/2309.01859} {Nllb-clip -- train
  performant multilingual image retrieval model on a budget}.
\newblock \emph{Preprint}, arXiv:2309.01859.

\bibitem[{Vrandecic and Kr{\"{o}}tzsch(2014)}]{vrandecic2014wikidata}
Denny Vrandecic and Markus Kr{\"{o}}tzsch. 2014.
\newblock \href {https://doi.org/10.1145/2629489} {Wikidata: a free
  collaborative knowledgebase}.
\newblock \emph{Commun. {ACM}}, 57(10):78--85.

\bibitem[{Wang et~al.(2025)Wang, Huang, Yang, Xie, and
  Kawahara}]{wang2025travelinglanguagesbenchmarkingcrosslingual}
Hao Wang, Pinzhi Huang, Jihan Yang, Saining Xie, and Daisuke Kawahara. 2025.
\newblock \href {https://arxiv.org/abs/2505.15075} {Traveling across languages:
  Benchmarking cross-lingual consistency in multimodal llms}.
\newblock \emph{Preprint}, arXiv:2505.15075.

\bibitem[{Wardle and Susnjak(2024)}]{wardle2024imagetextfirstoptimising}
Grant Wardle and Teo Susnjak. 2024.
\newblock \href {https://arxiv.org/abs/2410.03062} {Image first or text first?
  optimising the sequencing of modalities in large language model prompting and
  reasoning tasks}.
\newblock \emph{Preprint}, arXiv:2410.03062.

\bibitem[{Warstadt et~al.(2020)Warstadt, Parrish, Liu, Mohananey, Peng, Wang,
  and Bowman}]{warstadt-etal-2020-blimp-benchmark}
Alex Warstadt, Alicia Parrish, Haokun Liu, Anhad Mohananey, Wei Peng, Sheng-Fu
  Wang, and Samuel~R. Bowman. 2020.
\newblock \href {https://doi.org/10.1162/tacl_a_00321} {{BL}i{MP}: The
  benchmark of linguistic minimal pairs for {E}nglish}.
\newblock \emph{Transactions of the Association for Computational Linguistics},
  8:377--392.

\bibitem[{Wu et~al.(2024)Wu, Tang, Yang, Wang, Jia, Yu, Yao, and
  Su}]{li-etal-2024-not}
Suhang Wu, Jialong Tang, Baosong Yang, Ante Wang, Kaidi Jia, Jiawei Yu, Junfeng
  Yao, and Jinsong Su. 2024.
\newblock \href {https://doi.org/10.48550/ARXIV.2410.21970} {Not all languages
  are equal: Insights into multilingual retrieval-augmented generation}.
\newblock \emph{CoRR}, abs/2410.21970.

\bibitem[{Yue et~al.(2025)Yue, Song, Asai, Kim, de~Dieu~Nyandwi, Khanuja,
  Kantharuban, Sutawika, Ramamoorthy, and
  Neubig}]{yue2025pangeafullyopenmultilingual}
Xiang Yue, Yueqi Song, Akari Asai, Seungone Kim, Jean de~Dieu~Nyandwi, Simran
  Khanuja, Anjali Kantharuban, Lintang Sutawika, Sathyanarayanan Ramamoorthy,
  and Graham Neubig. 2025.
\newblock \href {https://arxiv.org/abs/2410.16153} {Pangea: A fully open
  multilingual multimodal llm for 39 languages}.
\newblock \emph{Preprint}, arXiv:2410.16153.

\bibitem[{Yuksekgonul et~al.(2023)Yuksekgonul, Bianchi, Kalluri, Jurafsky, and
  Zou}]{yuksekgonul2022when}
Mert Yuksekgonul, Federico Bianchi, Pratyusha Kalluri, Dan Jurafsky, and James
  Zou. 2023.
\newblock \href {https://openreview.net/forum?id=KRLUvxh8uaX} {When and why
  vision-language models behave like bags-of-words, and what to do about it?}
\newblock In \emph{International Conference on Learning Representations}.

\bibitem[{Zhai et~al.(2023)Zhai, Mustafa, Kolesnikov, and
  Beyer}]{zhai2023sigmoidlosslanguageimage}
Xiaohua Zhai, Basil Mustafa, Alexander Kolesnikov, and Lucas Beyer. 2023.
\newblock \href {https://arxiv.org/abs/2303.15343} {Sigmoid loss for language
  image pre-training}.
\newblock \emph{Preprint}, arXiv:2303.15343.

\bibitem[{Zhao et~al.(2018)Zhao, Wang, Yatskar, Ordonez, and
  Chang}]{zhao-etal-2018-gender}
Jieyu Zhao, Tianlu Wang, Mark Yatskar, Vicente Ordonez, and Kai-Wei Chang.
  2018.
\newblock \href {https://doi.org/10.18653/v1/N18-2003} {Gender bias in
  coreference resolution: Evaluation and debiasing methods}.
\newblock In \emph{Proceedings of the 2018 Conference of the North {A}merican
  Chapter of the Association for Computational Linguistics: Human Language
  Technologies, Volume 2 (Short Papers)}, pages 15--20, New Orleans, Louisiana.
  Association for Computational Linguistics.

\bibitem[{Zheng et~al.(2025)Zheng, Li, Du, Qu, and Song}]{zheng-etal-2025-shy}
Mao Zheng, Zheng Li, Yang Du, Bingxin Qu, and Mingyang Song. 2025.
\newblock \href {https://doi.org/10.18653/v1/2025.wmt-1.36} {Shy-hunyuan-{MT}
  at {WMT}25 general machine translation shared task}.
\newblock In \emph{Proceedings of the Tenth Conference on Machine Translation},
  pages 607--613, Suzhou, China. Association for Computational Linguistics.

\end{thebibliography}

\appendix

\section{Annotation Process}\label{ap:annotation}

The filtered dataset was translated into 13 additional languages. The filtering and corrections of the original dataset were carried out by researchers who are authors of this paper.

Translation into Czech, German, and Romanian was done by researchers. Translation into Russian was done by a language teacher of Russian. Translation into Arabic, Azerbaijani, Greek, Hindi, Italian, Japanese, Slovak, Ukrainian, and Vietnamese was done by university students. The students were recruited in classes. The students were paid an hourly rate that corresponds to 223\% of the minimum wage and 125\% of the median wage in the country where the translation was conducted.

\begin{figure}[t]

\includegraphics[width=\columnwidth]{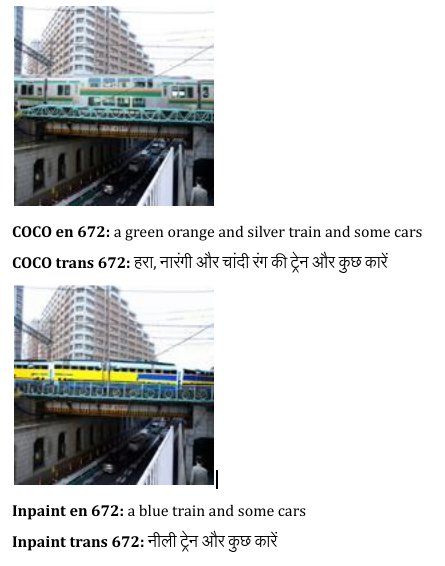}

\caption{A screenshot from the post-editing document for Hindi.}\label{fig:screenshot}

\end{figure}

We used Google Docs as the annotation interface (see Figure~\ref{fig:screenshot}). The annotators were presented with the images, original English captions, and the machine translation and were instructed to edit the translation without changing the formatting. We asked the annotators to keep a separate list of issues that they encountered in the dataset. Self-reported postediting times (biased by telling annotators it should take 6 hours) and the preferred MT system are reported in Table~\ref{tab:annotation}.

\begin{table}[t]
\centering\footnotesize
\setlength{\tabcolsep}{3pt}

\begin{tabular}{l ccccccccccccc}
\toprule
      & ar & az & cs & de & el & hi & it & ja & ro & ru & sk & uk & vi \\ \midrule
MT    & 4  &  4 &  4 &  4 &  4 &  4 &  G &  H &  4 &  4 &  4 &  4 &  H \\
Hours & 10 &  6 & 10 &  6 &  6 &  6 &  6 & 12 & 10 &  6 &  6 &  6 &  6 \\
\bottomrule
\end{tabular}

\caption{Machine translation system used for post-editing (G: Google Translate, H: Hunyuan, 4: GPT-4) and self-reported time spent by post-editing.}\label{tab:annotation}

\end{table}

\section{Hyperparameters}\label{app:hyperparams}

For the decoder-only models, we use the following instruction prompts:

English:
\begin{itemize}
    \item[\textit{2img}] Based on two images, which follow the caption [$C$] ? Answer with 'left' or 'right' and nothing else.
    \item[\textit{2txt}] Based on the 2 captions: [NEWLINE] (left) [$C_{1}$] [NEWLINE] (right) [$C_{2}$] [NEWLINE] Which one follows the image? Answer with 'left' or 'right' and nothing else.
\end{itemize}

French:
\begin{itemize}
    \item[\textit{2img}] À partir des deux images, qui suivent la description [$C$] ? Répondez par 'gauche' ou 'droite' et rien d'autre.
    \item[\textit{2txt}] À partir des deux descriptions suivantes: [NEWLINE] (gauche) [$C_{1}$] [NEWLINE] (droite) [$C_{2}$] [NEWLINE] Laquelle correspond à l'image? Répondez par 'gauche' ou 'droite', et rien d'autre.
\end{itemize}

In the \textit{2txt} task, the newline was necessary for the captions written in a right-to-left system (e.g. Arabic). Otherwise, the keywords left and right were confusing to the LLMs, causing them to produce a random input.

For encoder models, we note one small limitation of the SigLIP 2 text encoder: the maximum input length is 64 tokens. This impacts a maximum of 4 input captions, the highest number being in Greek, with the tokenizer of the base version.

% and the model is case-sensitive, so it is more familiar with lowercase input. \am{should I rather say that it has better results lowercase than the natural input case? It was more visible in German.}

%\section{Additional Model Details}\label{app:model_details}

\begin{table*}
\centering
\small
\begin{tabular}{ll c rllrccc}
\toprule
\multicolumn{2}{l}{\multirow{2}{*}{Language}} &
\multirow{2}{*}{\begin{minipage}{25pt}\centering Resource tier\end{minipage}} &
\multirow{2}{*}{\begin{minipage}{30pt}\centering Native speakers (M)\end{minipage}} &
\multirow{2}{*}{\begin{minipage}{50pt}Morphology\end{minipage}} &
\multirow{2}{*}{\begin{minipage}{50pt}Family\end{minipage}} &
\multirow{2}{*}{\begin{minipage}{30pt}\centering FineWeb (GB)\end{minipage}} &
\multirow{2}{*}{\begin{minipage}{30pt}\centering Wiki articles (M)\end{minipage}}  &
\multicolumn{2}{c}{WALS sim. w/} \\

\cmidrule(lr){9-10}

& & & & & & & & en & zh \\[.2ex]

\midrule
English            & en & 5 & 380  & Isolating & Indo-Europ.  &   2 $\cdot$ 10\textsuperscript{8} & 6.90 & ---   & .451 \\
Czech              & cs & 4 &  11  & Fusional  & Indo-Europ.  &    222 & 0.56 & .436  & .385 \\
Romanian           & ro & 3 &  24  & Fusional  & Indo-Europ.  &    200 & 0.43 & .537  & .469 \\
Vietnamese         & vi & 4 &  76  & Isolating & Austroasiatic&    343 & 1.31 & .525  & .423 \\
German             & de & 5 &  77  & Fusional  & Indo-Europ.  & 1\,655 & 2.98 & .597  & .450 \\
Slovak             & sk & 3 &   6  & Fusional  & Indo-Europ.  &     92 & 0.26 & .436  & .385 \\
Italian            & it & 4 &  65  & Fusional  & Indo-Europ.  &    794 & 1.86 & .524  & .421 \\
Azerbaijani        & az & 1 &  24  & Agglut.   & Turkic       &     29 & 0.22 & .451  & .362 \\
Japanese           & ja & 5 & 128  & Agglut.   & Japonic      & 1\,650 & 1.41 & .421  & .459 \\
Ukrainian          & uk & 3 &  27  & Fusional  & Indo-Europ.  &    274 & 1.33 & .541  & .465 \\
Arabic             & ar & 5 & 335  & Fusional  & Afro-Asiatic &    315 & 1.22 & .432  & .348 \\
Greek              & el & 3 &  15  & Fusional  & Indo-Europ.  &    238 & 0.21 & .500  & .403 \\
Hindi              & hi & 4 & 341  & Fusional  & Indo-Europ.  &    130 & 1.62 & .513  & .463 \\
Russian            & ru & 4 & 154  & Fusional  & Indo-Europ.  & 6\,400 & 1.99 & .503  & .469 \\
\bottomrule
\end{tabular}
\caption{Language metadata for the 14 languages in this study. ISO codes follow ISO~639-1
and ISO~639-3. Resource tier by \citep{joshi-etal-2020-state} (0 = lowest,
5 = highest). FineWeb2 corpus size is given in gigabytes. Linguistic similarity (WALS.\
sim.) is $1 - d_{\mathrm{featural}}$ computed via URIEL+~\citep{khan-etal-2025-uriel},
where higher values indicate greater typological similarity; EN = vs.\ English, ZH =
vs.\ Mandarin Chinese.}
\label{tab:languages}
\end{table*}

% \am{Qwen 2.5 VL has mentioned somewhere the use of at least 10 languages from the OCR dataset: "French, German, Italian, Spanish, Portuguese, Arabic, Russian, Japanese, Korean, and Vietnamese"}

\begin{table*}[t]
\centering\footnotesize
\setlength{\tabcolsep}{4.5pt}

\begin{tabular}{llllllllllllllll}
\toprule
Model & ar & az & cs & de & el & en & hi & it & ja & ro & ru & sk & uk & vi & Notes \\
\midrule
Gemma 3 12B & ? & ? & ? & ? & ? & ? & ? & ? & ? & ? & ? & ? & ? & ? & $>$ 140 languages \\
Qwen 2.5 VL 7B & \checkmark & ? & ? & \checkmark & ? & \checkmark & ? & \checkmark & \checkmark & ? & \checkmark & ? & ? & \checkmark & 10 languages (OCR data) \\
Qwen 3 VL 8B & \checkmark & ? & ? & \checkmark & ? & \checkmark & ? & \checkmark & \checkmark & ? & \checkmark & ? & ? & ? & 119 languages and dialects \\
Aya Vision 8B & \checkmark & - & \checkmark & \checkmark & \checkmark & \checkmark & \checkmark & \checkmark & \checkmark & \checkmark & \checkmark & - & \checkmark & \checkmark & 23 languages \\
Jina VL & \checkmark & - & - & \checkmark & - & \checkmark & \checkmark & \checkmark & \checkmark & - & \checkmark & - & - & \checkmark & 17 languages \\
% eurovllm-9b & \checkmark & - & \checkmark & \checkmark & \checkmark & \checkmark & \checkmark & \checkmark & \checkmark & \checkmark & \checkmark & \checkmark & \checkmark & - & 35 European languages \\
% llama4 scout & \checkmark & - & - & \checkmark & - & \checkmark & \checkmark & \checkmark & - & - & - & - & - & \checkmark & 12 languages\\
NLLB SigLIP Family& \checkmark & \checkmark & \checkmark & \checkmark & \checkmark & \checkmark & \checkmark & \checkmark & \checkmark & \checkmark & \checkmark & \checkmark & \checkmark & \checkmark & 200 languages \\
MEXMA SigLIP2 & \checkmark & \checkmark & \checkmark & \checkmark & \checkmark & \checkmark & \checkmark & \checkmark & \checkmark & \checkmark & \checkmark & \checkmark & \checkmark & \checkmark & 80 languages \\
SigLIP 2 Family & ? & ? & ? & ? & ? & ? & ? & ? & ? & ? & ? & ? & ? & ? & 109 languages \\
\bottomrule
\end{tabular}

\caption{Overview of training data availability across the evaluated models for the 14 languages included in our dataset. Symbols indicate whether the training data is available (\checkmark), the model was not trained on the language (-), or the exact data composition remains undisclosed (?).}

\label{tab:training_data}
\end{table*}

\begin{table*}[t]
\centering\footnotesize
\setlength{\tabcolsep}{4.5pt}

\begin{tabular}{lllllll}
\toprule
\multicolumn{2}{c}{Dataset} & \# lang. & GBI & \# captions & \# images & \% population \\
\toprule
MARVL & \citet{liu-etal-2021-visually} & 5 & 56.41 & 5.7K & 4.9K & 13.79\% \\
M5 (VGR,VLOD) & \citet{schneider-sitaram-2024-m5} & 12 & 74.87 & 1.4K & 6.8K & 16.79\% \\
ALM-bench & \citet{vayani2025languagesmatterevaluatinglmms} & 99 & 93.84 & 22K & 3K & 59.92\% \\
XM3600 & \citet{thapliyal-etal-2022-crossmodal} & 36 & 91.79 & 261K & 3.6K & 49.18\% \\
MaXM & \citet{changpinyo-etal-2023-maxm} & 7 & 64.61 & 2K & 0.3K & 21.45\% \\
MTVQA &  \citet{tang2025mtvqabenchmarkingmultilingualtextcentric} & 10 & 77.94 & 28.6K & 8.7K & 16.39\% \\
CVQA &  \citet{romero2024cvqaculturallydiversemultilingualvisual} & 31 & 88.2 & 10.3K & 5.2K & 40.89\% \\
\midrule
XVNLI & \citet{bugliarello2022} & 5 & 49.74 & 1.1k & 357 & 16.89\% \\
xGQA & \citet{nooralahzadeh2022} & 8 & 63.58 & 2.4M & 85K & 21.76\% \\
xFlickrCo & \citet{bugliarello2022} & 8 & 65.12 & 2k & 2k & 27.54\% \\
Multi30k & \citet{barrault-etal-2018-findings} & 4 & 33.84 & 620K & 31K & 6.60\% \\
XTD & \citet{xtd} & 7 & 71.28 & 7K & 1K & 22.13\% \\
CoMMuTE & \citet{futeral-etal-2023-tackling} & 6 & 66.15 & 1.9K & 0.3K & 18.61\% \\
xMMMU & \citet{yue2025pangeafullyopenmultilingual} & 7 & 69.23 & 2.6K & 300 & 18.04\% \\
KnowRecall & \citet{wang2025travelinglanguagesbenchmarkingcrosslingual}  & 15 & 77.94 & 45K & 1.5K & 35.26\% \\

\cellcolor{Goldenrod!60}\textbf{MultiInpaintCOCO} & \cellcolor{Goldenrod!60}  & \cellcolor{Goldenrod!60}14 & \cellcolor{Goldenrod!60}78.97 & \cellcolor{Goldenrod!60}26K & \cellcolor{Goldenrod!60}1.9K & \cellcolor{Goldenrod!60}19.26\% \\
\bottomrule
\end{tabular}

\caption{Dataset language coverage overview. The first half contains datasets with cultural knowledge, whereas the second ones are created by translating English samples.}

\label{tab:datasets}
\end{table*}

\begin{table*}

\centering\footnotesize
\renewcommand{\arraystretch}{1.6}
\setlength{\tabcolsep}{2.2pt}
\begin{tabular}{lcccccccccccccc}
\toprule
Model & ar & az & cs & de & el & en & hi & it & ja & ro & ru & sk & uk & vi \\
\midrule
Gemma 3 12B & \begin{minipage}{2.2em}\centering 16.6\\[-4pt] {\tiny 4--56}\end{minipage} & \begin{minipage}{2.2em}\centering 19.5\\[-4pt] {\tiny 5--78}\end{minipage} & \begin{minipage}{2.2em}\centering 17.2\\[-4pt] {\tiny 4--63}\end{minipage} & \begin{minipage}{2.2em}\centering 15.5\\[-4pt] {\tiny 5--51}\end{minipage} & \begin{minipage}{2.2em}\centering 25.9\\[-4pt] {\tiny 8--74}\end{minipage} & \begin{minipage}{2.2em}\centering 12.3\\[-4pt] {\tiny 4--40}\end{minipage} & \begin{minipage}{2.2em}\centering 16.3\\[-4pt] {\tiny 4--37}\end{minipage} & \begin{minipage}{2.2em}\centering 16.1\\[-4pt] {\tiny 5--50}\end{minipage} & \begin{minipage}{2.2em}\centering 13.5\\[-4pt] {\tiny 3--46}\end{minipage} & \begin{minipage}{2.2em}\centering 19.2\\[-4pt] {\tiny 5--63}\end{minipage} & \begin{minipage}{2.2em}\centering 16.6\\[-4pt] {\tiny 6--50}\end{minipage} & \begin{minipage}{2.2em}\centering 17.9\\[-4pt] {\tiny 4--69}\end{minipage} & \begin{minipage}{2.2em}\centering 19.4\\[-4pt] {\tiny 5--65}\end{minipage} & \begin{minipage}{2.2em}\centering 18.3\\[-4pt] {\tiny 6--60}\end{minipage} \\
Qwen 2.5 VL 7B & \begin{minipage}{2.2em}\centering 17.7\\[-4pt] {\tiny 2--62}\end{minipage} & \begin{minipage}{2.2em}\centering 22.0\\[-4pt] {\tiny 4--89}\end{minipage} & \begin{minipage}{2.2em}\centering 20.5\\[-4pt] {\tiny 4--83}\end{minipage} & \begin{minipage}{2.2em}\centering 16.7\\[-4pt] {\tiny 4--55}\end{minipage} & \begin{minipage}{2.2em}\centering 50.9\\[-4pt] {\tiny 13--144}\end{minipage} & \begin{minipage}{2.2em}\centering 11.5\\[-4pt] {\tiny 3--39}\end{minipage} & \begin{minipage}{2.2em}\centering 49.0\\[-4pt] {\tiny 9--96}\end{minipage} & \begin{minipage}{2.2em}\centering 18.2\\[-4pt] {\tiny 4--56}\end{minipage} & \begin{minipage}{2.2em}\centering 15.8\\[-4pt] {\tiny 3--59}\end{minipage} & \begin{minipage}{2.2em}\centering 21.1\\[-4pt] {\tiny 4--69}\end{minipage} & \begin{minipage}{2.2em}\centering 20.6\\[-4pt] {\tiny 6--62}\end{minipage} & \begin{minipage}{2.2em}\centering 21.1\\[-4pt] {\tiny 4--82}\end{minipage} & \begin{minipage}{2.2em}\centering 26.1\\[-4pt] {\tiny 4--85}\end{minipage} & \begin{minipage}{2.2em}\centering 19.4\\[-4pt] {\tiny 6--69}\end{minipage} \\
Qwen 3 VL 8B & \begin{minipage}{2.2em}\centering 17.7\\[-4pt] {\tiny 2--62}\end{minipage} & \begin{minipage}{2.2em}\centering 22.0\\[-4pt] {\tiny 4--89}\end{minipage} & \begin{minipage}{2.2em}\centering 20.5\\[-4pt] {\tiny 4--83}\end{minipage} & \begin{minipage}{2.2em}\centering 16.7\\[-4pt] {\tiny 4--55}\end{minipage} & \begin{minipage}{2.2em}\centering 50.9\\[-4pt] {\tiny 13--144}\end{minipage} & \begin{minipage}{2.2em}\centering 11.5\\[-4pt] {\tiny 3--39}\end{minipage} & \begin{minipage}{2.2em}\centering 49.0\\[-4pt] {\tiny 9--96}\end{minipage} & \begin{minipage}{2.2em}\centering 18.2\\[-4pt] {\tiny 4--56}\end{minipage} & \begin{minipage}{2.2em}\centering 15.8\\[-4pt] {\tiny 3--59}\end{minipage} & \begin{minipage}{2.2em}\centering 21.1\\[-4pt] {\tiny 4--69}\end{minipage} & \begin{minipage}{2.2em}\centering 20.6\\[-4pt] {\tiny 6--62}\end{minipage} & \begin{minipage}{2.2em}\centering 21.1\\[-4pt] {\tiny 4--82}\end{minipage} & \begin{minipage}{2.2em}\centering 26.1\\[-4pt] {\tiny 4--85}\end{minipage} & \begin{minipage}{2.2em}\centering 19.4\\[-4pt] {\tiny 6--69}\end{minipage} \\
Aya Vision & \begin{minipage}{2.2em}\centering 16.8\\[-4pt] {\tiny 4--54}\end{minipage} & \begin{minipage}{2.2em}\centering 20.2\\[-4pt] {\tiny 4--83}\end{minipage} & \begin{minipage}{2.2em}\centering 15.5\\[-4pt] {\tiny 4--54}\end{minipage} & \begin{minipage}{2.2em}\centering 15.7\\[-4pt] {\tiny 4--50}\end{minipage} & \begin{minipage}{2.2em}\centering 20.0\\[-4pt] {\tiny 5--57}\end{minipage} & \begin{minipage}{2.2em}\centering 12.4\\[-4pt] {\tiny 4--40}\end{minipage} & \begin{minipage}{2.2em}\centering 34.3\\[-4pt] {\tiny 7--74}\end{minipage} & \begin{minipage}{2.2em}\centering 15.8\\[-4pt] {\tiny 5--49}\end{minipage} & \begin{minipage}{2.2em}\centering 14.4\\[-4pt] {\tiny 4--51}\end{minipage} & \begin{minipage}{2.2em}\centering 16.9\\[-4pt] {\tiny 5--57}\end{minipage} & \begin{minipage}{2.2em}\centering 16.1\\[-4pt] {\tiny 4--45}\end{minipage} & \begin{minipage}{2.2em}\centering 18.1\\[-4pt] {\tiny 4--66}\end{minipage} & \begin{minipage}{2.2em}\centering 17.6\\[-4pt] {\tiny 5--58}\end{minipage} & \begin{minipage}{2.2em}\centering 17.7\\[-4pt] {\tiny 6--60}\end{minipage} \\
Jina VL & \begin{minipage}{2.2em}\centering 17.7\\[-4pt] {\tiny 2--62}\end{minipage} & \begin{minipage}{2.2em}\centering 22.0\\[-4pt] {\tiny 4--89}\end{minipage} & \begin{minipage}{2.2em}\centering 20.5\\[-4pt] {\tiny 4--83}\end{minipage} & \begin{minipage}{2.2em}\centering 16.7\\[-4pt] {\tiny 4--55}\end{minipage} & \begin{minipage}{2.2em}\centering 50.9\\[-4pt] {\tiny 13--144}\end{minipage} & \begin{minipage}{2.2em}\centering 11.5\\[-4pt] {\tiny 3--39}\end{minipage} & \begin{minipage}{2.2em}\centering 49.0\\[-4pt] {\tiny 9--96}\end{minipage} & \begin{minipage}{2.2em}\centering 18.2\\[-4pt] {\tiny 4--56}\end{minipage} & \begin{minipage}{2.2em}\centering 15.8\\[-4pt] {\tiny 3--59}\end{minipage} & \begin{minipage}{2.2em}\centering 21.1\\[-4pt] {\tiny 4--69}\end{minipage} & \begin{minipage}{2.2em}\centering 20.6\\[-4pt] {\tiny 6--62}\end{minipage} & \begin{minipage}{2.2em}\centering 21.1\\[-4pt] {\tiny 4--82}\end{minipage} & \begin{minipage}{2.2em}\centering 26.1\\[-4pt] {\tiny 4--85}\end{minipage} & \begin{minipage}{2.2em}\centering 19.4\\[-4pt] {\tiny 6--69}\end{minipage} \\
\midrule
NLLB-SigLIP Base & \begin{minipage}{2.2em}\centering 17.9\\[-4pt] {\tiny 5--59}\end{minipage} & \begin{minipage}{2.2em}\centering 17.1\\[-4pt] {\tiny 5--63}\end{minipage} & \begin{minipage}{2.2em}\centering 17.7\\[-4pt] {\tiny 5--63}\end{minipage} & \begin{minipage}{2.2em}\centering 18.6\\[-4pt] {\tiny 6--58}\end{minipage} & \begin{minipage}{2.2em}\centering 22.9\\[-4pt] {\tiny 8--66}\end{minipage} & \begin{minipage}{2.2em}\centering 14.8\\[-4pt] {\tiny 5--46}\end{minipage} & \begin{minipage}{2.2em}\centering 18.8\\[-4pt] {\tiny 4--40}\end{minipage} & \begin{minipage}{2.2em}\centering 17.9\\[-4pt] {\tiny 6--55}\end{minipage} & \begin{minipage}{2.2em}\centering 15.3\\[-4pt] {\tiny 5--48}\end{minipage} & \begin{minipage}{2.2em}\centering 19.5\\[-4pt] {\tiny 6--67}\end{minipage} & \begin{minipage}{2.2em}\centering 19.1\\[-4pt] {\tiny 7--55}\end{minipage} & \begin{minipage}{2.2em}\centering 17.6\\[-4pt] {\tiny 5--65}\end{minipage} & \begin{minipage}{2.2em}\centering 19.8\\[-4pt] {\tiny 5--63}\end{minipage} & \begin{minipage}{2.2em}\centering 19.7\\[-4pt] {\tiny 7--62}\end{minipage} \\
NLLB-SigLIP Large & \begin{minipage}{2.2em}\centering 17.9\\[-4pt] {\tiny 5--59}\end{minipage} & \begin{minipage}{2.2em}\centering 17.1\\[-4pt] {\tiny 5--63}\end{minipage} & \begin{minipage}{2.2em}\centering 17.7\\[-4pt] {\tiny 5--63}\end{minipage} & \begin{minipage}{2.2em}\centering 18.6\\[-4pt] {\tiny 6--58}\end{minipage} & \begin{minipage}{2.2em}\centering 22.9\\[-4pt] {\tiny 8--66}\end{minipage} & \begin{minipage}{2.2em}\centering 14.8\\[-4pt] {\tiny 5--46}\end{minipage} & \begin{minipage}{2.2em}\centering 18.8\\[-4pt] {\tiny 4--40}\end{minipage} & \begin{minipage}{2.2em}\centering 17.9\\[-4pt] {\tiny 6--55}\end{minipage} & \begin{minipage}{2.2em}\centering 15.3\\[-4pt] {\tiny 5--48}\end{minipage} & \begin{minipage}{2.2em}\centering 19.5\\[-4pt] {\tiny 6--67}\end{minipage} & \begin{minipage}{2.2em}\centering 19.1\\[-4pt] {\tiny 7--55}\end{minipage} & \begin{minipage}{2.2em}\centering 17.6\\[-4pt] {\tiny 5--65}\end{minipage} & \begin{minipage}{2.2em}\centering 19.8\\[-4pt] {\tiny 5--63}\end{minipage} & \begin{minipage}{2.2em}\centering 19.7\\[-4pt] {\tiny 7--62}\end{minipage} \\
MEXMA-SigLIP2 & \begin{minipage}{2.2em}\centering 16.5\\[-4pt] {\tiny 5--52}\end{minipage} & \begin{minipage}{2.2em}\centering 16.1\\[-4pt] {\tiny 5--60}\end{minipage} & \begin{minipage}{2.2em}\centering 16.5\\[-4pt] {\tiny 5--59}\end{minipage} & \begin{minipage}{2.2em}\centering 17.1\\[-4pt] {\tiny 6--53}\end{minipage} & \begin{minipage}{2.2em}\centering 20.9\\[-4pt] {\tiny 7--62}\end{minipage} & \begin{minipage}{2.2em}\centering 15.0\\[-4pt] {\tiny 5--47}\end{minipage} & \begin{minipage}{2.2em}\centering 19.2\\[-4pt] {\tiny 4--40}\end{minipage} & \begin{minipage}{2.2em}\centering 17.2\\[-4pt] {\tiny 5--53}\end{minipage} & \begin{minipage}{2.2em}\centering 16.8\\[-4pt] {\tiny 6--53}\end{minipage} & \begin{minipage}{2.2em}\centering 18.4\\[-4pt] {\tiny 5--62}\end{minipage} & \begin{minipage}{2.2em}\centering 17.4\\[-4pt] {\tiny 6--53}\end{minipage} & \begin{minipage}{2.2em}\centering 16.6\\[-4pt] {\tiny 5--62}\end{minipage} & \begin{minipage}{2.2em}\centering 18.2\\[-4pt] {\tiny 5--61}\end{minipage} & \begin{minipage}{2.2em}\centering 19.3\\[-4pt] {\tiny 7--62}\end{minipage} \\
SigLIP2 Base & \begin{minipage}{2.2em}\centering 16.9\\[-4pt] {\tiny 4--56}\end{minipage} & \begin{minipage}{2.2em}\centering 19.5\\[-4pt] {\tiny 5--80}\end{minipage} & \begin{minipage}{2.2em}\centering 16.8\\[-4pt] {\tiny 5--62}\end{minipage} & \begin{minipage}{2.2em}\centering 14.5\\[-4pt] {\tiny 4--48}\end{minipage} & \begin{minipage}{2.2em}\centering 26.4\\[-4pt] {\tiny 8--76}\end{minipage} & \begin{minipage}{2.2em}\centering 12.3\\[-4pt] {\tiny 4--40}\end{minipage} & \begin{minipage}{2.2em}\centering 22.7\\[-4pt] {\tiny 6--48}\end{minipage} & \begin{minipage}{2.2em}\centering 15.1\\[-4pt] {\tiny 5--46}\end{minipage} & \begin{minipage}{2.2em}\centering 12.6\\[-4pt] {\tiny 3--43}\end{minipage} & \begin{minipage}{2.2em}\centering 18.8\\[-4pt] {\tiny 5--61}\end{minipage} & \begin{minipage}{2.2em}\centering 15.7\\[-4pt] {\tiny 5--50}\end{minipage} & \begin{minipage}{2.2em}\centering 17.4\\[-4pt] {\tiny 5--67}\end{minipage} & \begin{minipage}{2.2em}\centering 19.2\\[-4pt] {\tiny 5--65}\end{minipage} & \begin{minipage}{2.2em}\centering 17.9\\[-4pt] {\tiny 6--62}\end{minipage} \\
SigLIP2 Large & \begin{minipage}{2.2em}\centering 16.9\\[-4pt] {\tiny 4--56}\end{minipage} & \begin{minipage}{2.2em}\centering 19.5\\[-4pt] {\tiny 5--80}\end{minipage} & \begin{minipage}{2.2em}\centering 16.8\\[-4pt] {\tiny 5--62}\end{minipage} & \begin{minipage}{2.2em}\centering 14.5\\[-4pt] {\tiny 4--48}\end{minipage} & \begin{minipage}{2.2em}\centering 26.4\\[-4pt] {\tiny 8--76}\end{minipage} & \begin{minipage}{2.2em}\centering 12.3\\[-4pt] {\tiny 4--40}\end{minipage} & \begin{minipage}{2.2em}\centering 22.7\\[-4pt] {\tiny 6--48}\end{minipage} & \begin{minipage}{2.2em}\centering 15.1\\[-4pt] {\tiny 5--46}\end{minipage} & \begin{minipage}{2.2em}\centering 12.6\\[-4pt] {\tiny 3--43}\end{minipage} & \begin{minipage}{2.2em}\centering 18.8\\[-4pt] {\tiny 5--61}\end{minipage} & \begin{minipage}{2.2em}\centering 15.7\\[-4pt] {\tiny 5--50}\end{minipage} & \begin{minipage}{2.2em}\centering 17.4\\[-4pt] {\tiny 5--67}\end{minipage} & \begin{minipage}{2.2em}\centering 19.2\\[-4pt] {\tiny 5--65}\end{minipage} & \begin{minipage}{2.2em}\centering 17.9\\[-4pt] {\tiny 6--62}\end{minipage} \\
SigLIP2 SO400M & \begin{minipage}{2.2em}\centering 16.9\\[-4pt] {\tiny 4--56}\end{minipage} & \begin{minipage}{2.2em}\centering 19.5\\[-4pt] {\tiny 5--80}\end{minipage} & \begin{minipage}{2.2em}\centering 16.8\\[-4pt] {\tiny 5--62}\end{minipage} & \begin{minipage}{2.2em}\centering 14.5\\[-4pt] {\tiny 4--48}\end{minipage} & \begin{minipage}{2.2em}\centering 26.4\\[-4pt] {\tiny 8--76}\end{minipage} & \begin{minipage}{2.2em}\centering 12.3\\[-4pt] {\tiny 4--40}\end{minipage} & \begin{minipage}{2.2em}\centering 22.7\\[-4pt] {\tiny 6--48}\end{minipage} & \begin{minipage}{2.2em}\centering 15.1\\[-4pt] {\tiny 5--46}\end{minipage} & \begin{minipage}{2.2em}\centering 12.6\\[-4pt] {\tiny 3--43}\end{minipage} & \begin{minipage}{2.2em}\centering 18.8\\[-4pt] {\tiny 5--61}\end{minipage} & \begin{minipage}{2.2em}\centering 15.7\\[-4pt] {\tiny 5--50}\end{minipage} & \begin{minipage}{2.2em}\centering 17.4\\[-4pt] {\tiny 5--67}\end{minipage} & \begin{minipage}{2.2em}\centering 19.2\\[-4pt] {\tiny 5--65}\end{minipage} & \begin{minipage}{2.2em}\centering 17.9\\[-4pt] {\tiny 6--62}\end{minipage} \\
SigLIP2 Giant & \begin{minipage}{2.2em}\centering 16.9\\[-4pt] {\tiny 4--56}\end{minipage} & \begin{minipage}{2.2em}\centering 19.5\\[-4pt] {\tiny 5--80}\end{minipage} & \begin{minipage}{2.2em}\centering 16.8\\[-4pt] {\tiny 5--62}\end{minipage} & \begin{minipage}{2.2em}\centering 14.5\\[-4pt] {\tiny 4--48}\end{minipage} & \begin{minipage}{2.2em}\centering 26.4\\[-4pt] {\tiny 8--76}\end{minipage} & \begin{minipage}{2.2em}\centering 12.3\\[-4pt] {\tiny 4--40}\end{minipage} & \begin{minipage}{2.2em}\centering 22.7\\[-4pt] {\tiny 6--48}\end{minipage} & \begin{minipage}{2.2em}\centering 15.1\\[-4pt] {\tiny 5--46}\end{minipage} & \begin{minipage}{2.2em}\centering 12.6\\[-4pt] {\tiny 3--43}\end{minipage} & \begin{minipage}{2.2em}\centering 18.8\\[-4pt] {\tiny 5--61}\end{minipage} & \begin{minipage}{2.2em}\centering 15.7\\[-4pt] {\tiny 5--50}\end{minipage} & \begin{minipage}{2.2em}\centering 17.4\\[-4pt] {\tiny 5--67}\end{minipage} & \begin{minipage}{2.2em}\centering 19.2\\[-4pt] {\tiny 5--65}\end{minipage} & \begin{minipage}{2.2em}\centering 17.9\\[-4pt] {\tiny 6--62}\end{minipage} \\
\bottomrule
\end{tabular}
\caption{Tokenizer statistics of the tested models for the MultiInpaintCOCO dataset. The table shows the average number of tokens per model-language pair and the range of text lengths.}
\label{tab:tokenizer}
\end{table*}

%----------------
\begin{table*}[t]
\centering
\small
\setlength{\tabcolsep}{8pt}
\begin{tabular}{>{\centering\arraybackslash}p{0.4cm} p{2cm} p{5.3cm} p{5.3cm}}
\toprule
\# & Query & Option (1) & Option (2) \\
\midrule
1 & \includegraphics[width=2cm]{images/coco_110.png}
  & \textbf{A man is feeding an elephant over a fence}
  & A man is feeding a giraffe over a fence \\[1em]
2 & \includegraphics[width=2cm]{images/coco_110.png}
  & A man is feeding a giraffe over a fence
  & \textbf{A man is feeding an elephant over a fence} \\[1em]
3 & \includegraphics[width=2cm]{images/inpaint_110.png}
  & A man is feeding an elephant over a fence
  & \textbf{A man is feeding a giraffe over a fence} \\[1em]
4 & \includegraphics[width=2cm]{images/inpaint_110.png}
  & \textbf{A man is feeding a giraffe over a fence}
  & A man is feeding an elephant over a fence \\
\bottomrule
\end{tabular}

\begin{tabular}{>{\centering\arraybackslash}p{0.4cm} p{6cm} >{\centering\arraybackslash}p{3.3cm} >{\centering\arraybackslash}p{3.3cm}}
%\# & Query & Option (1) & Option (2) \\
%\midrule
5 & A man is feeding an elephant over a fence
  & \fbox{\includegraphics[width=2cm]{images/coco_110.png}}
  & \includegraphics[width=2cm]{images/inpaint_110.png} \\
&  & \textbf{(1) correct} & \\[1em]
6 & A man is feeding an elephant over a fence
  & \includegraphics[width=2cm]{images/inpaint_110.png}
  & \fbox{\includegraphics[width=2cm]{images/coco_110.png}} \\
&  & & \textbf{(2) correct} \\[1em]
7 & A man is feeding a giraffe over a fence
  & \includegraphics[width=2cm]{images/coco_110.png}
  & \fbox{\includegraphics[width=2cm]{images/inpaint_110.png}} \\
&  & & \textbf{(2) correct} \\[1em]
8 & A man is feeding a giraffe over a fence
  & \fbox{\includegraphics[width=2cm]{images/inpaint_110.png}}
  & \includegraphics[width=2cm]{images/coco_110.png} \\
&  & \textbf{(1) correct} & \\
\bottomrule
\end{tabular}
%\caption{Above -- \textit{2txt} configurations: the image is the query and the model selects the correct caption. Correct answers in \textbf{bold}. \\
%Below -- \textit{2img} configurations: the caption is the query and the model selects the correct image. The correct option is framed and labeled.}
\caption{For decoder-only models, each example is evaluated in both input orders to account for order sensitivity, resulting in eight comparison configurations per example. Above -- \textit{2txt} configurations: the image is the query and the model selects the correct caption; correct answers are shown in \textbf{bold}. Below -- \textit{2img} configurations: the caption is the query and the model selects the correct image; the correct option is framed and labeled.}\label{tab:configs_2img}
\end{table*}

%-------------------------------------------
\begin{table*}[t]
\centering
\small
\setlength{\tabcolsep}{4pt}
\begin{tabular}{lllll}
\toprule
% Params (V + T  + (X)= Total)
Model & Hugging Face Identifiers & Params (V + T + (X)= Total) & Img Size & Vocab Size \\
\midrule
Gemma 3 12B  & \href{https://huggingface.co/google/gemma-3-12b-it}{google/gemma-3-12b-it} & $417\,\text{M} + 11.8\,\text{B} = 12.2\,\text{B}$ & 896 & 262,144 \\
Qwen 7B & \href{https://huggingface.co/Qwen/Qwen2.5-VL-7B-Instruct}{Qwen/Qwen2.5-VL-7B-Instruct} & $677\,\text{M} + 7.1\,\text{B} = 7.8\,\text{B}$ & 3584 & 151,643 \\
Qwen3 8B & \href{https://huggingface.co/Qwen/Qwen3-VL-8B-Instruct}{Qwen/Qwen3-VL-8B-Instruct} & $576\,\text{M} + 7.6\,\text{B} = 8.1\,\text{B}$ & 3584 & 151,643 \\
Aya Vision 8B & \href{https://huggingface.co/CohereLabs/aya-vision-8b}{CohereLabs/aya-vision-8b} & $413\,\text{M} + 8.0\,\text{B} + 190.9\,\text{M}= 8.6\,\text{B}$ & 364 & 255,000 \\
Jina VL & \href{https://huggingface.co/jinaai/jina-vlm}{jinaai/jina-vlm} & $449\,\text{M} + 1.72\,\text{B} + 311\,\text{M} = 2.5\,\text{B}$ & 1176×910 & 151,643 \\
NLLB-SigLIP Base & \href{https://huggingface.co/visheratin/nllb-clip-base-siglip}{nllb-clip-base-siglip} & $93\,\text{M} + 414\,\text{M} = 508\,\text{M}$ & 384 & 256,204 \\
NLLB-SigLIP Large & \href{https://huggingface.co/visheratin/nllb-clip-large-siglip}{nllb-clip-large-siglip}& $428\,\text{M} + 767\,\text{M} = 1.2\,\text{B}$ & 384 & 256,204 \\
MEXMA-SigLIP2 & \href{https://huggingface.co/visheratin/mexma-siglip2}{visheratin/mexma-siglip2} & $428\,\text{M} + 573\,\text{M} = \text{1B}$ & 512 & 250,002 \\
SigLIP2 Base & \href{https://huggingface.co/google/siglip2-base-patch16-224}{google/siglip2-base-patch16-224} & $92\,\text{M} + 282\,\text{M} = 375\,\text{M}$ & 224 & 256,000 \\
SigLIP2 Large & \href{https://huggingface.co/google/siglip2-large-patch16-256}{google/siglip2-large-patch16-256} & $316\,\text{M} + 565\,\text{M} = 881\,\text{M}$ & 256 & 256,000 \\
SigLIP2 SO400M & \href{https://huggingface.co/google/siglip2-so400m-patch16-256}{google/siglip2-so400m-patch16-256} & $428\,\text{M} + 708\,\text{M} = 1.14\,\text{B}$ & 256 & 256,000 \\
SigLIP2 Giant & \href{https://huggingface.co/google/siglip2-giant-opt-patch16-256}{google/siglip2-giant-opt-patch16-256} & $1.16\,\text{B} + 708\,\text{M} = 1.87\,\text{B}$ & 256 & 256,000 \\
\bottomrule
\end{tabular}
\caption{Model overview with Hugging Face identifiers, parameter counts split into vision (V), text (T), and optional cross-modal (X) components, maximum image size, and tokenizer vocabulary size. Image sizes are square resolutions unless otherwise indicated.}\label{tab:model-details}
\end{table*}
%-------------------------------------------

\section{Position Bias}\label{app:position}

Since the image order can introduce a small amount of noise in the results, we analyze how significant the noise is. We compute the number of mistakes for both query texts when the images are given in the direct order, $M_{\text{dir}}$, and in the swapped order, $M_{\text{swap}}$.
\begin{eqnarray}
    M_{\text{dir}} & = & \mathbb{0}_{c(T_\text{o}, (I_\text{o}, I_\text{e}))} + \mathbb{0}_{c(T_\text{e}, (I_\text{o}, I_\text{e}))} \label{eq:m-dir}  \\
    M_{\text{swap}} & = & \mathbb{0}_{c(T_\text{o}, (I_\text{e}, I_\text{o}))} + \mathbb{0}_{c(T_\text{e}, (I_\text{e}, I_\text{o}))} \label{eq:m-swap}  \\
    M_{\text{order}} & = & M_{\text{dir}} - M_{\text{swap}} \label{eq:m-order} 
\end{eqnarray}
where the $\mathbb{0}_{c(q, (a, b))}$ represents the number of times that $c(q, (a, b)) = 0$. We report, for each model, the difference $M_{\text{order}}$: a negative value indicates a bias towards the dataset order, while a positive value indicates a bias towards the swap. 

We report the position bias in Table~\ref{tab:order_image_corr}. The impact of the generative decoder models is visible but not highly significant. However, we can still see that the lowest values are for the Aya model, indicating a preference for the swapped order. On the opposite pole, the Jina model has the highest values, preferring the dataset image order. For the dual encoders, we do not plot the values because they are always 0. This is because the modality similarity is obtained from an outer product, except in the case when the similarity is equal. 

\begin{table*}[t]
\centering\footnotesize
\setlength{\tabcolsep}{4.5pt}
\begin{tabular}{lccccccccccccccc}
\toprule
Model & ar & az & cs & de & el & en & hi & it & ja & ro & ru & sk & uk & vi & Avg. \\
\midrule
Gemma 3 12B & \FP{.01} & \FP{.02} & \FP{-.03} & \FP{.00} & \FP{-.02} & \FP{-.06} & \FP{-.02} & \FP{.00} & \FP{-.03} & \FP{-.03} & \FP{-.02} & \FP{.01} & \FP{-.03} & \FP{-.02} & \FP{-.01} \\
Qwen 2.5 VL 7B & \FP{-.04} & \FP{-.03} & \FP{-.03} & \FP{-.06} & \FP{-.05} & \FP{-.03} & \FP{-.04} & \FP{-.04} & \FP{-.04} & \FP{-.01} & \FP{-.03} & \FP{-.01} & \FP{-.04} & \FP{-.04} & \FP{-.04} \\
Qwen 3 VL 8B & \FP{-.03} & \FP{-.05} & \FP{-.02} & \FP{-.04} & \FP{-.03} & \FP{-.06} & \FP{-.03} & \FP{-.05} & \FP{-.03} & \FP{-.03} & \FP{-.05} & \FP{-.03} & \FP{-.02} & \FP{-.03} & \FP{-.04} \\
Aya Vision 8B & \FP{-.07} & \FP{-.04} & \FP{-.06} & \FP{-.06} & \FP{-.07} & \FP{-.07} & \FP{-.10} & \FP{-.10} & \FP{-.07} & \FP{-.06} & \FP{-.06} & \FP{-.06} & \FP{-.07} & \FP{-.09} & \FP{-.07} \\
Jina VL & \FP{.03} & \FP{-.01} & \FP{-.03} & \FP{.05} & \FP{.11} & \FP{-.04} & \FP{.04} & \FP{.03} & \FP{.01} & \FP{.05} & \FP{.03} & \FP{.04} & \FP{.05} & \FP{.04} & \FP{.03} \\
\midrule
\midrule
Avg. & \FP{-.02} & \FP{-.02} & \FP{-.04} & \FP{-.02} & \FP{-.01} & \FP{-.05} & \FP{-.03} & \FP{-.03} & \FP{-.03} & \FP{-.02} & \FP{-.02} & \FP{-.01} & \FP{-.02} & \FP{-.03} & \FP{-.03} \\
\bottomrule
\end{tabular}
\caption{Per-language position bias for each model, measured as $M_{\text{order}}$ from Equation~(\ref{eq:m-order}). Negative values indicate a preference for the dataset image order, whereas positive values indicate a preference for the swapped order.}
\label{tab:order_image_corr}
\end{table*}

\section{Additional Results}\label{app:results}
% ################################################

\begin{table*}[htbp]
\centering\footnotesize
\setlength{\tabcolsep}{4.2pt}
\begin{tabular}{lcccccccccccccc@{\hskip 16pt}c}
\toprule
Model & ar & az & cs & de & el & en & hi & it & ja & ro & ru & sk & uk & vi & Avg. \\
\midrule
Gemma 3 12B & \Acc{.639} & \Acc{.501} & \Acc{.562} & \Acc{.638} & \Acc{.652} & \Acc{.608} & \Acc{.578} & \Acc{.648} & \Acc{.571} & \Acc{.616} & \Acc{.663} & \Acc{.516} & \Acc{.650} & \Acc{.659} & \Acc{.607} \\
Qwen 2.5 VL 7B & \Acc{.651} & \Acc{.414} & \Acc{.641} & \Acc{.726} & \Acc{.614} & \Acc{.766} & \Acc{.615} & \Acc{.700} & \Acc{.643} & \Acc{.616} & \Acc{.724} & \Acc{.592} & \Acc{.672} & \Acc{.667} & \Acc{.646} \\
Qwen 3 VL 8B & \Acc{.745} & \Acc{.581} & \Acc{.739} & \Acc{.773} & \Acc{.771} & \Acc{.773} & \Acc{.769} & \Acc{.761} & \Acc{.747} & \Acc{.739} & \Acc{.770} & \Acc{.704} & \Acc{.761} & \Acc{.757} & \Acc{.742} \\
Aya Vision 8B & \Acc{.264} & \Acc{.163} & \Acc{.336} & \Acc{.459} & \Acc{.445} & \Acc{.548} & \Acc{.332} & \Acc{.412} & \Acc{.388} & \Acc{.387} & \Acc{.429} & \Acc{.268} & \Acc{.411} & \Acc{.470} & \Acc{.379} \\
Jina VL & \Acc{.023} & \Acc{.002} & \Acc{.008} & \Acc{.037} & \Acc{.018} & \Acc{.021} & \Acc{.016} & \Acc{.031} & \Acc{.026} & \Acc{.016} & \Acc{.016} & \Acc{.009} & \Acc{.017} & \Acc{.036} & \Acc{.020} \\
\midrule
NLLB-SigLIP Base & \Acc{.768} & \Acc{.512} & \Acc{.548} & \Acc{.644} & \Acc{.754} & \Acc{.771} & \Acc{.759} & \Acc{.613} & \Acc{.726} & \Acc{.608} & \Acc{.775} & \Acc{.586} & \Acc{.748} & \Acc{.642} & \Acc{.675} \\
NLLB-SigLIP Large & \Acc{.766} & \Acc{.593} & \Acc{.701} & \Acc{.708} & \Acc{.774} & \Acc{.778} & \Acc{.771} & \Acc{.733} & \Acc{.740} & \Acc{.708} & \Acc{.770} & \Acc{.692} & \Acc{.767} & \Acc{.666} & \Acc{.726} \\
\bottomrule
\end{tabular}
\caption{Accuracy \textit{2img} for bidirectional retrieval, i.e. $c_\textit{2img}$ from Equation~(\ref{eq:c2img}). The upper part of the tables contains LLMs, while the lower has the dual encoders.}
\label{tab:results_1}
\end{table*}

\begin{table*}[htbp]
\centering\footnotesize
\setlength{\tabcolsep}{4.2pt}
\begin{tabular}{lcccccccccccccc@{\hskip 16pt}c}
\toprule
Model & ar & az & cs & de & el & en & hi & it & ja & ro & ru & sk & uk & vi & Avg. \\
\midrule
Gemma 3 12B & \Acc{.737} & \Acc{.630} & \Acc{.705} & \Acc{.751} & \Acc{.733} & \Acc{.767} & \Acc{.724} & \Acc{.742} & \Acc{.711} & \Acc{.750} & \Acc{.760} & \Acc{.704} & \Acc{.742} & \Acc{.747} & \Acc{.729} \\
Qwen 2.5 VL 7B & \Acc{.719} & \Acc{.505} & \Acc{.720} & \Acc{.786} & \Acc{.741} & \Acc{.801} & \Acc{.677} & \Acc{.785} & \Acc{.772} & \Acc{.737} & \Acc{.788} & \Acc{.714} & \Acc{.765} & \Acc{.758} & \Acc{.733} \\
Qwen 3 VL 8B & \Acc{.791} & \Acc{.692} & \Acc{.761} & \Acc{.803} & \Acc{.787} & \Acc{.818} & \Acc{.789} & \Acc{.815} & \Acc{.793} & \Acc{.794} & \Acc{.802} & \Acc{.759} & \Acc{.780} & \Acc{.794} & \Acc{.784} \\
Aya Vision 8B & \Acc{.635} & \Acc{.355} & \Acc{.595} & \Acc{.648} & \Acc{.630} & \Acc{.688} & \Acc{.617} & \Acc{.660} & \Acc{.613} & \Acc{.644} & \Acc{.651} & \Acc{.508} & \Acc{.617} & \Acc{.646} & \Acc{.608} \\
Jina VL & \Acc{.445} & \Acc{.187} & \Acc{.417} & \Acc{.512} & \Acc{.448} & \Acc{.675} & \Acc{.418} & \Acc{.554} & \Acc{.447} & \Acc{.477} & \Acc{.485} & \Acc{.445} & \Acc{.449} & \Acc{.443} & \Acc{.457} \\
\midrule
NLLB-SigLIP Base & \Acc{.756} & \Acc{.463} & \Acc{.488} & \Acc{.605} & \Acc{.763} & \Acc{.782} & \Acc{.750} & \Acc{.626} & \Acc{.742} & \Acc{.578} & \Acc{.775} & \Acc{.522} & \Acc{.759} & \Acc{.659} & \Acc{.662} \\
NLLB-SigLIP Large & \Acc{.765} & \Acc{.567} & \Acc{.672} & \Acc{.735} & \Acc{.779} & \Acc{.792} & \Acc{.784} & \Acc{.750} & \Acc{.759} & \Acc{.720} & \Acc{.779} & \Acc{.669} & \Acc{.764} & \Acc{.685} & \Acc{.730} \\
\bottomrule
\end{tabular}
\caption{Accuracy \textit{2txt} for bidirectional retrieval, i.e. $c_\textit{2txt}$ from Equation~(\ref{eq:c2txt}). The upper part of the tables contains LLMs, while the lower has the dual encoders.}
\label{tab:results_2}
\end{table*}

\begin{table*}[htbp]
\centering\footnotesize
\setlength{\tabcolsep}{4.2pt}
\begin{tabular}{lcccccccccccccc@{\hskip 16pt}c}
\toprule
& ar & az & cs & de & el & en & hi & it & ja & ro & ru & sk & uk & vi & Avg. \\
\midrule
ar & --- & \Acc{.502} & \Acc{.622} & \Acc{.646} & \Acc{.665} & \Acc{.660} & \Acc{.621} & \Acc{.651} & \Acc{.600} & \Acc{.636} & \Acc{.683} & \Acc{.609} & \Acc{.660} & \Acc{.647} & \Acc{.657} \\
az & \Acc{.518} & --- & \Acc{.508} & \Acc{.529} & \Acc{.520} & \Acc{.541} & \Acc{.490} & \Acc{.526} & \Acc{.475} & \Acc{.508} & \Acc{.545} & \Acc{.502} & \Acc{.531} & \Acc{.516} & \Acc{.551} \\
cs & \Acc{.601} & \Acc{.476} & --- & \Acc{.626} & \Acc{.613} & \Acc{.630} & \Acc{.568} & \Acc{.621} & \Acc{.557} & \Acc{.614} & \Acc{.650} & \Acc{.652} & \Acc{.637} & \Acc{.602} & \Acc{.632} \\
de & \Acc{.644} & \Acc{.516} & \Acc{.645} & --- & \Acc{.647} & \Acc{.669} & \Acc{.604} & \Acc{.655} & \Acc{.594} & \Acc{.647} & \Acc{.678} & \Acc{.630} & \Acc{.662} & \Acc{.632} & \Acc{.659} \\
el & \Acc{.663} & \Acc{.506} & \Acc{.628} & \Acc{.643} & --- & \Acc{.680} & \Acc{.624} & \Acc{.654} & \Acc{.602} & \Acc{.645} & \Acc{.684} & \Acc{.617} & \Acc{.672} & \Acc{.653} & \Acc{.662} \\
en & \Acc{.661} & \Acc{.524} & \Acc{.648} & \Acc{.672} & \Acc{.690} & --- & \Acc{.644} & \Acc{.669} & \Acc{.630} & \Acc{.670} & \Acc{.703} & \Acc{.631} & \Acc{.683} & \Acc{.672} & \Acc{.678} \\
hi & \Acc{.639} & \Acc{.489} & \Acc{.598} & \Acc{.616} & \Acc{.643} & \Acc{.652} & --- & \Acc{.620} & \Acc{.591} & \Acc{.612} & \Acc{.657} & \Acc{.583} & \Acc{.639} & \Acc{.626} & \Acc{.640} \\
\bottomrule
\end{tabular}
\caption{Accuracy of the bilingual retrieval, \textit{2txt} only.}
\label{tab:bilingual_acc}
\end{table*}

% ################################################

% Gemma 3 12B

\begin{table*}[htbp]
\begin{minipage}{.50\textwidth}
    
\centering
Gemma 3 12B

\footnotesize
\setlength{\tabcolsep}{2.0pt}
\scalebox{.72}{
\begin{tabular}{lcccccccccccccc@{\hskip 8pt}c}
\toprule
Lng. & ar & az & cs & de & el & en & hi & it & ja & ro & ru & sk & uk & vi & Avg. \\
\midrule
ar & --- & \FP{.14} & \FP{-.12} & \FP{-.15} & \FP{.00} & \FP{-.29} & \FP{.04} & \FP{-.11} & \FP{-.03} & \FP{-.18} & \FP{-.10} & \FP{-.10} & \FP{-.04} & \FP{-.11} & \FP{-.08} \\
az & \FP{-.14} & --- & \FP{-.33} & \FP{-.37} & \FP{-.22} & \FP{-.47} & \FP{-.16} & \FP{-.32} & \FP{-.22} & \FP{-.36} & \FP{-.28} & \FP{-.29} & \FP{-.20} & \FP{-.35} & \FP{-.28} \\
cs & \FP{.12} & \FP{.33} & --- & \FP{-.10} & \FP{.04} & \FP{-.29} & \FP{.19} & \FP{-.02} & \FP{.12} & \FP{-.05} & \FP{-.02} & \FP{.02} & \FP{.06} & \FP{-.07} & \FP{.02} \\
de & \FP{.15} & \FP{.37} & \FP{.10} & --- & \FP{.14} & \FP{-.15} & \FP{.25} & \FP{.09} & \FP{.19} & \FP{.06} & \FP{.04} & \FP{.14} & \FP{.12} & \FP{.04} & \FP{.12} \\
el & \FP{.00} & \FP{.22} & \FP{-.04} & \FP{-.14} & --- & \FP{-.26} & \FP{.04} & \FP{-.09} & \FP{.01} & \FP{-.12} & \FP{-.09} & \FP{-.02} & \FP{-.01} & \FP{-.11} & \FP{-.05} \\
en & \FP{.29} & \FP{.47} & \FP{.29} & \FP{.15} & \FP{.26} & --- & \FP{.37} & \FP{.26} & \FP{.35} & \FP{.24} & \FP{.18} & \FP{.30} & \FP{.24} & \FP{.19} & \FP{.28} \\
hi & \FP{-.04} & \FP{.16} & \FP{-.19} & \FP{-.25} & \FP{-.04} & \FP{-.37} & --- & \FP{-.22} & \FP{-.11} & \FP{-.29} & \FP{-.18} & \FP{-.16} & \FP{-.11} & \FP{-.21} & \FP{-.15} \\
\bottomrule
\end{tabular}
}
\end{minipage}\begin{minipage}{.50\textwidth}
\centering
Qwen 2.5 VL 7B
\footnotesize
\setlength{\tabcolsep}{2.0pt}

\scalebox{.72}{
\begin{tabular}{lcccccccccccccc@{\hskip 8pt}c}
\toprule
Lng. & ar & az & cs & de & el & en & hi & it & ja & ro & ru & sk & uk & vi & Avg. \\
\midrule
ar & --- & \FP{.14} & \FP{-.07} & \FP{-.12} & \FP{-.06} & \FP{-.18} & \FP{.07} & \FP{-.13} & \FP{-.06} & \FP{-.10} & \FP{-.08} & \FP{-.04} & \FP{-.03} & \FP{-.09} & \FP{-.06} \\
az & \FP{-.14} & --- & \FP{-.26} & \FP{-.28} & \FP{-.25} & \FP{-.30} & \FP{-.22} & \FP{-.32} & \FP{-.27} & \FP{-.26} & \FP{-.21} & \FP{-.26} & \FP{-.17} & \FP{-.23} & \FP{-.24} \\
cs & \FP{.07} & \FP{.26} & --- & \FP{-.08} & \FP{-.05} & \FP{-.18} & \FP{.06} & \FP{-.11} & \FP{-.03} & \FP{-.03} & \FP{-.05} & \FP{.05} & \FP{.04} & \FP{-.02} & \FP{-.01} \\
de & \FP{.12} & \FP{.28} & \FP{.08} & --- & \FP{.05} & \FP{-.08} & \FP{.14} & \FP{.00} & \FP{.05} & \FP{.08} & \FP{.04} & \FP{.12} & \FP{.12} & \FP{.05} & \FP{.08} \\
el & \FP{.06} & \FP{.25} & \FP{.05} & \FP{-.05} & --- & \FP{-.12} & \FP{.05} & \FP{-.03} & \FP{.02} & \FP{.00} & \FP{-.02} & \FP{.09} & \FP{.04} & \FP{.00} & \FP{.03} \\
en & \FP{.18} & \FP{.30} & \FP{.18} & \FP{.08} & \FP{.12} & --- & \FP{.17} & \FP{.08} & \FP{.12} & \FP{.15} & \FP{.08} & \FP{.19} & \FP{.14} & \FP{.12} & \FP{.15} \\
hi & \FP{-.07} & \FP{.22} & \FP{-.06} & \FP{-.14} & \FP{-.05} & \FP{-.17} & --- & \FP{-.16} & \FP{-.10} & \FP{-.09} & \FP{-.11} & \FP{-.01} & \FP{-.04} & \FP{-.10} & \FP{-.07} \\
\bottomrule
\end{tabular}
}
\end{minipage}

\begin{minipage}{.50\textwidth}
\centering
Qwen 3 VL 8B
\footnotesize
\setlength{\tabcolsep}{2.0pt}
\scalebox{.72}{
\begin{tabular}{lcccccccccccccc@{\hskip 8pt}c}
\toprule
Lng. & ar & az & cs & de & el & en & hi & it & ja & ro & ru & sk & uk & vi & Avg. \\
\midrule
ar & --- & \FP{.08} & \FP{-.01} & \FP{-.05} & \FP{-.02} & \FP{-.09} & \FP{-.01} & \FP{-.04} & \FP{-.08} & \FP{-.02} & \FP{-.01} & \FP{.01} & \FP{.02} & \FP{-.08} & \FP{-.02} \\
az & \FP{-.08} & --- & \FP{-.15} & \FP{-.14} & \FP{-.12} & \FP{-.17} & \FP{-.13} & \FP{-.14} & \FP{-.18} & \FP{-.11} & \FP{-.08} & \FP{-.12} & \FP{-.05} & \FP{-.15} & \FP{-.12} \\
cs & \FP{.01} & \FP{.15} & --- & \FP{-.05} & \FP{-.02} & \FP{-.10} & \FP{.00} & \FP{-.06} & \FP{-.08} & \FP{-.01} & \FP{-.02} & \FP{.04} & \FP{.03} & \FP{-.06} & \FP{-.01} \\
de & \FP{.05} & \FP{.14} & \FP{.05} & --- & \FP{.02} & \FP{-.08} & \FP{.03} & \FP{.01} & \FP{-.02} & \FP{.04} & \FP{.02} & \FP{.07} & \FP{.06} & \FP{-.02} & \FP{.03} \\
el & \FP{.02} & \FP{.12} & \FP{.02} & \FP{-.02} & --- & \FP{-.05} & \FP{.02} & \FP{-.01} & \FP{-.03} & \FP{.01} & \FP{.01} & \FP{.04} & \FP{.04} & \FP{-.03} & \FP{.01} \\
en & \FP{.09} & \FP{.17} & \FP{.10} & \FP{.08} & \FP{.05} & --- & \FP{.06} & \FP{.08} & \FP{.05} & \FP{.10} & \FP{.07} & \FP{.12} & \FP{.09} & \FP{.03} & \FP{.08} \\
hi & \FP{.01} & \FP{.13} & \FP{.00} & \FP{-.03} & \FP{-.02} & \FP{-.06} & --- & \FP{-.04} & \FP{-.05} & \FP{.01} & \FP{-.01} & \FP{.03} & \FP{.04} & \FP{-.05} & \FP{.00} \\
\bottomrule
\end{tabular}
}
\end{minipage}\begin{minipage}{.50\textwidth}
\centering
Aya Vision 8B
\footnotesize
\setlength{\tabcolsep}{2.0pt}

\scalebox{.72}{
\begin{tabular}{lcccccccccccccc@{\hskip 8pt}c}
\toprule
Lng. & ar & az & cs & de & el & en & hi & it & ja & ro & ru & sk & uk & vi & Avg. \\
\midrule
ar & --- & \FP{.31} & \FP{.11} & \FP{.19} & \FP{-.09} & \FP{.21} & \FP{-.22} & \FP{.08} & \FP{-.13} & \FP{.04} & \FP{-.03} & \FP{.19} & \FP{-.03} & \FP{-.08} & \FP{.04} \\
az & \FP{-.31} & --- & \FP{-.27} & \FP{-.11} & \FP{-.43} & \FP{-.27} & \FP{-.55} & \FP{-.22} & \FP{-.45} & \FP{-.34} & \FP{-.34} & \FP{-.15} & \FP{-.34} & \FP{-.43} & \FP{-.32} \\
cs & \FP{-.11} & \FP{.27} & --- & \FP{.06} & \FP{-.19} & \FP{-.10} & \FP{-.29} & \FP{-.01} & \FP{-.20} & \FP{-.08} & \FP{-.07} & \FP{.12} & \FP{-.07} & \FP{-.19} & \FP{-.07} \\
de & \FP{-.19} & \FP{.11} & \FP{-.06} & --- & \FP{-.22} & \FP{-.22} & \FP{-.36} & \FP{-.07} & \FP{-.26} & \FP{-.13} & \FP{-.14} & \FP{.02} & \FP{-.14} & \FP{-.21} & \FP{-.14} \\
el & \FP{.09} & \FP{.43} & \FP{.19} & \FP{.22} & --- & \FP{.19} & \FP{-.09} & \FP{.14} & \FP{-.01} & \FP{.08} & \FP{.07} & \FP{.31} & \FP{.10} & \FP{.01} & \FP{.13} \\
en & \FP{-.21} & \FP{.27} & \FP{.10} & \FP{.22} & \FP{-.19} & --- & \FP{-.40} & \FP{.07} & \FP{-.06} & \FP{.02} & \FP{-.05} & \FP{.15} & \FP{-.01} & \FP{-.12} & \FP{-.02} \\
hi & \FP{.22} & \FP{.55} & \FP{.29} & \FP{.36} & \FP{.09} & \FP{.40} & --- & \FP{.26} & \FP{.08} & \FP{.20} & \FP{.19} & \FP{.39} & \FP{.19} & \FP{.09} & \FP{.26} \\
\bottomrule
\end{tabular}
}
\end{minipage}

\begin{minipage}{.50\textwidth}
\centering
Jina VL
\footnotesize
\setlength{\tabcolsep}{2.0pt}

\scalebox{.72}{
\begin{tabular}{lcccccccccccccc@{\hskip 8pt}c}
\toprule
Lng. & ar & az & cs & de & el & en & hi & it & ja & ro & ru & sk & uk & vi & Avg. \\
\midrule
ar & --- & \FP{.24} & \FP{.05} & \FP{.00} & \FP{.03} & \FP{-.06} & \FP{-.15} & \FP{-.05} & \FP{-.03} & \FP{-.01} & \FP{.13} & \FP{.02} & \FP{.19} & \FP{.01} & \FP{.03} \\
az & \FP{-.24} & --- & \FP{-.23} & \FP{-.21} & \FP{-.28} & \FP{-.28} & \FP{-.44} & \FP{-.30} & \FP{-.22} & \FP{-.29} & \FP{-.15} & \FP{-.25} & \FP{-.15} & \FP{-.21} & \FP{-.25} \\
cs & \FP{-.05} & \FP{.23} & --- & \FP{-.06} & \FP{-.09} & \FP{-.22} & \FP{-.26} & \FP{-.11} & \FP{-.10} & \FP{-.06} & \FP{-.01} & \FP{-.01} & \FP{.02} & \FP{-.08} & \FP{-.06} \\
de & \FP{.00} & \FP{.21} & \FP{.06} & --- & \FP{.01} & \FP{-.16} & \FP{-.19} & \FP{-.08} & \FP{-.08} & \FP{.00} & \FP{.14} & \FP{.03} & \FP{.16} & \FP{-.09} & \FP{.00} \\
el & \FP{-.03} & \FP{.28} & \FP{.09} & \FP{-.01} & --- & \FP{-.34} & \FP{-.28} & \FP{-.05} & \FP{-.07} & \FP{-.02} & \FP{.04} & \FP{.08} & \FP{.08} & \FP{-.07} & \FP{-.02} \\
en & \FP{.06} & \FP{.28} & \FP{.22} & \FP{.16} & \FP{.34} & --- & \FP{.12} & \FP{.10} & \FP{.00} & \FP{.12} & \FP{.17} & \FP{.21} & \FP{.24} & \FP{.03} & \FP{.16} \\
hi & \FP{.15} & \FP{.44} & \FP{.26} & \FP{.19} & \FP{.28} & \FP{-.12} & --- & \FP{.16} & \FP{.13} & \FP{.15} & \FP{.30} & \FP{.23} & \FP{.36} & \FP{.04} & \FP{.20} \\
\bottomrule
\end{tabular}
}
\end{minipage}

\caption{The difference between the language-specific mistakes, for decoder-only model.}
\label{tab:bilingual_fp_llms}
\end{table*}

% NLLB-SigLIP and SigLIP2 Models

\begin{table*}[htbp]
\begin{minipage}{.50\textwidth}
\centering
NLLB-SigLIP Base
\footnotesize
\setlength{\tabcolsep}{2.0pt}
\scalebox{.72}{
\begin{tabular}{lcccccccccccccc@{\hskip 8pt}c}
\toprule
Lng. & ar & az & cs & de & el & en & hi & it & ja & ro & ru & sk & uk & vi & Avg. \\
\midrule
ar & --- & \FP{.43} & \FP{.40} & \FP{.23} & \FP{-.01} & \FP{-.03} & \FP{-.01} & \FP{.29} & \FP{.01} & \FP{.18} & \FP{-.01} & \FP{.46} & \FP{.00} & \FP{.09} & \FP{.16} \\
az & \FP{-.43} & --- & \FP{-.06} & \FP{-.21} & \FP{-.47} & \FP{-.49} & \FP{-.45} & \FP{-.19} & \FP{-.46} & \FP{-.33} & \FP{-.46} & \FP{.01} & \FP{-.44} & \FP{-.41} & \FP{-.34} \\
cs & \FP{-.40} & \FP{.06} & --- & \FP{-.17} & \FP{-.41} & \FP{-.43} & \FP{-.42} & \FP{-.13} & \FP{-.40} & \FP{-.27} & \FP{-.42} & \FP{.09} & \FP{-.41} & \FP{-.39} & \FP{-.28} \\
de & \FP{-.23} & \FP{.21} & \FP{.17} & --- & \FP{-.26} & \FP{-.29} & \FP{-.25} & \FP{.05} & \FP{-.26} & \FP{-.10} & \FP{-.26} & \FP{.24} & \FP{-.24} & \FP{-.20} & \FP{-.11} \\
el & \FP{.01} & \FP{.47} & \FP{.41} & \FP{.26} & --- & \FP{-.02} & \FP{.01} & \FP{.31} & \FP{.04} & \FP{.21} & \FP{.01} & \FP{.47} & \FP{.03} & \FP{.11} & \FP{.18} \\
en & \FP{.03} & \FP{.49} & \FP{.43} & \FP{.29} & \FP{.02} & --- & \FP{.02} & \FP{.33} & \FP{.05} & \FP{.24} & \FP{.02} & \FP{.48} & \FP{.06} & \FP{.12} & \FP{.20} \\
hi & \FP{.01} & \FP{.45} & \FP{.42} & \FP{.25} & \FP{-.01} & \FP{-.02} & --- & \FP{.31} & \FP{.02} & \FP{.20} & \FP{.01} & \FP{.47} & \FP{.01} & \FP{.09} & \FP{.17} \\
\bottomrule
\end{tabular}
}
\end{minipage}\begin{minipage}{.50\textwidth}
\centering
NLLB-SigLIP Large
\footnotesize
\setlength{\tabcolsep}{2.0pt}
\scalebox{.72}{
\begin{tabular}{lcccccccccccccc@{\hskip 8pt}c}
\toprule
Lng. & ar & az & cs & de & el & en & hi & it & ja & ro & ru & sk & uk & vi & Avg. \\
\midrule
ar & --- & \FP{.21} & \FP{.16} & \FP{.05} & \FP{-.02} & \FP{-.05} & \FP{.01} & \FP{.02} & \FP{.01} & \FP{.04} & \FP{-.01} & \FP{.15} & \FP{.02} & \FP{.10} & \FP{.05} \\
az & \FP{-.21} & --- & \FP{-.05} & \FP{-.18} & \FP{-.23} & \FP{-.25} & \FP{-.21} & \FP{-.18} & \FP{-.21} & \FP{-.17} & \FP{-.20} & \FP{-.06} & \FP{-.19} & \FP{-.13} & \FP{-.17} \\
cs & \FP{-.16} & \FP{.05} & --- & \FP{-.12} & \FP{-.18} & \FP{-.20} & \FP{-.17} & \FP{-.12} & \FP{-.16} & \FP{-.12} & \FP{-.17} & \FP{.00} & \FP{-.14} & \FP{-.07} & \FP{-.12} \\
de & \FP{-.05} & \FP{.18} & \FP{.12} & --- & \FP{-.06} & \FP{-.08} & \FP{-.05} & \FP{.00} & \FP{-.04} & \FP{.00} & \FP{-.06} & \FP{.11} & \FP{-.02} & \FP{.05} & \FP{.01} \\
el & \FP{.02} & \FP{.23} & \FP{.18} & \FP{.06} & --- & \FP{-.03} & \FP{.02} & \FP{.05} & \FP{.02} & \FP{.07} & \FP{.03} & \FP{.17} & \FP{.04} & \FP{.12} & \FP{.07} \\
en & \FP{.05} & \FP{.25} & \FP{.20} & \FP{.08} & \FP{.03} & --- & \FP{.04} & \FP{.07} & \FP{.05} & \FP{.08} & \FP{.04} & \FP{.18} & \FP{.07} & \FP{.13} & \FP{.10} \\
hi & \FP{-.01} & \FP{.21} & \FP{.17} & \FP{.05} & \FP{-.02} & \FP{-.04} & --- & \FP{.03} & \FP{.01} & \FP{.04} & \FP{-.01} & \FP{.14} & \FP{.02} & \FP{.09} & \FP{.05} \\
\bottomrule
\end{tabular}
}
\end{minipage}

\begin{minipage}{.50\textwidth}
\centering
MEXMA-SigLIP2
\footnotesize
\setlength{\tabcolsep}{2.0pt}
\scalebox{.72}{
\begin{tabular}{lcccccccccccccc@{\hskip 8pt}c}
\toprule
Lng. & ar & az & cs & de & el & en & hi & it & ja & ro & ru & sk & uk & vi & Avg. \\
\midrule
ar & --- & \FP{.04} & \FP{.00} & \FP{-.01} & \FP{.01} & \FP{-.04} & \FP{.00} & \FP{-.03} & \FP{.08} & \FP{.01} & \FP{.00} & \FP{-.02} & \FP{.00} & \FP{.05} & \FP{.01} \\
az & \FP{-.04} & --- & \FP{-.04} & \FP{-.05} & \FP{-.04} & \FP{-.08} & \FP{-.02} & \FP{-.05} & \FP{.02} & \FP{-.04} & \FP{-.03} & \FP{-.04} & \FP{-.04} & \FP{.00} & \FP{-.03} \\
cs & \FP{.00} & \FP{.04} & --- & \FP{-.01} & \FP{.01} & \FP{-.04} & \FP{.02} & \FP{-.02} & \FP{.07} & \FP{.00} & \FP{.00} & \FP{.00} & \FP{.00} & \FP{.03} & \FP{.01} \\
de & \FP{.01} & \FP{.05} & \FP{.01} & --- & \FP{.02} & \FP{-.02} & \FP{.02} & \FP{.00} & \FP{.07} & \FP{.02} & \FP{.02} & \FP{.01} & \FP{.02} & \FP{.05} & \FP{.02} \\
el & \FP{-.01} & \FP{.04} & \FP{-.01} & \FP{-.02} & --- & \FP{-.05} & \FP{-.01} & \FP{-.02} & \FP{.07} & \FP{-.01} & \FP{.00} & \FP{-.01} & \FP{.00} & \FP{.03} & \FP{.00} \\
en & \FP{.04} & \FP{.08} & \FP{.04} & \FP{.02} & \FP{.05} & --- & \FP{.05} & \FP{.02} & \FP{.10} & \FP{.04} & \FP{.05} & \FP{.04} & \FP{.04} & \FP{.07} & \FP{.05} \\
hi & \FP{.00} & \FP{.02} & \FP{-.02} & \FP{-.02} & \FP{.01} & \FP{-.05} & --- & \FP{-.02} & \FP{.07} & \FP{.00} & \FP{.00} & \FP{-.01} & \FP{.00} & \FP{.03} & \FP{.00} \\
\bottomrule
\end{tabular}
}
\end{minipage}\begin{minipage}{.50\textwidth}
\centering
SigLIP2 Base
\footnotesize
\setlength{\tabcolsep}{2.0pt}
\scalebox{.72}{
\begin{tabular}{lcccccccccccccc@{\hskip 8pt}c}
\toprule
Lng. & ar & az & cs & de & el & en & hi & it & ja & ro & ru & sk & uk & vi & Avg. \\
\midrule
ar & --- & \FP{.12} & \FP{-.02} & \FP{-.02} & \FP{.01} & \FP{-.04} & \FP{.05} & \FP{-.02} & \FP{.07} & \FP{.02} & \FP{-.03} & \FP{-.01} & \FP{.00} & \FP{.00} & \FP{.01} \\
az & \FP{-.12} & --- & \FP{-.13} & \FP{-.14} & \FP{-.11} & \FP{-.16} & \FP{-.07} & \FP{-.13} & \FP{-.04} & \FP{-.09} & \FP{-.14} & \FP{-.12} & \FP{-.12} & \FP{-.11} & \FP{-.11} \\
cs & \FP{.02} & \FP{.13} & --- & \FP{-.01} & \FP{.03} & \FP{-.02} & \FP{.06} & \FP{.00} & \FP{.09} & \FP{.04} & \FP{-.01} & \FP{.01} & \FP{.02} & \FP{.02} & \FP{.03} \\
de & \FP{.02} & \FP{.14} & \FP{.01} & --- & \FP{.03} & \FP{-.02} & \FP{.07} & \FP{.00} & \FP{.10} & \FP{.05} & \FP{.00} & \FP{.02} & \FP{.02} & \FP{.02} & \FP{.04} \\
el & \FP{-.01} & \FP{.11} & \FP{-.03} & \FP{-.03} & --- & \FP{-.05} & \FP{.04} & \FP{-.03} & \FP{.06} & \FP{.01} & \FP{-.04} & \FP{-.02} & \FP{-.01} & \FP{-.01} & \FP{.00} \\
en & \FP{.04} & \FP{.16} & \FP{.02} & \FP{.02} & \FP{.05} & --- & \FP{.09} & \FP{.02} & \FP{.11} & \FP{.06} & \FP{.01} & \FP{.03} & \FP{.04} & \FP{.04} & \FP{.05} \\
hi & \FP{-.05} & \FP{.07} & \FP{-.06} & \FP{-.07} & \FP{-.04} & \FP{-.09} & --- & \FP{-.07} & \FP{.03} & \FP{-.03} & \FP{-.07} & \FP{-.05} & \FP{-.05} & \FP{-.05} & \FP{-.04} \\
\bottomrule
\end{tabular}
}
\end{minipage}

\begin{minipage}{.50\textwidth}
\centering
SigLIP2 Large
\footnotesize
\setlength{\tabcolsep}{2.0pt}
\scalebox{.72}{
\begin{tabular}{lcccccccccccccc@{\hskip 8pt}c}
\toprule
Lng. & ar & az & cs & de & el & en & hi & it & ja & ro & ru & sk & uk & vi & Avg. \\
\midrule
ar & --- & \FP{.12} & \FP{.00} & \FP{-.02} & \FP{.00} & \FP{-.03} & \FP{.01} & \FP{-.02} & \FP{.06} & \FP{.00} & \FP{-.02} & \FP{.01} & \FP{.00} & \FP{-.01} & \FP{.01} \\
az & \FP{-.12} & --- & \FP{-.13} & \FP{-.14} & \FP{-.12} & \FP{-.15} & \FP{-.11} & \FP{-.14} & \FP{-.06} & \FP{-.12} & \FP{-.14} & \FP{-.11} & \FP{-.13} & \FP{-.13} & \FP{-.12} \\
cs & \FP{.00} & \FP{.13} & --- & \FP{-.01} & \FP{.01} & \FP{-.02} & \FP{.02} & \FP{-.01} & \FP{.07} & \FP{.01} & \FP{-.01} & \FP{.01} & \FP{.00} & \FP{.00} & \FP{.01} \\
de & \FP{.02} & \FP{.14} & \FP{.01} & --- & \FP{.02} & \FP{-.01} & \FP{.03} & \FP{.00} & \FP{.08} & \FP{.02} & \FP{.00} & \FP{.02} & \FP{.01} & \FP{.01} & \FP{.03} \\
el & \FP{.00} & \FP{.12} & \FP{-.01} & \FP{-.02} & --- & \FP{-.03} & \FP{.01} & \FP{-.02} & \FP{.06} & \FP{.00} & \FP{-.02} & \FP{.00} & \FP{-.01} & \FP{-.01} & \FP{.00} \\
en & \FP{.03} & \FP{.15} & \FP{.02} & \FP{.01} & \FP{.03} & --- & \FP{.04} & \FP{.01} & \FP{.09} & \FP{.03} & \FP{.01} & \FP{.03} & \FP{.02} & \FP{.02} & \FP{.04} \\
hi & \FP{-.01} & \FP{.11} & \FP{-.02} & \FP{-.03} & \FP{-.01} & \FP{-.04} & --- & \FP{-.03} & \FP{.05} & \FP{-.01} & \FP{-.03} & \FP{-.01} & \FP{-.02} & \FP{-.02} & \FP{.00} \\
\bottomrule
\end{tabular}
}
\end{minipage}\begin{minipage}{.50\textwidth}
\centering
SigLIP2 SO400M
\footnotesize
\setlength{\tabcolsep}{2.0pt}
\scalebox{.72}{
\begin{tabular}{lcccccccccccccc@{\hskip 8pt}c}
\toprule
Lng. & ar & az & cs & de & el & en & hi & it & ja & ro & ru & sk & uk & vi & Avg. \\
\midrule
ar & --- & \FP{.09} & \FP{-.02} & \FP{-.02} & \FP{-.01} & \FP{-.02} & \FP{.00} & \FP{-.01} & \FP{.05} & \FP{-.01} & \FP{-.01} & \FP{-.01} & \FP{-.01} & \FP{-.01} & \FP{.00} \\
az & \FP{-.09} & --- & \FP{-.11} & \FP{-.11} & \FP{-.10} & \FP{-.12} & \FP{-.09} & \FP{-.10} & \FP{-.04} & \FP{-.10} & \FP{-.10} & \FP{-.10} & \FP{-.10} & \FP{-.10} & \FP{-.10} \\
cs & \FP{.02} & \FP{.11} & --- & \FP{-.01} & \FP{.01} & \FP{-.01} & \FP{.01} & \FP{.00} & \FP{.07} & \FP{.01} & \FP{.00} & \FP{.01} & \FP{.01} & \FP{.01} & \FP{.02} \\
de & \FP{.02} & \FP{.11} & \FP{.01} & --- & \FP{.01} & \FP{.00} & \FP{.02} & \FP{.01} & \FP{.08} & \FP{.01} & \FP{.01} & \FP{.02} & \FP{.01} & \FP{.02} & \FP{.02} \\
el & \FP{.01} & \FP{.10} & \FP{-.01} & \FP{-.01} & --- & \FP{-.02} & \FP{.01} & \FP{.00} & \FP{.06} & \FP{.00} & \FP{-.01} & \FP{.00} & \FP{.00} & \FP{.00} & \FP{.01} \\
en & \FP{.02} & \FP{.12} & \FP{.01} & \FP{.00} & \FP{.02} & --- & \FP{.02} & \FP{.01} & \FP{.08} & \FP{.01} & \FP{.01} & \FP{.02} & \FP{.01} & \FP{.02} & \FP{.03} \\
hi & \FP{.00} & \FP{.09} & \FP{-.01} & \FP{-.02} & \FP{-.01} & \FP{-.02} & --- & \FP{-.01} & \FP{.06} & \FP{-.01} & \FP{-.01} & \FP{.00} & \FP{-.01} & \FP{.00} & \FP{.00} \\
\bottomrule
\end{tabular}
}
\end{minipage}

\begin{minipage}{.50\textwidth}
\centering
SigLIP2 Giant
\footnotesize
\setlength{\tabcolsep}{2.0pt}
\scalebox{.72}{
\begin{tabular}{lcccccccccccccc@{\hskip 8pt}c}
\toprule
Lng. & ar & az & cs & de & el & en & hi & it & ja & ro & ru & sk & uk & vi & Avg. \\
\midrule
ar & --- & \FP{.11} & \FP{-.02} & \FP{-.02} & \FP{-.01} & \FP{-.02} & \FP{.02} & \FP{-.02} & \FP{.06} & \FP{-.02} & \FP{-.02} & \FP{-.02} & \FP{-.01} & \FP{-.01} & \FP{.00} \\
az & \FP{-.11} & --- & \FP{-.13} & \FP{-.13} & \FP{-.12} & \FP{-.13} & \FP{-.10} & \FP{-.13} & \FP{-.06} & \FP{-.13} & \FP{-.13} & \FP{-.13} & \FP{-.13} & \FP{-.13} & \FP{-.12} \\
cs & \FP{.02} & \FP{.13} & --- & \FP{.00} & \FP{.01} & \FP{.00} & \FP{.04} & \FP{.00} & \FP{.08} & \FP{.00} & \FP{.00} & \FP{.00} & \FP{.01} & \FP{.00} & \FP{.02} \\
de & \FP{.02} & \FP{.13} & \FP{.00} & --- & \FP{.01} & \FP{.00} & \FP{.03} & \FP{.00} & \FP{.08} & \FP{.00} & \FP{.00} & \FP{.00} & \FP{.00} & \FP{.00} & \FP{.02} \\
el & \FP{.01} & \FP{.12} & \FP{-.01} & \FP{-.01} & --- & \FP{-.01} & \FP{.03} & \FP{-.01} & \FP{.07} & \FP{.00} & \FP{-.01} & \FP{-.01} & \FP{.00} & \FP{.00} & \FP{.01} \\
en & \FP{.02} & \FP{.13} & \FP{.00} & \FP{.00} & \FP{.01} & --- & \FP{.04} & \FP{.00} & \FP{.08} & \FP{.00} & \FP{.00} & \FP{.00} & \FP{.01} & \FP{.00} & \FP{.02} \\
hi & \FP{-.02} & \FP{.10} & \FP{-.04} & \FP{-.03} & \FP{-.03} & \FP{-.04} & --- & \FP{-.03} & \FP{.04} & \FP{-.03} & \FP{-.04} & \FP{-.04} & \FP{-.03} & \FP{-.03} & \FP{-.02} \\
\bottomrule
\end{tabular}
}
\end{minipage}

\caption{The difference between the language-specific mistakes, for encoder-only models.}
\label{tab:bilingual_fp_encoders}
\end{table*}

% ################################################

\begin{table*}[htbp]
\begin{minipage}{.333\textwidth}
    
\centering
\phantom{Q}Gemma 3 12B

\footnotesize
\setlength{\tabcolsep}{2.0pt}
\scalebox{.62}{
\begin{tabular}{lcccccccccccccc}
\toprule
& ar & az & cs & de & el & en & hi & it & ja & ro & ru & sk & uk & vi \\
\midrule
ar & --- & \Agree{.70} & \Agree{.75} & \Agree{.78} & \Agree{.80} & \Agree{.78} & \Agree{.75} & \Agree{.80} & \Agree{.76} & \Agree{.78} & \Agree{.81} & \Agree{.71} & \Agree{.80} & \Agree{.81} \\
az & \Agree{.80} & --- & \Agree{.67} & \Agree{.70} & \Agree{.69} & \Agree{.69} & \Agree{.67} & \Agree{.69} & \Agree{.69} & \Agree{.68} & \Agree{.71} & \Agree{.65} & \Agree{.70} & \Agree{.69} \\
cs & \Agree{.82} & \Agree{.76} & --- & \Agree{.75} & \Agree{.74} & \Agree{.72} & \Agree{.69} & \Agree{.73} & \Agree{.71} & \Agree{.74} & \Agree{.76} & \Agree{.78} & \Agree{.77} & \Agree{.73} \\
de & \Agree{.89} & \Agree{.80} & \Agree{.83} & --- & \Agree{.81} & \Agree{.81} & \Agree{.78} & \Agree{.80} & \Agree{.76} & \Agree{.78} & \Agree{.83} & \Agree{.71} & \Agree{.81} & \Agree{.80} \\
el & \Agree{.87} & \Agree{.80} & \Agree{.83} & \Agree{.89} & --- & \Agree{.80} & \Agree{.78} & \Agree{.81} & \Agree{.74} & \Agree{.79} & \Agree{.82} & \Agree{.72} & \Agree{.81} & \Agree{.80} \\
en & \Agree{.88} & \Agree{.80} & \Agree{.82} & \Agree{.90} & \Agree{.89} & --- & \Agree{.78} & \Agree{.80} & \Agree{.75} & \Agree{.77} & \Agree{.80} & \Agree{.69} & \Agree{.79} & \Agree{.80} \\
hi & \Agree{.85} & \Agree{.80} & \Agree{.81} & \Agree{.87} & \Agree{.86} & \Agree{.87} & --- & \Agree{.75} & \Agree{.73} & \Agree{.73} & \Agree{.78} & \Agree{.65} & \Agree{.76} & \Agree{.76} \\
\bottomrule
\end{tabular}
}
\end{minipage}\begin{minipage}{.333\textwidth}
\centering
Qwen 2.5 VL 7B

\footnotesize
\setlength{\tabcolsep}{2.0pt}

\scalebox{.62}{
\begin{tabular}{lcccccccccccccc}
\toprule
& ar & az & cs & de & el & en & hi & it & ja & ro & ru & sk & uk & vi \\
\midrule
ar & --- & \Agree{.63} & \Agree{.79} & \Agree{.84} & \Agree{.79} & \Agree{.83} & \Agree{.78} & \Agree{.84} & \Agree{.81} & \Agree{.79} & \Agree{.85} & \Agree{.78} & \Agree{.82} & \Agree{.84} \\
az & \Agree{.69} & --- & \Agree{.64} & \Agree{.64} & \Agree{.62} & \Agree{.61} & \Agree{.64} & \Agree{.65} & \Agree{.64} & \Agree{.63} & \Agree{.65} & \Agree{.64} & \Agree{.65} & \Agree{.63} \\
cs & \Agree{.83} & \Agree{.68} & --- & \Agree{.85} & \Agree{.78} & \Agree{.81} & \Agree{.77} & \Agree{.84} & \Agree{.81} & \Agree{.80} & \Agree{.85} & \Agree{.85} & \Agree{.85} & \Agree{.80} \\
de & \Agree{.88} & \Agree{.70} & \Agree{.87} & --- & \Agree{.81} & \Agree{.89} & \Agree{.81} & \Agree{.89} & \Agree{.84} & \Agree{.82} & \Agree{.92} & \Agree{.81} & \Agree{.88} & \Agree{.85} \\
el & \Agree{.86} & \Agree{.69} & \Agree{.84} & \Agree{.90} & --- & \Agree{.79} & \Agree{.77} & \Agree{.82} & \Agree{.79} & \Agree{.76} & \Agree{.82} & \Agree{.75} & \Agree{.79} & \Agree{.79} \\
en & \Agree{.87} & \Agree{.68} & \Agree{.86} & \Agree{.94} & \Agree{.89} & --- & \Agree{.80} & \Agree{.88} & \Agree{.83} & \Agree{.80} & \Agree{.89} & \Agree{.79} & \Agree{.84} & \Agree{.84} \\
hi & \Agree{.81} & \Agree{.69} & \Agree{.81} & \Agree{.83} & \Agree{.82} & \Agree{.83} & --- & \Agree{.80} & \Agree{.78} & \Agree{.75} & \Agree{.81} & \Agree{.75} & \Agree{.80} & \Agree{.77} \\
\bottomrule
\end{tabular}
}
\end{minipage}\begin{minipage}{.333\textwidth}
\centering
Qwen 3 VL 8B

\footnotesize
\setlength{\tabcolsep}{2.0pt}
\scalebox{.62}{
\begin{tabular}{lcccccccccccccc}
\toprule
& ar & az & cs & de & el & en & hi & it & ja & ro & ru & sk & uk & vi \\
\midrule
ar & --- & \Agree{.73} & \Agree{.84} & \Agree{.89} & \Agree{.88} & \Agree{.89} & \Agree{.87} & \Agree{.88} & \Agree{.87} & \Agree{.85} & \Agree{.88} & \Agree{.84} & \Agree{.88} & \Agree{.88} \\
az & \Agree{.82} & --- & \Agree{.73} & \Agree{.76} & \Agree{.74} & \Agree{.75} & \Agree{.76} & \Agree{.75} & \Agree{.74} & \Agree{.74} & \Agree{.75} & \Agree{.73} & \Agree{.75} & \Agree{.74} \\
cs & \Agree{.87} & \Agree{.79} & --- & \Agree{.87} & \Agree{.86} & \Agree{.86} & \Agree{.86} & \Agree{.85} & \Agree{.84} & \Agree{.83} & \Agree{.87} & \Agree{.87} & \Agree{.87} & \Agree{.85} \\
de & \Agree{.91} & \Agree{.83} & \Agree{.87} & --- & \Agree{.89} & \Agree{.92} & \Agree{.90} & \Agree{.90} & \Agree{.88} & \Agree{.87} & \Agree{.91} & \Agree{.86} & \Agree{.90} & \Agree{.89} \\
el & \Agree{.90} & \Agree{.81} & \Agree{.84} & \Agree{.90} & --- & \Agree{.88} & \Agree{.89} & \Agree{.87} & \Agree{.86} & \Agree{.86} & \Agree{.89} & \Agree{.85} & \Agree{.88} & \Agree{.87} \\
en & \Agree{.92} & \Agree{.83} & \Agree{.88} & \Agree{.94} & \Agree{.91} & --- & \Agree{.89} & \Agree{.90} & \Agree{.88} & \Agree{.86} & \Agree{.89} & \Agree{.85} & \Agree{.89} & \Agree{.89} \\
hi & \Agree{.90} & \Agree{.82} & \Agree{.86} & \Agree{.90} & \Agree{.88} & \Agree{.91} & --- & \Agree{.88} & \Agree{.87} & \Agree{.86} & \Agree{.89} & \Agree{.85} & \Agree{.89} & \Agree{.88} \\
\bottomrule
\end{tabular}
}
\end{minipage}

\begin{minipage}{.33\textwidth}
\centering
Aya Vision 8B
\footnotesize
\setlength{\tabcolsep}{2.0pt}

\scalebox{.62}{
\begin{tabular}{lcccccccccccccc}
\toprule
& ar & az & cs & de & el & en & hi & it & ja & ro & ru & sk & uk & vi \\
\midrule
ar & --- & \Agree{.40} & \Agree{.59} & \Agree{.54} & \Agree{.64} & \Agree{.48} & \Agree{.65} & \Agree{.58} & \Agree{.58} & \Agree{.63} & \Agree{.61} & \Agree{.56} & \Agree{.62} & \Agree{.62} \\
az & \Agree{.54} & --- & \Agree{.44} & \Agree{.43} & \Agree{.45} & \Agree{.40} & \Agree{.45} & \Agree{.43} & \Agree{.45} & \Agree{.44} & \Agree{.45} & \Agree{.47} & \Agree{.44} & \Agree{.44} \\
cs & \Agree{.72} & \Agree{.53} & --- & \Agree{.66} & \Agree{.69} & \Agree{.62} & \Agree{.64} & \Agree{.69} & \Agree{.66} & \Agree{.71} & \Agree{.72} & \Agree{.68} & \Agree{.72} & \Agree{.67} \\
de & \Agree{.77} & \Agree{.55} & \Agree{.77} & --- & \Agree{.72} & \Agree{.73} & \Agree{.61} & \Agree{.75} & \Agree{.67} & \Agree{.69} & \Agree{.73} & \Agree{.57} & \Agree{.73} & \Agree{.72} \\
el & \Agree{.78} & \Agree{.54} & \Agree{.74} & \Agree{.82} & --- & \Agree{.71} & \Agree{.71} & \Agree{.74} & \Agree{.71} & \Agree{.74} & \Agree{.77} & \Agree{.62} & \Agree{.77} & \Agree{.78} \\
en & \Agree{.79} & \Agree{.53} & \Agree{.76} & \Agree{.84} & \Agree{.84} & --- & \Agree{.58} & \Agree{.71} & \Agree{.66} & \Agree{.65} & \Agree{.67} & \Agree{.53} & \Agree{.68} & \Agree{.72} \\
hi & \Agree{.75} & \Agree{.53} & \Agree{.73} & \Agree{.78} & \Agree{.77} & \Agree{.80} & --- & \Agree{.64} & \Agree{.67} & \Agree{.67} & \Agree{.68} & \Agree{.59} & \Agree{.68} & \Agree{.68} \\
\bottomrule
\end{tabular}
}
\end{minipage}\begin{minipage}{.33\textwidth}
\centering
Jina VL
\footnotesize
\setlength{\tabcolsep}{2.0pt}

\scalebox{.62}{
\begin{tabular}{lcccccccccccccc}
\toprule
& ar & az & cs & de & el & en & hi & it & ja & ro & ru & sk & uk & vi \\
\midrule
ar & --- & \Agree{.26} & \Agree{.54} & \Agree{.58} & \Agree{.43} & \Agree{.57} & \Agree{.45} & \Agree{.61} & \Agree{.48} & \Agree{.56} & \Agree{.56} & \Agree{.55} & \Agree{.55} & \Agree{.47} \\
az & \Agree{.46} & --- & \Agree{.30} & \Agree{.24} & \Agree{.15} & \Agree{.26} & \Agree{.16} & \Agree{.23} & \Agree{.16} & \Agree{.29} & \Agree{.33} & \Agree{.34} & \Agree{.31} & \Agree{.16} \\
cs & \Agree{.65} & \Agree{.43} & --- & \Agree{.53} & \Agree{.36} & \Agree{.55} & \Agree{.38} & \Agree{.53} & \Agree{.39} & \Agree{.55} & \Agree{.58} & \Agree{.68} & \Agree{.55} & \Agree{.40} \\
de & \Agree{.75} & \Agree{.48} & \Agree{.70} & --- & \Agree{.44} & \Agree{.61} & \Agree{.48} & \Agree{.61} & \Agree{.50} & \Agree{.56} & \Agree{.62} & \Agree{.53} & \Agree{.57} & \Agree{.53} \\
el & \Agree{.70} & \Agree{.41} & \Agree{.64} & \Agree{.72} & --- & \Agree{.38} & \Agree{.54} & \Agree{.44} & \Agree{.45} & \Agree{.39} & \Agree{.39} & \Agree{.35} & \Agree{.42} & \Agree{.48} \\
en & \Agree{.72} & \Agree{.41} & \Agree{.65} & \Agree{.80} & \Agree{.71} & --- & \Agree{.44} & \Agree{.61} & \Agree{.49} & \Agree{.57} & \Agree{.58} & \Agree{.56} & \Agree{.55} & \Agree{.48} \\
hi & \Agree{.69} & \Agree{.46} & \Agree{.63} & \Agree{.71} & \Agree{.67} & \Agree{.69} & --- & \Agree{.46} & \Agree{.52} & \Agree{.42} & \Agree{.39} & \Agree{.36} & \Agree{.43} & \Agree{.48} \\
\bottomrule
\end{tabular}
}
\end{minipage}\begin{minipage}{.33\textwidth}
\centering
NLLB-SigLIP Base
\footnotesize
\setlength{\tabcolsep}{2.0pt}
\scalebox{.62}{
\begin{tabular}{lcccccccccccccc}
\toprule
& ar & az & cs & de & el & en & hi & it & ja & ro & ru & sk & uk & vi \\
\midrule
ar & --- & \Agree{.57} & \Agree{.58} & \Agree{.70} & \Agree{.85} & \Agree{.90} & \Agree{.88} & \Agree{.68} & \Agree{.86} & \Agree{.66} & \Agree{.88} & \Agree{.61} & \Agree{.85} & \Agree{.72} \\
az & \Agree{.52} & --- & \Agree{.51} & \Agree{.53} & \Agree{.58} & \Agree{.57} & \Agree{.57} & \Agree{.54} & \Agree{.56} & \Agree{.55} & \Agree{.57} & \Agree{.55} & \Agree{.57} & \Agree{.49} \\
cs & \Agree{.49} & \Agree{.39} & --- & \Agree{.58} & \Agree{.60} & \Agree{.61} & \Agree{.60} & \Agree{.58} & \Agree{.57} & \Agree{.60} & \Agree{.59} & \Agree{.68} & \Agree{.59} & \Agree{.51} \\
de & \Agree{.66} & \Agree{.43} & \Agree{.40} & --- & \Agree{.71} & \Agree{.73} & \Agree{.72} & \Agree{.66} & \Agree{.70} & \Agree{.62} & \Agree{.72} & \Agree{.59} & \Agree{.71} & \Agree{.58} \\
el & \Agree{.85} & \Agree{.50} & \Agree{.48} & \Agree{.66} & --- & \Agree{.89} & \Agree{.87} & \Agree{.70} & \Agree{.84} & \Agree{.68} & \Agree{.88} & \Agree{.62} & \Agree{.84} & \Agree{.72} \\
en & \Agree{.88} & \Agree{.52} & \Agree{.50} & \Agree{.67} & \Agree{.88} & --- & \Agree{.93} & \Agree{.71} & \Agree{.87} & \Agree{.70} & \Agree{.91} & \Agree{.64} & \Agree{.89} & \Agree{.73} \\
hi & \Agree{.84} & \Agree{.51} & \Agree{.48} & \Agree{.64} & \Agree{.84} & \Agree{.89} & --- & \Agree{.69} & \Agree{.86} & \Agree{.69} & \Agree{.91} & \Agree{.63} & \Agree{.88} & \Agree{.72} \\
\bottomrule
\end{tabular}
}
\end{minipage}

\begin{minipage}{.33\textwidth}
\centering
NLLB-SigLIP Large
\footnotesize
\setlength{\tabcolsep}{2.0pt}
\scalebox{.62}{
\begin{tabular}{lcccccccccccccc}
\toprule
& ar & az & cs & de & el & en & hi & it & ja & ro & ru & sk & uk & vi \\
\midrule
ar & --- & \Agree{.67} & \Agree{.76} & \Agree{.82} & \Agree{.88} & \Agree{.91} & \Agree{.90} & \Agree{.85} & \Agree{.86} & \Agree{.81} & \Agree{.89} & \Agree{.79} & \Agree{.87} & \Agree{.77} \\
az & \Agree{.61} & --- & \Agree{.65} & \Agree{.63} & \Agree{.69} & \Agree{.69} & \Agree{.70} & \Agree{.68} & \Agree{.63} & \Agree{.65} & \Agree{.68} & \Agree{.69} & \Agree{.68} & \Agree{.57} \\
cs & \Agree{.69} & \Agree{.56} & --- & \Agree{.74} & \Agree{.79} & \Agree{.78} & \Agree{.80} & \Agree{.77} & \Agree{.74} & \Agree{.76} & \Agree{.79} & \Agree{.80} & \Agree{.79} & \Agree{.69} \\
de & \Agree{.78} & \Agree{.57} & \Agree{.68} & --- & \Agree{.82} & \Agree{.85} & \Agree{.85} & \Agree{.80} & \Agree{.80} & \Agree{.77} & \Agree{.84} & \Agree{.76} & \Agree{.82} & \Agree{.71} \\
el & \Agree{.86} & \Agree{.63} & \Agree{.71} & \Agree{.79} & --- & \Agree{.93} & \Agree{.92} & \Agree{.87} & \Agree{.86} & \Agree{.84} & \Agree{.92} & \Agree{.79} & \Agree{.88} & \Agree{.78} \\
en & \Agree{.87} & \Agree{.62} & \Agree{.73} & \Agree{.82} & \Agree{.89} & --- & \Agree{.95} & \Agree{.89} & \Agree{.88} & \Agree{.85} & \Agree{.92} & \Agree{.80} & \Agree{.90} & \Agree{.78} \\
hi & \Agree{.84} & \Agree{.60} & \Agree{.69} & \Agree{.80} & \Agree{.86} & \Agree{.89} & --- & \Agree{.88} & \Agree{.89} & \Agree{.84} & \Agree{.92} & \Agree{.81} & \Agree{.90} & \Agree{.79} \\
\bottomrule
\end{tabular}
}
\end{minipage}\begin{minipage}{.33\textwidth}
\centering
MEXMA-SigLIP2
\footnotesize
\setlength{\tabcolsep}{2.0pt}
\scalebox{.62}{
\begin{tabular}{lcccccccccccccc}
\toprule
& ar & az & cs & de & el & en & hi & it & ja & ro & ru & sk & uk & vi \\
\midrule
ar & --- & \Agree{.83} & \Agree{.86} & \Agree{.88} & \Agree{.86} & \Agree{.89} & \Agree{.88} & \Agree{.87} & \Agree{.86} & \Agree{.88} & \Agree{.88} & \Agree{.88} & \Agree{.88} & \Agree{.88} \\
az & \Agree{.84} & --- & \Agree{.82} & \Agree{.84} & \Agree{.85} & \Agree{.86} & \Agree{.84} & \Agree{.85} & \Agree{.79} & \Agree{.86} & \Agree{.84} & \Agree{.85} & \Agree{.86} & \Agree{.81} \\
cs & \Agree{.85} & \Agree{.85} & --- & \Agree{.89} & \Agree{.86} & \Agree{.89} & \Agree{.88} & \Agree{.88} & \Agree{.83} & \Agree{.88} & \Agree{.89} & \Agree{.91} & \Agree{.91} & \Agree{.87} \\
de & \Agree{.87} & \Agree{.87} & \Agree{.88} & --- & \Agree{.87} & \Agree{.92} & \Agree{.90} & \Agree{.88} & \Agree{.86} & \Agree{.90} & \Agree{.90} & \Agree{.91} & \Agree{.91} & \Agree{.87} \\
el & \Agree{.85} & \Agree{.83} & \Agree{.84} & \Agree{.86} & --- & \Agree{.89} & \Agree{.87} & \Agree{.88} & \Agree{.82} & \Agree{.88} & \Agree{.86} & \Agree{.87} & \Agree{.88} & \Agree{.85} \\
en & \Agree{.88} & \Agree{.87} & \Agree{.87} & \Agree{.92} & \Agree{.88} & --- & \Agree{.91} & \Agree{.92} & \Agree{.86} & \Agree{.91} & \Agree{.89} & \Agree{.91} & \Agree{.92} & \Agree{.87} \\
hi & \Agree{.85} & \Agree{.84} & \Agree{.85} & \Agree{.88} & \Agree{.85} & \Agree{.88} & --- & \Agree{.88} & \Agree{.88} & \Agree{.90} & \Agree{.88} & \Agree{.89} & \Agree{.89} & \Agree{.88} \\
\bottomrule
\end{tabular}
}
\end{minipage}\begin{minipage}{.33\textwidth}
\centering
SigLIP2 Base
\footnotesize
\setlength{\tabcolsep}{2.0pt}
\scalebox{.62}{
\begin{tabular}{lcccccccccccccc}
\toprule
& ar & az & cs & de & el & en & hi & it & ja & ro & ru & sk & uk & vi \\
\midrule
ar & --- & \Agree{.47} & \Agree{.74} & \Agree{.78} & \Agree{.70} & \Agree{.75} & \Agree{.63} & \Agree{.76} & \Agree{.61} & \Agree{.67} & \Agree{.77} & \Agree{.72} & \Agree{.72} & \Agree{.73} \\
az & \Agree{.55} & --- & \Agree{.45} & \Agree{.50} & \Agree{.45} & \Agree{.47} & \Agree{.42} & \Agree{.49} & \Agree{.35} & \Agree{.48} & \Agree{.49} & \Agree{.46} & \Agree{.46} & \Agree{.51} \\
cs & \Agree{.78} & \Agree{.55} & --- & \Agree{.77} & \Agree{.69} & \Agree{.76} & \Agree{.60} & \Agree{.78} & \Agree{.58} & \Agree{.69} & \Agree{.78} & \Agree{.76} & \Agree{.71} & \Agree{.74} \\
de & \Agree{.79} & \Agree{.55} & \Agree{.83} & --- & \Agree{.74} & \Agree{.85} & \Agree{.66} & \Agree{.85} & \Agree{.62} & \Agree{.72} & \Agree{.87} & \Agree{.78} & \Agree{.75} & \Agree{.80} \\
el & \Agree{.78} & \Agree{.57} & \Agree{.78} & \Agree{.77} & --- & \Agree{.72} & \Agree{.62} & \Agree{.71} & \Agree{.56} & \Agree{.65} & \Agree{.74} & \Agree{.67} & \Agree{.69} & \Agree{.71} \\
en & \Agree{.81} & \Agree{.54} & \Agree{.81} & \Agree{.88} & \Agree{.77} & --- & \Agree{.65} & \Agree{.84} & \Agree{.61} & \Agree{.70} & \Agree{.85} & \Agree{.77} & \Agree{.75} & \Agree{.80} \\
hi & \Agree{.73} & \Agree{.53} & \Agree{.72} & \Agree{.72} & \Agree{.70} & \Agree{.71} & --- & \Agree{.66} & \Agree{.53} & \Agree{.59} & \Agree{.67} & \Agree{.63} & \Agree{.62} & \Agree{.66} \\
\bottomrule
\end{tabular}
}
\end{minipage}

\begin{minipage}{.33\textwidth}
\centering
SigLIP2 Large
\footnotesize
\setlength{\tabcolsep}{2.0pt}
\scalebox{.62}{
\begin{tabular}{lcccccccccccccc}
\toprule
& ar & az & cs & de & el & en & hi & it & ja & ro & ru & sk & uk & vi \\
\midrule
ar & --- & \Agree{.55} & \Agree{.78} & \Agree{.81} & \Agree{.75} & \Agree{.80} & \Agree{.72} & \Agree{.82} & \Agree{.63} & \Agree{.75} & \Agree{.82} & \Agree{.76} & \Agree{.76} & \Agree{.78} \\
az & \Agree{.61} & --- & \Agree{.54} & \Agree{.58} & \Agree{.52} & \Agree{.56} & \Agree{.52} & \Agree{.56} & \Agree{.47} & \Agree{.54} & \Agree{.58} & \Agree{.56} & \Agree{.54} & \Agree{.53} \\
cs & \Agree{.81} & \Agree{.59} & --- & \Agree{.84} & \Agree{.76} & \Agree{.83} & \Agree{.71} & \Agree{.85} & \Agree{.65} & \Agree{.78} & \Agree{.83} & \Agree{.82} & \Agree{.79} & \Agree{.81} \\
de & \Agree{.83} & \Agree{.60} & \Agree{.87} & --- & \Agree{.78} & \Agree{.89} & \Agree{.75} & \Agree{.88} & \Agree{.65} & \Agree{.79} & \Agree{.89} & \Agree{.83} & \Agree{.83} & \Agree{.83} \\
el & \Agree{.80} & \Agree{.59} & \Agree{.81} & \Agree{.83} & --- & \Agree{.79} & \Agree{.71} & \Agree{.78} & \Agree{.60} & \Agree{.72} & \Agree{.79} & \Agree{.74} & \Agree{.75} & \Agree{.76} \\
en & \Agree{.84} & \Agree{.59} & \Agree{.86} & \Agree{.90} & \Agree{.82} & --- & \Agree{.73} & \Agree{.87} & \Agree{.64} & \Agree{.78} & \Agree{.87} & \Agree{.82} & \Agree{.80} & \Agree{.83} \\
hi & \Agree{.79} & \Agree{.60} & \Agree{.78} & \Agree{.80} & \Agree{.76} & \Agree{.82} & --- & \Agree{.74} & \Agree{.60} & \Agree{.67} & \Agree{.74} & \Agree{.72} & \Agree{.72} & \Agree{.71} \\
\bottomrule
\end{tabular}
}
\end{minipage}\begin{minipage}{.33\textwidth}
\centering
SigLIP2 SO400M
\footnotesize
\setlength{\tabcolsep}{2.0pt}
\scalebox{.62}{
\begin{tabular}{lcccccccccccccc}
\toprule
& ar & az & cs & de & el & en & hi & it & ja & ro & ru & sk & uk & vi \\
\midrule
ar & --- & \Agree{.59} & \Agree{.80} & \Agree{.83} & \Agree{.76} & \Agree{.81} & \Agree{.74} & \Agree{.82} & \Agree{.67} & \Agree{.79} & \Agree{.82} & \Agree{.81} & \Agree{.81} & \Agree{.78} \\
az & \Agree{.63} & --- & \Agree{.58} & \Agree{.59} & \Agree{.55} & \Agree{.57} & \Agree{.54} & \Agree{.60} & \Agree{.49} & \Agree{.58} & \Agree{.59} & \Agree{.57} & \Agree{.56} & \Agree{.56} \\
cs & \Agree{.81} & \Agree{.65} & --- & \Agree{.84} & \Agree{.76} & \Agree{.81} & \Agree{.73} & \Agree{.83} & \Agree{.68} & \Agree{.77} & \Agree{.82} & \Agree{.84} & \Agree{.80} & \Agree{.79} \\
de & \Agree{.84} & \Agree{.64} & \Agree{.87} & --- & \Agree{.82} & \Agree{.89} & \Agree{.77} & \Agree{.89} & \Agree{.72} & \Agree{.82} & \Agree{.90} & \Agree{.86} & \Agree{.83} & \Agree{.84} \\
el & \Agree{.83} & \Agree{.62} & \Agree{.84} & \Agree{.84} & --- & \Agree{.80} & \Agree{.73} & \Agree{.80} & \Agree{.66} & \Agree{.76} & \Agree{.80} & \Agree{.79} & \Agree{.78} & \Agree{.77} \\
en & \Agree{.84} & \Agree{.61} & \Agree{.85} & \Agree{.89} & \Agree{.84} & --- & \Agree{.75} & \Agree{.88} & \Agree{.70} & \Agree{.81} & \Agree{.88} & \Agree{.83} & \Agree{.81} & \Agree{.81} \\
hi & \Agree{.81} & \Agree{.66} & \Agree{.80} & \Agree{.81} & \Agree{.80} & \Agree{.83} & --- & \Agree{.77} & \Agree{.64} & \Agree{.74} & \Agree{.75} & \Agree{.75} & \Agree{.74} & \Agree{.73} \\
\bottomrule
\end{tabular}
}
\end{minipage}\begin{minipage}{.33\textwidth}
\centering
SigLIP2 Giant
\footnotesize
\setlength{\tabcolsep}{2.0pt}
\scalebox{.62}{
\begin{tabular}{lcccccccccccccc}
\toprule
& ar & az & cs & de & el & en & hi & it & ja & ro & ru & sk & uk & vi \\
\midrule
ar & --- & \Agree{.54} & \Agree{.78} & \Agree{.81} & \Agree{.75} & \Agree{.80} & \Agree{.71} & \Agree{.79} & \Agree{.68} & \Agree{.77} & \Agree{.81} & \Agree{.79} & \Agree{.79} & \Agree{.80} \\
az & \Agree{.62} & --- & \Agree{.53} & \Agree{.55} & \Agree{.54} & \Agree{.56} & \Agree{.51} & \Agree{.56} & \Agree{.50} & \Agree{.55} & \Agree{.55} & \Agree{.55} & \Agree{.54} & \Agree{.53} \\
cs & \Agree{.83} & \Agree{.61} & --- & \Agree{.84} & \Agree{.76} & \Agree{.81} & \Agree{.72} & \Agree{.81} & \Agree{.69} & \Agree{.79} & \Agree{.84} & \Agree{.84} & \Agree{.79} & \Agree{.79} \\
de & \Agree{.85} & \Agree{.62} & \Agree{.87} & --- & \Agree{.80} & \Agree{.88} & \Agree{.76} & \Agree{.88} & \Agree{.71} & \Agree{.82} & \Agree{.89} & \Agree{.84} & \Agree{.84} & \Agree{.84} \\
el & \Agree{.82} & \Agree{.61} & \Agree{.84} & \Agree{.86} & --- & \Agree{.81} & \Agree{.69} & \Agree{.79} & \Agree{.66} & \Agree{.74} & \Agree{.79} & \Agree{.75} & \Agree{.77} & \Agree{.77} \\
en & \Agree{.85} & \Agree{.61} & \Agree{.86} & \Agree{.91} & \Agree{.84} & --- & \Agree{.75} & \Agree{.88} & \Agree{.70} & \Agree{.80} & \Agree{.87} & \Agree{.81} & \Agree{.83} & \Agree{.82} \\
hi & \Agree{.82} & \Agree{.62} & \Agree{.80} & \Agree{.81} & \Agree{.80} & \Agree{.82} & --- & \Agree{.76} & \Agree{.64} & \Agree{.71} & \Agree{.75} & \Agree{.72} & \Agree{.72} & \Agree{.73} \\
\bottomrule
\end{tabular}
}
\end{minipage}

\caption{Languages' agreement as Cohen-Kappa coefficient for all model.}
\label{tab:cohen_kappa_lang_corr_all}
\end{table*}

% ################################################

\begin{table*}[htbp]
\begin{minipage}{.50\textwidth}
    
\centering
Arabic

\footnotesize
\setlength{\tabcolsep}{2.0pt}
\scalebox{.70}{
\begin{tabular}{lcccccccccccc}
\toprule
& \rotatebox{90}{Gemma 3 12B} & \rotatebox{90}{Qwen 2.5 VL 7B} & \rotatebox{90}{Qwen 3 VL 8B} & \rotatebox{90}{Aya Vision 8B} & \rotatebox{90}{Jina VL} & \rotatebox{90}{NLLB-SigLIP Base} & \rotatebox{90}{NLLB-SigLIP Large} & \rotatebox{90}{MEXMA-SigLIP2} & \rotatebox{90}{SigLIP2 Base} & \rotatebox{90}{SigLIP2 Large} & \rotatebox{90}{SigLIP2 SO400M} & \rotatebox{90}{SigLIP2 Giant} \\
\midrule
Gemma 3 12B & --- & \Agree{.68} & \Agree{.72} & \Agree{.40} & \Agree{.08} & \Agree{.64} & \Agree{.66} & \Agree{.64} & \Agree{.61} & \Agree{.64} & \Agree{.66} & \Agree{.64} \\
Qwen 2.5 VL 7B & \Agree{.79} & --- & \Agree{.78} & \Agree{.34} & \Agree{.10} & \Agree{.67} & \Agree{.68} & \Agree{.66} & \Agree{.62} & \Agree{.62} & \Agree{.64} & \Agree{.63} \\
Qwen 3 VL 8B & \Agree{.82} & \Agree{.83} & --- & \Agree{.40} & \Agree{.09} & \Agree{.73} & \Agree{.73} & \Agree{.73} & \Agree{.65} & \Agree{.67} & \Agree{.68} & \Agree{.68} \\
Aya Vision 8B & \Agree{.74} & \Agree{.73} & \Agree{.76} & --- & \Agree{.06} & \Agree{.39} & \Agree{.39} & \Agree{.38} & \Agree{.36} & \Agree{.37} & \Agree{.38} & \Agree{.37} \\
Jina VL & \Agree{.60} & \Agree{.67} & \Agree{.62} & \Agree{.59} & --- & \Agree{.07} & \Agree{.06} & \Agree{.06} & \Agree{.07} & \Agree{.07} & \Agree{.07} & \Agree{.07} \\
\midrule
NLLB-SigLIP Base & \Agree{.72} & \Agree{.72} & \Agree{.75} & \Agree{.68} & \Agree{.57} & --- & \Agree{.78} & \Agree{.75} & \Agree{.63} & \Agree{.64} & \Agree{.67} & \Agree{.66} \\
NLLB-SigLIP Large & \Agree{.73} & \Agree{.72} & \Agree{.76} & \Agree{.70} & \Agree{.57} & \Agree{.80} & --- & \Agree{.75} & \Agree{.63} & \Agree{.65} & \Agree{.65} & \Agree{.65} \\
\bottomrule
\end{tabular}
}
\end{minipage}\begin{minipage}{.50\textwidth}
\centering
Azerbajani

\footnotesize
\setlength{\tabcolsep}{2.0pt}

\scalebox{.70}{
\begin{tabular}{lcccccccccccc}
\toprule
& \rotatebox{90}{Gemma 3 12B} & \rotatebox{90}{Qwen 2.5 VL 7B} & \rotatebox{90}{Qwen 3 VL 8B} & \rotatebox{90}{Aya Vision 8B} & \rotatebox{90}{Jina VL} & \rotatebox{90}{NLLB-SigLIP Base} & \rotatebox{90}{NLLB-SigLIP Large} & \rotatebox{90}{MEXMA-SigLIP2} & \rotatebox{90}{SigLIP2 Base} & \rotatebox{90}{SigLIP2 Large} & \rotatebox{90}{SigLIP2 SO400M} & \rotatebox{90}{SigLIP2 Giant} \\
\midrule
Gemma 3 12B & --- & \Agree{.50} & \Agree{.58} & \Agree{.30} & \Agree{.05} & \Agree{.39} & \Agree{.44} & \Agree{.55} & \Agree{.36} & \Agree{.42} & \Agree{.43} & \Agree{.42} \\
Qwen 2.5 VL 7B & \Agree{.63} & --- & \Agree{.61} & \Agree{.25} & \Agree{.07} & \Agree{.38} & \Agree{.43} & \Agree{.51} & \Agree{.33} & \Agree{.39} & \Agree{.40} & \Agree{.38} \\
Qwen 3 VL 8B & \Agree{.74} & \Agree{.65} & --- & \Agree{.32} & \Agree{.04} & \Agree{.46} & \Agree{.49} & \Agree{.64} & \Agree{.40} & \Agree{.45} & \Agree{.47} & \Agree{.44} \\
Aya Vision 8B & \Agree{.49} & \Agree{.46} & \Agree{.51} & --- & \Agree{.03} & \Agree{.26} & \Agree{.29} & \Agree{.30} & \Agree{.22} & \Agree{.27} & \Agree{.26} & \Agree{.26} \\
Jina VL & \Agree{.35} & \Agree{.47} & \Agree{.35} & \Agree{.29} & --- & \Agree{.03} & \Agree{.02} & \Agree{.02} & \Agree{.03} & \Agree{.02} & \Agree{.01} & \Agree{.01} \\
\midrule
NLLB-SigLIP Base & \Agree{.43} & \Agree{.43} & \Agree{.46} & \Agree{.33} & \Agree{.25} & --- & \Agree{.55} & \Agree{.56} & \Agree{.32} & \Agree{.36} & \Agree{.38} & \Agree{.33} \\
NLLB-SigLIP Large & \Agree{.51} & \Agree{.47} & \Agree{.56} & \Agree{.39} & \Agree{.28} & \Agree{.55} & --- & \Agree{.58} & \Agree{.39} & \Agree{.43} & \Agree{.43} & \Agree{.43} \\
\bottomrule
\end{tabular}
}
\end{minipage}

\begin{minipage}{.50\textwidth}
\centering
Czech

\footnotesize
\setlength{\tabcolsep}{2.0pt}
\scalebox{.70}{
\begin{tabular}{lcccccccccccc}
\toprule
& \rotatebox{90}{Gemma 3 12B} & \rotatebox{90}{Qwen 2.5 VL 7B} & \rotatebox{90}{Qwen 3 VL 8B} & \rotatebox{90}{Aya Vision 8B} & \rotatebox{90}{Jina VL} & \rotatebox{90}{NLLB-SigLIP Base} & \rotatebox{90}{NLLB-SigLIP Large} & \rotatebox{90}{MEXMA-SigLIP2} & \rotatebox{90}{SigLIP2 Base} & \rotatebox{90}{SigLIP2 Large} & \rotatebox{90}{SigLIP2 SO400M} & \rotatebox{90}{SigLIP2 Giant} \\
\midrule
Gemma 3 12B & --- & \Agree{.68} & \Agree{.69} & \Agree{.44} & \Agree{.09} & \Agree{.46} & \Agree{.57} & \Agree{.63} & \Agree{.59} & \Agree{.62} & \Agree{.61} & \Agree{.62} \\
Qwen 2.5 VL 7B & \Agree{.78} & --- & \Agree{.76} & \Agree{.41} & \Agree{.11} & \Agree{.50} & \Agree{.62} & \Agree{.68} & \Agree{.63} & \Agree{.65} & \Agree{.65} & \Agree{.65} \\
Qwen 3 VL 8B & \Agree{.80} & \Agree{.81} & --- & \Agree{.47} & \Agree{.09} & \Agree{.52} & \Agree{.65} & \Agree{.72} & \Agree{.68} & \Agree{.70} & \Agree{.68} & \Agree{.71} \\
Aya Vision 8B & \Agree{.72} & \Agree{.71} & \Agree{.72} & --- & \Agree{.06} & \Agree{.34} & \Agree{.43} & \Agree{.47} & \Agree{.45} & \Agree{.46} & \Agree{.46} & \Agree{.46} \\
Jina VL & \Agree{.58} & \Agree{.64} & \Agree{.59} & \Agree{.55} & --- & \Agree{.05} & \Agree{.06} & \Agree{.05} & \Agree{.06} & \Agree{.05} & \Agree{.05} & \Agree{.06} \\
\midrule
NLLB-SigLIP Base & \Agree{.45} & \Agree{.48} & \Agree{.48} & \Agree{.43} & \Agree{.39} & --- & \Agree{.56} & \Agree{.55} & \Agree{.47} & \Agree{.45} & \Agree{.46} & \Agree{.48} \\
NLLB-SigLIP Large & \Agree{.62} & \Agree{.65} & \Agree{.65} & \Agree{.58} & \Agree{.50} & \Agree{.57} & --- & \Agree{.68} & \Agree{.59} & \Agree{.61} & \Agree{.61} & \Agree{.60} \\
\bottomrule
\end{tabular}%
}
\end{minipage}

\caption{Models' agreement as Cohen-Kappa coefficient, for each language, part 1.}
\label{tab:cohen_kappa_model_corr_1}
\end{table*}

\begin{table*}[htbp]
\begin{minipage}{.50\textwidth}
    
\centering
Japanese

\footnotesize
\setlength{\tabcolsep}{2.0pt}
\scalebox{.70}{
%
}
\end{minipage}

\caption{Models' agreement as Cohen-Kappa coefficient, for each language, part 2.}
\label{tab:cohen_kappa_model_corr_2}
\end{table*}

% ################################################

\begin{table*}[htbp]
\centering\footnotesize
\setlength{\tabcolsep}{4.2pt}
%
\caption{Confidence intervals for \textit{strict accuracy} obtained with bootstrap.}
\label{tab:strict_acc_bootstrap}
\end{table*}

% ################################################

\begin{table*}[htbp]
\centering\footnotesize
\setlength{\tabcolsep}{4.2pt}
%
\caption{\textit{Strict accuracy} for bidirectional retrieval with the instruction prompt in French, only for LLMs, i.e. $c_{\text{strict}}$ from Equation~(\ref{eq:cstrict}).}
\label{tab:strict_acc_fr_prompt}
\end{table*}

\begin{table*}[htbp]
\centering\footnotesize
\setlength{\tabcolsep}{4.2pt}
%
\caption{Difference between language-specific mistakes with the instruction prompt in French.}

\label{tab:lang_mistakes_fr_prompt}
\end{table*}

% ################################################

\begin{table*}[htbp]
\centering\footnotesize
\setlength{\tabcolsep}{4.2pt}
%
\caption{\textit{Accuracy} of the bilingual retrieval for in the random pairs scenario, \textit{2txt} only.}
\label{tab:bilingual_acc_in_random_pairs}
\end{table*}

\begin{table*}[htbp]
\centering\footnotesize
\setlength{\tabcolsep}{4.2pt}
%
\caption{Difference between language-specific mistakes in the random pairs scenario.}

\label{tab:lang_mistakes_random_pairs}
\end{table*}

% ###################################################

\begin{figure}[t]

\footnotesize
\setlength{\tabcolsep}{1pt}
\begin{tabular}{p{10pt} >{\centering\arraybackslash}p{100pt} >{\centering\arraybackslash}p{100pt}}
\toprule
& Original image & Edited image \\
&
\begin{minipage}[t]{100pt}
    \includegraphics[width=\textwidth]{images/coco_536.png}
\end{minipage}
&
\begin{minipage}[t]{100pt}
    \includegraphics[width=\textwidth]{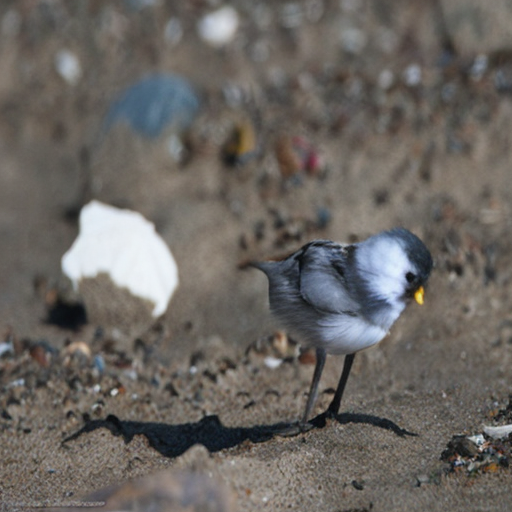}
\end{minipage} \\ \midrule
en & A small bird standing on a rocky ground. & A tiny bird standing on a rocky ground. \\
\bottomrule

\end{tabular}

\caption{Sample from the MultiInpaintCOCO dataset, with a high number of English preference ($ep_{536}=161$).}
\label{fig:mic_sample_2}

\end{figure}

\begin{figure}[t]

\footnotesize
\setlength{\tabcolsep}{1pt}
%

\caption{Sample from the MultiInpaintCOCO dataset, with a high number of English preference ($ep_{185}=0$).}
\label{fig:mic_sample_3}

\end{figure}

\end{document}